\documentclass[11pt]{article}

\usepackage{xcolor}
\usepackage[a4paper,margin=1in]{geometry}
\usepackage{times}
\usepackage{url}
\usepackage{hyperref}
\usepackage{microtype}
\usepackage{setspace}
\usepackage{graphicx}
\usepackage{amsmath,amssymb}
\usepackage{mathtools}
\usepackage{caption}
\usepackage{subcaption}
\definecolor{npciorange}{HTML}{D9782C}
\usepackage{titling}
\usepackage{float}
\usepackage{multirow}
\usepackage{adjustbox}
\usepackage{pgfplots}
\usepackage{xcolor}
\usepackage{booktabs}
\usepackage{float}
\usepackage{colortbl}
\usepackage{tabularx}
\usepackage{enumitem}
\usepackage{titlesec}
\usepackage{array}
\usepackage{parskip}
\usepackage[round]{natbib}
\usepackage{tikz}
\usetikzlibrary{shapes.geometric, arrows.meta, positioning, calc}
\titleformat{\paragraph}
  {\normalfont\normalsize\bfseries}
  {\theparagraph}{1em}{}
\titlespacing*{\paragraph}{0pt}{1ex}{0.5ex}

\hypersetup{
    colorlinks=true,
    linkcolor=blue,
    citecolor=blue,
    urlcolor=blue
}           

\pretitle{%
\noindent
\begin{minipage}[t]{\textwidth}
    \begin{minipage}[t]{0.5\textwidth}
        \includegraphics[height=0.7cm]{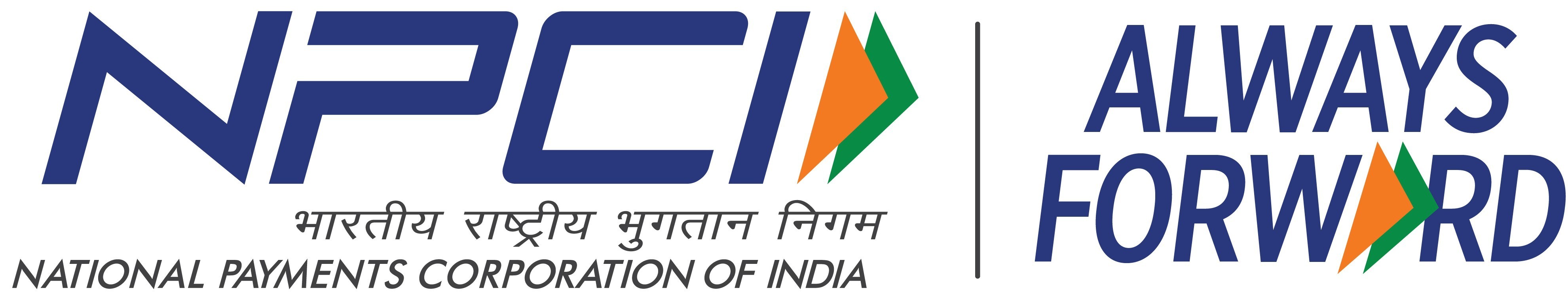}
    \end{minipage}%
    \begin{minipage}[t]{0.5\textwidth}
        \raggedleft
         \today
    \end{minipage}
\end{minipage}
\vspace{-0.35em}
\noindent\rule{\textwidth}{0.4pt}

\vspace{1.0em}   
\centering
}
\title{%
{\color{npciorange}\bfseries\Large
\mbox{FiMI: A Domain-Specific Language Model for Indian Finance Ecosystem}}
}
\posttitle{%
    \par
    \vspace{1.0em} 
}
\preauthor{\centering\large\color{black}\mdseries} 
\author{National Payments Corporation of India (NPCI)}
\postauthor{%
    \par
    \vspace{1.0em} 
}
\predate{}
\date{}
\postdate{}
\author{
National Payments Corporation of India
}
\date{}

\usepackage{fancyhdr}

\fancypagestyle{plain}{
  \fancyhf{}
  \fancyhead[L]{} 
  \fancyhead[R]{} 
  \fancyfoot[R]{\thepage} 
}

\begin{document}

\maketitle

\begin{abstract}
\noindent
We present FiMI (Finance Model for India), a domain-specialized financial language model developed by National Payments Corporation of India (NPCI) for Indian digital payment systems. We develop two model variants: FiMI Base and FiMI Instruct. FiMI adapts the Mistral Small 24B architecture through a multi-stage training pipeline, beginning with continuous pre-training on 68 Billion tokens of curated financial, multilingual (English, Hindi, Hinglish), and synthetic data. This is followed by instruction fine-tuning and domain-specific supervised fine-tuning focused on multi-turn, tool-driven conversations that model real-world workflows, such as transaction disputes and mandate lifecycle management. Evaluations reveal that FiMI Base achieves a 20\% improvement over the Mistral Small 24B Base model on finance reasoning benchmark, while FiMI Instruct outperforms the Mistral Small 24B Instruct model by 87\% on domain-specific tool-calling. Moreover, FiMI achieves these significant domain gains while maintaining comparable performance to models of similar size on general benchmarks.
\end{abstract}

\section{Introduction}

India’s digital payments ecosystem has evolved into one of the world’s most sophisticated, high velocity financial networks, powered by innovations such as the Unified Payments Interface (UPI). Unlike many global financial systems, India’s payments landscape is characterised by real time settlement at massive scale, multi Payment Service Provider (PSP) routing, multilingual consumer behaviour, and a distinctive interplay of regulatory, operational, and ecosystem level dynamics. As the organisation responsible for the design, operation, and governance of UPI, National Payments Corporation of India (NPCI) possesses deep system level insights into transaction patterns, user behaviour, dispute lifecycles, and linguistic diversity that collectively define India’s financial fabric.

In parallel with the rapid evolution of this payments ecosystem, recent advances in Large Language Models (LLMs) have demonstrated strong capabilities in general reasoning and language understanding \citep{DBLP:journals/corr/abs-2005-14165},  \citep{DBLP:journals/corr/VaswaniSPUJGKP17}. However, models trained primarily for broad, general purpose use often struggle to capture the operational depth and contextual specificity required in highly specialised domains such as financial infrastructure. Prior work within the financial sector has shown that domain aligned models significantly outperform generalist baselines by internalising industry specific terminology, workflows, and constraints, as evidenced by systems such as BloombergGPT \citep{wu2023bloomberggptlargelanguagemodel}. These findings highlight the limitations of generic language models when applied to complex, regulated environments.

 We introduce FiMI, a domain specific language model developed by NPCI for the Indian financial ecosystem. FiMI is designed to internalise the terminology, contextual nuances, end to end workflows, and ecosystem level processes intrinsic to India’s digital payments infrastructure. By embedding knowledge of regulatory frameworks, compliance structures, dispute mechanisms, and customer support interactions, FiMI aims to enable AI driven solutions that are aligned with real world operational and regulatory realities across banks, Payment Service Providers (PSPs), fintechs, and other ecosystem participants \citep{kaddour2023challengesapplicationslargelanguage}.

FiMI is developed using a multi stage training paradigm that combines large scale domain exposure with targeted alignment. FiMI Base is obtained by continuously pre-training the Mistral Small 24B Base architecture on a large corpus of India specific financial data. Building on this foundation, FiMI Instruct is further refined through instruction fine-tuning and domain specific supervised fine-tuning. Across both stages, the models are trained on approximately 68 Billion tokens, sourced from curated financial datasets for continuous pre-training (CPT), open source corpora for instruction fine-tuning (IFT), and high quality synthetic datasets for supervised fine-tuning (SFT). This multiphase training approach enables FiMI to internalise domain knowledge spanning UPI terminology and workflows, customer support interactions, compliance requirements, dispute resolution mechanisms, and operational FAQs, resulting in a model that is both linguistically capable and operationally grounded.

The model has already been leveraged in NPCI’s~\href{https://www.upihelp.npci.org.in/}{UPI Help}, enabling natural language conversations for customer assistance, transaction issue  explanations, mandate management (view, pause, unpause, or revoke automatic payments), problem reporting, and general UPI information access. It supports English, Hindi, and Hinglish, reflecting the linguistic reality of user interactions in India’s payments systems. Across these tasks, FiMI demonstrates better performance than general-purpose large language models that lack India specific financial grounding.

\section{Methodology}

\subsection{Why Context Engineering Fails for NPCI ?}
\label{sec:context_limitations}

During the initial development phase, we attempted to build an agentic system using a general-purpose language model and relied heavily on context engineering and prompt design. However, this approach presented several limitations. Although the model demonstrated reasonable tool-calling capabilities, it struggled to consistently interpret domain-specific terminology such as RRN (Retrieval Reference Number), UMN (Unique Mandate Number), and other acronyms and constructs unique to UPI and the Indian digital payments ecosystem.

Multiple prompting methodologies like structured prompting, chain-of-thought (CoT) guidance, and role-based prompting were explored to improve reliability. While these techniques yielded partial improvements, the results remained inconsistent and unpredictable. The agent responses often varied across repeated runs, and the system required extensive trial and error cycles to achieve acceptable accuracy. Additionally, any modification to the workflow or introduction of new capabilities required rewriting or tuning the prompts, making the solution fragile and difficult to scale.



This experience highlighted the limitations of relying solely on prompt engineering to adapt general-purpose large language models to specialised financial workflows. The lack of domain understanding became a bottleneck, directly impacting tool-call accuracy, consistency, and system reliability. These observations led to the conclusion that a domain-aligned model trained or adapted using NPCI-specific terminology, rules, transaction semantics, and contextual data would provide significantly better stability, interpretability, and operational robustness for production-grade use cases.

\subsection{Why Mistral Small 24B?}
\label{sec:why_mistral_24b}

Choosing an appropriate foundation model was critical to ensuring accuracy, reliability, and scalability for NPCI’s agentic systems. The model needed to balance multiple requirements: strong general reasoning, robust Indic-language performance (especially Hindi and Hinglish), low hallucination in structured workflows, predictable tool-calling behaviour, and efficient fine-tuning support. Practical factors such as inference cost, latency, and licensing flexibility were also essential for both experimentation and production deployment.

We benchmarked four candidate models against NPCI specific criteria, including instruction following, multilingual capability, tool-calling stability, multi-task accuracy, fine-tuning readiness, and operational cost. Initial evaluations showed that smaller models such as Mistral 7B \citep{jiang2023mistral7b} and Llama~3~8B \citep{llama3modelcard} were efficient but lacked the depth required for compliance-heavy, multilingual Indian financial workflows. Mistral Nemo~12B \citep{mistralnemo2024} demonstrated strong Indic grounding but delivered lower stability in structured tool-calling settings. A quantitative summary of these comparisons across reasoning, instruction following, and operational suitability is presented in \hyperref[tab:model_comparison] {Table~\ref*{tab:model_comparison}}.

Across all evaluations, Mistral Small~24B \citep{MistralSmall24B} consistently outperformed alternatives, offering superior agentic behaviour, multilingual robustness, and domain adaptability while maintaining manageable inference costs. Its open and permissive nature further enables flexible experimentation and seamless future integration across NPCI systems. Based on these findings, Mistral Small~24B was selected as the foundation for domain adaptation and fine-tuning, providing the optimal balance of performance, reliability, and operational practicality for India-scale financial applications.

\begin{table}[htbp]
\centering
\begin{tabular}{>{\raggedright\arraybackslash}p{2.0cm} c >{\raggedright\arraybackslash}p{2.0cm} >{\raggedright\arraybackslash}p{2.0cm} >{\raggedright\arraybackslash}p{3.0cm} >{\raggedright\arraybackslash}p{3.0cm}}
\toprule
\textbf{Model} &
\textbf{Parameters} &
\textbf{Multi-task accuracy (MMLU)} &
\textbf{Instruction Following (IFEval)} &
\textbf{NPCI-Specific Suitability} &
\textbf{Licensing \& Notes} \\
\midrule

Mistral Small 24B & 24B & 79 & 82.9 &
Best agentic consistency; strong multilingual capability; stable tool calling &
Apache 2.0 \\
\midrule

Llama 3 8B & 8B & 69.4 & 80.4 &
Good general reasoning; insufficient multilingual and compliance-heavy context handling &
Llama 3.1 Community License \\
\midrule

Mistral Nemo 12B & 12B & 68.0 & 62.9 &
Strong Indic grounding; moderate instability in structured tasks &
Apache 2.0 \\
\midrule

Mistral 7B & 7B & 60 & 60 &
Cost-efficient but lacks domain depth and Indic understanding &
Apache 2.0 \\
\bottomrule
\end{tabular}

\caption{Benchmarking foundation model candidates: Comparative analysis of model performance across reasoning and instruction-following metrics, highlighting operational suitability for NPCI-scale financial systems.}
\label{tab:model_comparison}
\end{table}

\subsection{Training Phases}
\label{sec:training_phases}
Building upon the selection of the 
Mistral Small~24B (\S\hyperref[sec:why_mistral_24b]{\ref*{sec:why_mistral_24b}}), we implemented a multi-phase training strategy to seamlessly adapt the foundation model to the complexities of the Indian financial ecosystem.

This approach was designed to progressively internalize specialized knowledge and operational intelligence, moving beyond simple instruction following to achieve true domain-aligned performance. The model was trained in following phases:

\begin{enumerate}
    \item \textbf{Continuous Pre-Training (CPT):} This initial phase involved adapting the Mistral~Small 24B Base model using a curated corpus of India-specific financial data. The objective was to inject core terminology, contextual knowledge, and ecosystem-level processes (such as those related to UPI) without suffering catastrophic forgetting of the model’s general reasoning capabilities.

    \item \textbf{Instruction Fine-Tuning (IFT):} Following CPT, the model underwent instruction fine-tuning on broad, high-quality open-source corpora. This phase was critical for enhancing the model’s ability to interpret complex natural language queries, follow structured multi-step instructions, and ensure robust multilingual capability across English, Hindi, and Hinglish.

    \item \textbf{Domain-Specific Supervised Fine-Tuning (SFT):} The final stage involved SFT using a high-quality synthetic dataset. This task-specific tuning focused on practical, production-grade use cases such as customer support interactions, compliance structures, dispute mechanisms, and operational FAQs, as demonstrated in NPCI’s UPI Help.
\end{enumerate}

This layered training strategy ensures that FiMI possesses the stability, interpretability, and operational robustness required for deployment within the strict compliance boundaries of Indian financial institutions, directly resolving the consistency and accuracy issues highlighted in \S\hyperref[sec:context_limitations]{\ref*{sec:context_limitations}}.

\section{Continuous Pre-Training (CPT)}
\label{sec:cpt}


We apply CPT to adapt the base model more
closely to the Indian financial domain before instruction and
task-specific fine-tuning. The primary motivations are outlined below.

\begin{itemize}   \setlength{\itemsep}{0pt}   \setlength{\parskip}{0pt}   \setlength{\parsep}{0pt}
    \item Increase the model’s exposure to Indian financial and payments terminologies:  
    CPT enables the model to encounter a broader set of documents and communication styles used across the Indian financial ecosystem, helping it form a more grounded understanding of key concepts and practices. When the model already has basic familiarity with the domain, later training stages can focus on instruction quality and task behaviour.

    \item Reduce the mismatch between general pre-training data and local financial usage:  
    Since most large-scale corpora are globally oriented, CPT helps correct for this imbalance by introducing material that reflects Indian regulatory structures and domain-specific conventions. This reduces reliance on assumptions that may not hold in the local financial context.
\end{itemize}

As the next steps in this work, we outline three components that will be discussed in detail: Data Preparation, Evaluation, and Training. The data preparation section will describe how domain-relevant corpora are identified, curated, and filtered for use in the CPT stage. The evaluation section will outline the methods used to assess the impact of CPT on both general and domain-specific performance. Finally, the CPT section will discuss the training methodology.

\subsection{Data Preparation}
\label{sec:cpt_data}

As part of the CPT data curation pipeline, several components are defined to structure how data is collected, processed, evaluated, and prepared for subsequent training stages. These steps help ensure that the model receives well-organized and domain-relevant material during CPT.

\begin{enumerate}
    \item Data Sourcing: Collecting open-source datasets appropriate for the domain.

    \item Dataset Profiling: Examining the sourced dataset and categorizing it into quality tiers to support informed training decisions.
    
    \item Pre-Processing: Cleaning, normalizing and formatting the collected data for use in CPT.
    
    \item Data Composition: Assembling the final mixture of datasets that will be sampled during CPT.
    \item Auxiliary Preparation which includes:
    \begin{enumerate}
        \item Topic Modelling: Identifying themes based structures within the corpus to support later training and evaluation. This method is used to create training and test splits that maintain balanced topic coverage across both sets.

        \item Question and Answer (QA) Generation: Creating QA material that supports downstream training. This component is derived from a large portion of the dataset and is introduced toward the end of CPT to further strengthen domain coverage.
    \end{enumerate}


\end{enumerate}

\subsubsection{Data Sourcing}
\label{sec:data_sourcing}

We collected large-scale and diverse textual data from multiple open datasets released under permissive licenses. All sources were reasonably reviewed for quality, and suitable subsets were selected for different stages of CPT, following the data processing workflow described earlier. 

\noindent The data collection strategy was guided by two primary objectives: broad general-purpose coverage and strong India-specific financial context. In total, we assembled a corpus of approximately 30 Trillion tokens, spanning general English, mathematics, code, multilingual data, and finance-domain context.
A high-level categorization of the general-purpose and domain-specific components within this corpus is provided in \hyperref[tab:dataset_overview] {Table~\ref*{tab:dataset_overview}}.

\begin{table}[htbp]
\centering
\begin{tabular}{@{}p{8cm} p{5cm}@{}}
\hline
\textbf{General Dataset} & \textbf{Indian Finance} \\
\hline
Open Crawl based web corpus data & Investment schemes \\
Strong foundation for reasoning and code & Insurance services \\
Broad general knowledge coverage & Finance news articles \\
Linguistic diversity, including Indic languages & Payments ecosystem \\
High-quality academic and reference data & Economic and monetary system \\
\hline
\end{tabular}

\caption{Data corpus stratification: Categorization of the 30 Trillion token pre-training corpus, distinguishing between general-purpose foundational data and localized Indian financial domain content.}
\label{tab:dataset_overview}
\end{table}

\subsubsection{Dataset Profiling}
\label{sec:profiled_datasets}



Our data profiling methodology employs a Human-in-the-Loop (HITL) \citep{tarun2025human} alignment framework. We aggregated samples from our General Text, Mathematics, and Code corpora and subjected them to rigorous HITL evaluation. These human annotations served as the ground truth to create a specialized pipeline. This pipeline was then deployed to score and profile the corpus for categorizing data based on its alignment with our training objectives. 


The resulting profiles revealed distinct but complementary roles for different data sources. Segments with high alignment scores often found in synthetic or curated academic subsets are utilized to anchor the model’s reasoning capabilities and stabilize training. Meanwhile, segments with broader alignment scores, such as those from diverse web scrapes, are retained not as noise but as critical exposure to real-world variability. 

Based on these alignment scores, subsets were selected
from each source. After initial profiling, approximately 30 Trillion tokens were identified, which were
then further filtered through a pre-processing pipeline that curated the dataset down to 2.2 Trillion tokens.
From this filtered corpus, 68 Billion high-quality tokens were selected for CPT.
The collated dataset contains the following data types:
\begin{itemize}   \setlength{\itemsep}{0pt}   \setlength{\parskip}{0pt}   \setlength{\parsep}{0pt}
  \setlength{\itemsep}{0pt}
  \setlength{\parskip}{0pt}
  \setlength{\parsep}{0pt}
  \item Text Data
  \item Math Data
  \item Code Data
  \item Finance Data
  \item Multilingual (Hindi) Data
\end{itemize}






The comprehensive mapping of raw data sources, their specific partitions, and their functional roles across text, mathematics, and code domains is detailed in \hyperref[tab:data_sources] {Table~\ref*{tab:data_sources}}.

\begin{table}[htbp]
\centering
\begin{tabular}{
>{\raggedright\arraybackslash}p{2cm}
>{\raggedright\arraybackslash}p{11cm}
>{\raggedright\arraybackslash}p{1.5cm}
}
\toprule
\textbf{Data Source} & \textbf{Partitions} & \textbf{Used For} \\
\midrule

\multirow{3}{*}{Dolma} &
Falcon, c4\_filtered, cc\_en\_tail, cc\_en\_middle, wikiref\_megawika, cc\_en\_head, wiki &
Text \\
\cmidrule(lr){2-3}
&
proof\_pile\_2-algebraic\_stack, proof\_pile\_2-open\_web\_math &
Math \\
\cmidrule(lr){2-3}
&
starcoder &
Code \\
\midrule

\multirow{3}{*}{NeMo} &
For\_General, Synthetic &
Text \\
\cmidrule(lr){2-3}
&
4plus, 4plus\_MIND &
Math \\
\cmidrule(lr){2-3}
&
nemo\_synthetic\_code &
Code \\
\midrule

FineWeb &
CC-MAIN Partitions &
Text \\
\midrule

Ai4Bharat &
Sangraha Corpus &
Text \\
\midrule

CommonPile &
arxiv\_abstracts\_filtered, arxiv\_papers\_filtered, oercommons\_filtered, caselaw\_access\_project\_filtered, cccc\_filtered, doab\_filtered, pressbooks\_filtered, project\_gutenberg\_filtered, uspto\_filtered, wikimedia\_filtered, wikiteam\_filtered, peS2o\_filtered &
Text \\
\midrule

FineMath &
finemath-3plus, finemath-4plus, infiwebmath-3plus, infiwebmath-4plus &
Math \\
\bottomrule
\end{tabular}

\caption{Data Source and Partition Taxonomy. The table highlights how heterogeneous corpora are systematically organized and selectively employed to support domain-specialized learning objectives.}

\label{tab:data_sources}
\end{table}

\noindent \textbf{Text Dataset}

\begin{figure}[htbp]
    \centering
    \includegraphics[width=0.9\linewidth]{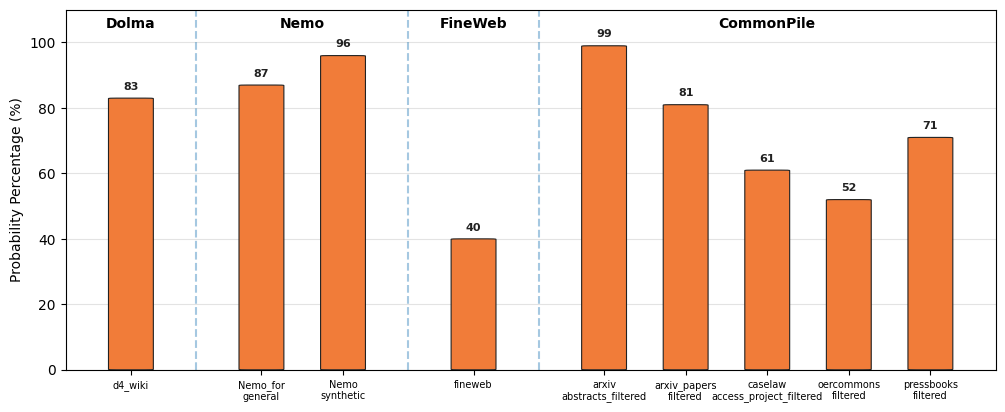}
    \caption{Analysis of token distribution across the selected corpora, illustrating the effectiveness of the filtering pipeline in isolating high-Alignment segments from diverse sources like Dolma, NeMo, FineWeb, and CommonPile.}    
    \label{fig:quality_distribution_text}
\end{figure}

\noindent Text dataset comprises of web content, articles, narratives, wikipedia-like content, any non-code and non-math natural language. The alignment distribution shown in Figure~\ref{fig:quality_distribution_text} illustrates two complementary contributions within the CPT corpus. Datasets such as d4-wiki (Dolma) 
\citep{soldaini2024dolmaopencorpustrillion},  
NeMo Synthetic and For\_General \citep{nvidia2025nvidianemotronnano2} 
contain a higher proportion of strongly aligned text, reflected in improved grammatical consistency, clearer discourse organization, and overall readability. Complementing this foundation, carefully selected high-alignment subsets from CommonPile \citep{kandpal2025common} including academic, legal, and educational sources provide more formal, domain-specific content, characterized by richer structural patterns and longer-range linguistic dependencies.

\noindent Together, the combination of clean general-language data and curated domain-focused text enables the model to balance fluency with depth, improving both generalization and practical applicability across diverse text-based tasks.

\noindent Overall, the CPT corpus now emphasizes alignment-first curation, combining clean general-purpose datasets (Dolma, Nemo) with structured CommonPile sources, while using FineWeb \citep{penedo2024the} to maintain real-world linguistic diversity. A detailed breakdown of dataset-wise alignment statistics and folder-level contributions is provided later in the document in the Data Composition \hyperref[tab:data_composition_table]{Table~\ref*{tab:data_composition_table}} .

\vspace{0.5em}
\noindent \textbf{Math Dataset}

\begin{figure}[htbp]
    \centering
    \includegraphics[height=0.25\textheight]{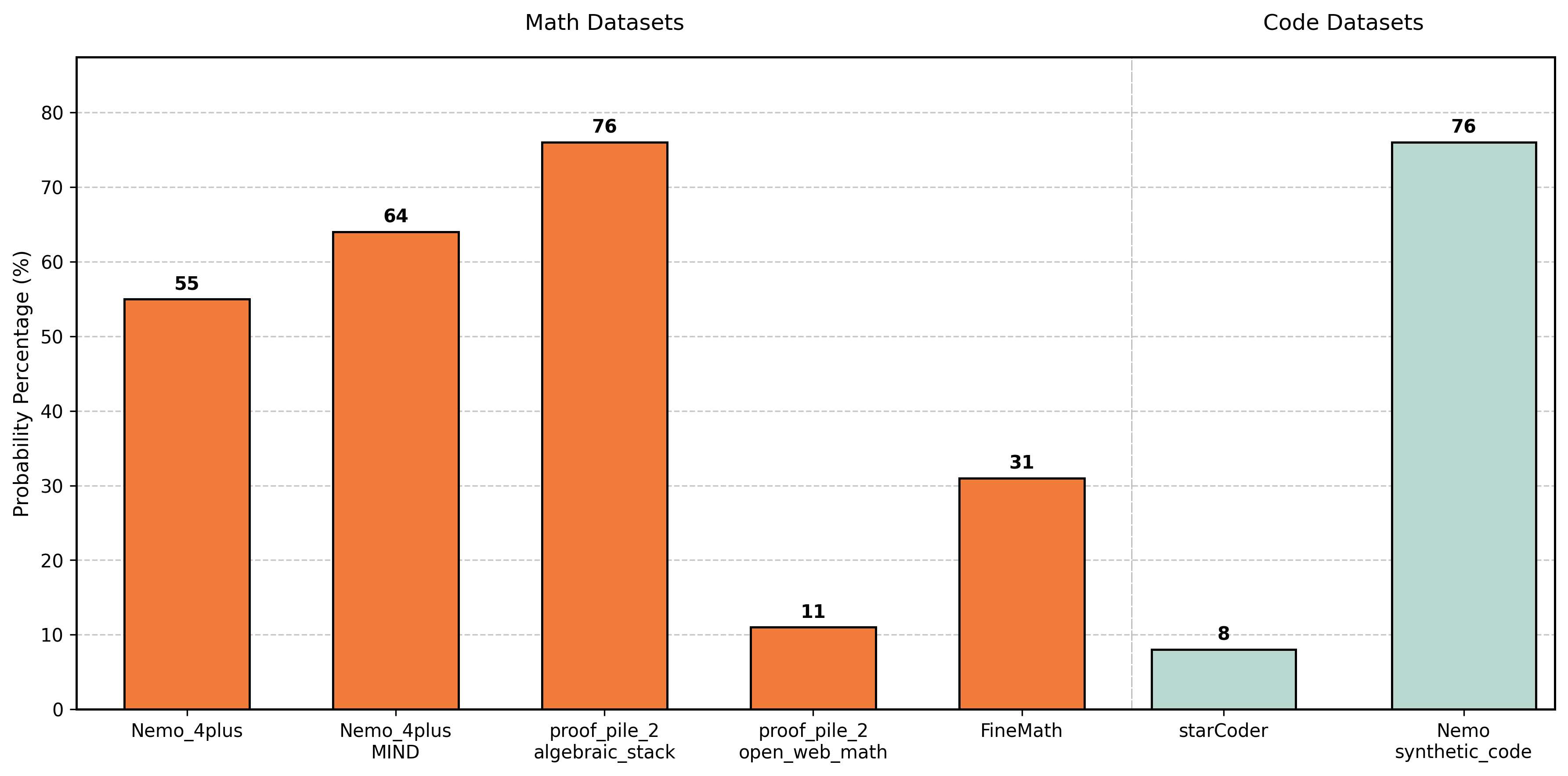}
    \caption{Reasoning Density and Structural Quality Across Math and Code Corpora. NeMo and ProofPile-2 exhibit high structured reasoning, while code datasets balance real-world diversity with curated synthetic structure.}
    \label{fig:math_and_code_distribution}
\end{figure}

The math dataset contains mathematical expressions, structures, proofs, equations, theorems, and reasoning-style mathematical text. The Alignment distribution across these sources, as illustrated in \hyperref[fig:math_and_code_distribution] {Figure~\ref*{fig:math_and_code_distribution}}, reveals two distinct categories of data based on their structural and logical rigor.

Datasets such as NeMo-4Plus, NeMo-4Plus-MIND \citep{nvidia2025nvidianemotronnano2}, and ProofPile-2 Algebraic Stack \citep{dolma} contain a higher proportion of high-Alignment mathematical text, with strong coherence, structured formatting, and reliable step-by-step reasoning. These datasets provide the core mathematical grounding required for stable pre-training and improve the model’s ability to handle formal reasoning and structured explanation.
 
In contrast, datasets such as ProofPile-2 OpenWebMath \citep{dolma}, FineMath \citep{allal2025smollm2smolgoesbig} reflect a much broader range of writing styles, as they draw directly from open web sources where mathematical expressions and explanations can vary significantly. This diversity introduces material that is more informal, conversational, or statistically inconsistent compared to curated datasets. However, it also brings in a wide spectrum of real-world problem-solving approaches, intuitive explanations, and community-generated mathematical discussions. These characteristics make the dataset valuable for helping the model understand and adapt to the varied mathematical inputs it may encounter in practical usage.

\vspace{0.5em}
\noindent \textbf{Code-centric Dataset}
.\noindent Code-centric data contains code blocks, syntax highlighting patterns, language keywords, brackets, and indentation patterns. The Alignment distribution across these sources is summarized in \hyperref[fig:math_and_code_distribution] {Figure~\ref*{fig:math_and_code_distribution}}.

\noindent The Code-centric datasets show two complementary strengths. StarCoder \citep{dolma} draws from a very broad range of real-world code, capturing diverse programming styles, informal patterns, and community-generated solutions. On the other hand, the Nemo\_synthetic\_code \citep{nvidia2025nvidianemotronnano2} dataset provides more structured, consistently formatted examples created with clearer instructions. This kind of data is especially useful for learning precise reasoning patterns, clean abstractions, and reliable coding structures.

\subsubsection{Data Pre-processing}
\label{sec:data_pre-processing}

\begin{figure}[htbp]
    \centering
    \includegraphics[width=0.9\linewidth]{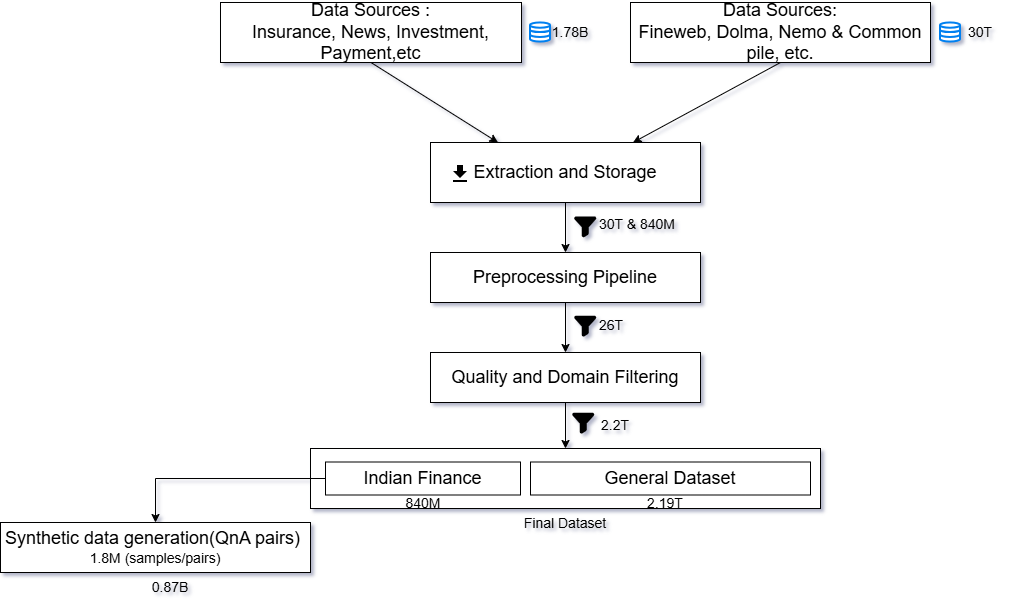}
    \caption{Data Curation and CPT Pipeline. End-to-end workflow illustrating raw data ingestion through multi-stage pre-processing, quality filtering, anonymization, and domain-specific partitioning for the construction of general-purpose and specialized financial corpora.}
    \label{fig:data_curation_pipeline}
\end{figure}

The high-alignment profiled dataset described in
\S\hyperref[sec:profiled_datasets]{\ref*{sec:profiled_datasets}} was pre-processed using a combination
of automated tools and custom scripts to produce an optimally curated
set of tokens. An overview of the resulting data composition is provided
in \hyperref[tab:data_composition_table] {Table~\ref*{tab:data_composition_table}}, and the end-to-end workflow is
illustrated in \hyperref[fig:data_curation_pipeline] {Figure~\ref*{fig:data_curation_pipeline}}.

The key pre-processing steps are summarized below.

\begin{enumerate}
    \item Text extracted from the collected sources was cleaned using
    reliable parsing techniques to isolate meaningful content and ensure
    consistent formatting across diverse inputs.

    \item Common noise, inconsistencies, and structural irregularities
    were addressed through normalization and targeted filtering, allowing
    the text to retain semantic value while removing elements that could
    interfere with downstream training.

    \item Duplicate and near-duplicate passages were identified using
    semantic similarity techniques and removed to preserve corpus
    diversity and avoid over-representation of repetitive patterns,
    following established best practices for large-scale data
    deduplication~\citep{you2025evaluatingdeduplicationtechniqueseconomic}.

    \item A rule-based filtering stage was applied to exclude entries
    that were unusually short, noisy, or lacking substantive
    information, thereby improving overall dataset quality.

    \item Personally Identifiable Information (PII), including names,
    account references, and contact details, was replaced with
    anonymized placeholders in accordance with ethical and regulatory
    standards for model training.

    \item The curated dataset was further classified using a three-level
    quality system (High, Medium, Low). Despite extensive cleaning, some
    residual noise is expected; this categorization helps identify
    incomplete, irrelevant, or low-value samples. Low-quality content
    primarily included poorly written text, unrelated material, and
    samples with limited informational value.

    \item To enhance multilingual coverage, a high-throughput,
    multi-node deployable translation pipeline was developed to generate
    Hindi tokens and additional multilingual data.

    \item To analyze and refine the distribution of domain-relevant
    content, a dedicated domain classifier based on a BERT-style
    architecture was applied~\citep{devlin2019bertpretrainingdeepbidirectional}.
    This classifier enabled separation of Indian finance–specific
    content from general text and facilitated diversity analysis within
    the domain.

    \item The final curated dataset was converted into a
    text-completion format and used for multi-stage CPT, with
    question–answer pairs generated from the text and introduced
    progressively across training stages.
\end{enumerate}

\subsubsection{Data Composition}
\label{sec:data_composition}

All the corpus consisting of 30 Trillion tokens went through our pre-processing steps, which brought the total down to about 2.2 Trillion filtered tokens balancing overall quality, domain coverage, and diversity, and is used as the final input for CPT pre-training. We then ran a separate quality-evaluation stage to check how clean, useful, and relevant the remaining data was. Based on this review, we sampled a final set of 68 Billion high-quality tokens to use for CPT. 
Summary of token counts before and after pre-processing for the various components of the CPT corpus, featuring datasets from Dolma, FineWeb, Nemotron Pre-training Dataset, Common Pile, Ai4Bharat Sangraha \citep{khan2024indicllmsuite}, and FineMath are shown in \hyperref[tab:data_composition_table]{Table~\ref*{tab:data_composition_table}}. 

\begin{table}[htbp]
\centering
\renewcommand{\arraystretch}{1.3}
\begin{tabular}{llrr}
\toprule
\textbf{Data Type} & \textbf{Dataset} & \multicolumn{2}{c}{\textbf{Token count in Billions}} \\ 
\cmidrule(lr){3-4}
 & & \textbf{Downloaded / Available} & \textbf{Filtered / Curated} \\ 
\midrule
General English & Dolma v1.7 & 2,019.50 & 608.50 \\
 & FineWeb & 18,427.00 & 337.40 \\
 & Common Pile v0.1 Filtered Data & 2,134.80 & 90.00 \\
 & Nemotron-CC-v2 & 6,585.80 & 810.00 \\
\midrule
Maths & Nemotron-pre-training-Math & 133.00 & 116.00 \\
 & Dolma--Proof Pile II & 25.20 & -- \\
 & Finemath & 88.00 & 35.00 \\
\midrule
Code & Nemotron-pre-training-Code & 223.40 & 134.90 \\
 & Dolma Starcoder & 263.80 & -- \\
\midrule
Hindi & Ai4Bharat Sangraha & 34.54 & 32.00 \\
\midrule
Finance English & FineWeb & 60.00 & 51.00 \\
\midrule
Finance Hindi & Ai4Bharat Sangraha & 4.00 & 4.00 \\
\midrule
Indian Finance & Economic and monetary system & -- & 0.24 \\
 & Investment schemes & -- & 0.11 \\
 & Insurance services & -- & 0.15 \\
 & Finance news articles & -- & 0.20 \\
\midrule
\textbf{Total} & & \textbf{30,000.00} & \textbf{2,220.00} \\ 
\bottomrule
\end{tabular}
\caption{Training data composition for CPT before and after pre-processing in tokens (Billions).}
\label{tab:data_composition_table}
\end{table}

\noindent \textbf{Topic Modelling for Auxiliary Data Preparation}
\label{sec:topic_modelling}

\noindent To effectively tune the model for the Indian financial context, a Top-Down Topic Modelling approach was developed to identify and hierarchically structure the major themes present in the corpus. This method provides a clear view of how different topics are distributed across the dataset. The topic-modelled corpus is then used to create balanced training and test splits, ensuring that each topic is proportionally represented in both subsets rather than being unevenly distributed through random partitioning. Using this approach, the dataset is divided into a 95\% training split and a 5\% test split. This train dataset is introduced at the last stage of CPT as this dataset tunes the model most towards Indian Financial Context.

The primary domains considered in this work, sourced from open web data, include the economic and monetary system, investment schemes, insurance services, financial news articles, and the payments ecosystem.


\vspace{0.5em}
\noindent \textbf{Top-Down Topic Modelling Approach for Structuring Finance Data}

\vspace{0.3em}

\noindent This process uses an LLM and semantic analysis to automatically construct a detailed, multi-level taxonomy. The method deploys a top-down hierarchical grouping strategy to organize the Indian Finance dataset, systematically dividing the data into smaller, more specific subdomains. The resulting structure is a tree where each leaf node uniquely represents a distinct sub-topic along with its associated data.

\vspace{0.3em}




\noindent Phase 1: Domain Initialization\\
The dataset is initially partitioned into a set of broad financial domains based on domain knowledge and corpus-level inspection. Each document is assigned to a relevant top-level domain to establish a coarse hierarchical structure that serves as the starting point for downstream topic modeling.
\vspace{0.3em}

\noindent Phase 2: LLM-Driven Semantic Assignment\\
An LLM first proposes potential subtopics derived directly from the domain data. Both source data and these proposed subtopics are then converted into vector embeddings, allowing each data row to be assigned to its most relevant subtopic based on cosine similarity scores.
\vspace{0.3em}

\noindent Phase 3: Adaptive Tree Refinement\\
The hierarchy undergoes dynamic modification where over-sized groups are treated as new parent nodes and recursively split via the previous phase until specificity targets are met. Conversely, groups that fall below a size threshold are preserved as leaf nodes and flagged for future data augmentation. Finally, unclustered residual groups are analyzed by the LLM to generate precise, descriptive labels, avoiding generic categorizations.

\vspace{0.5em}

\begin{figure}[htbp]
    \centering
    \includegraphics[width=0.7\linewidth]{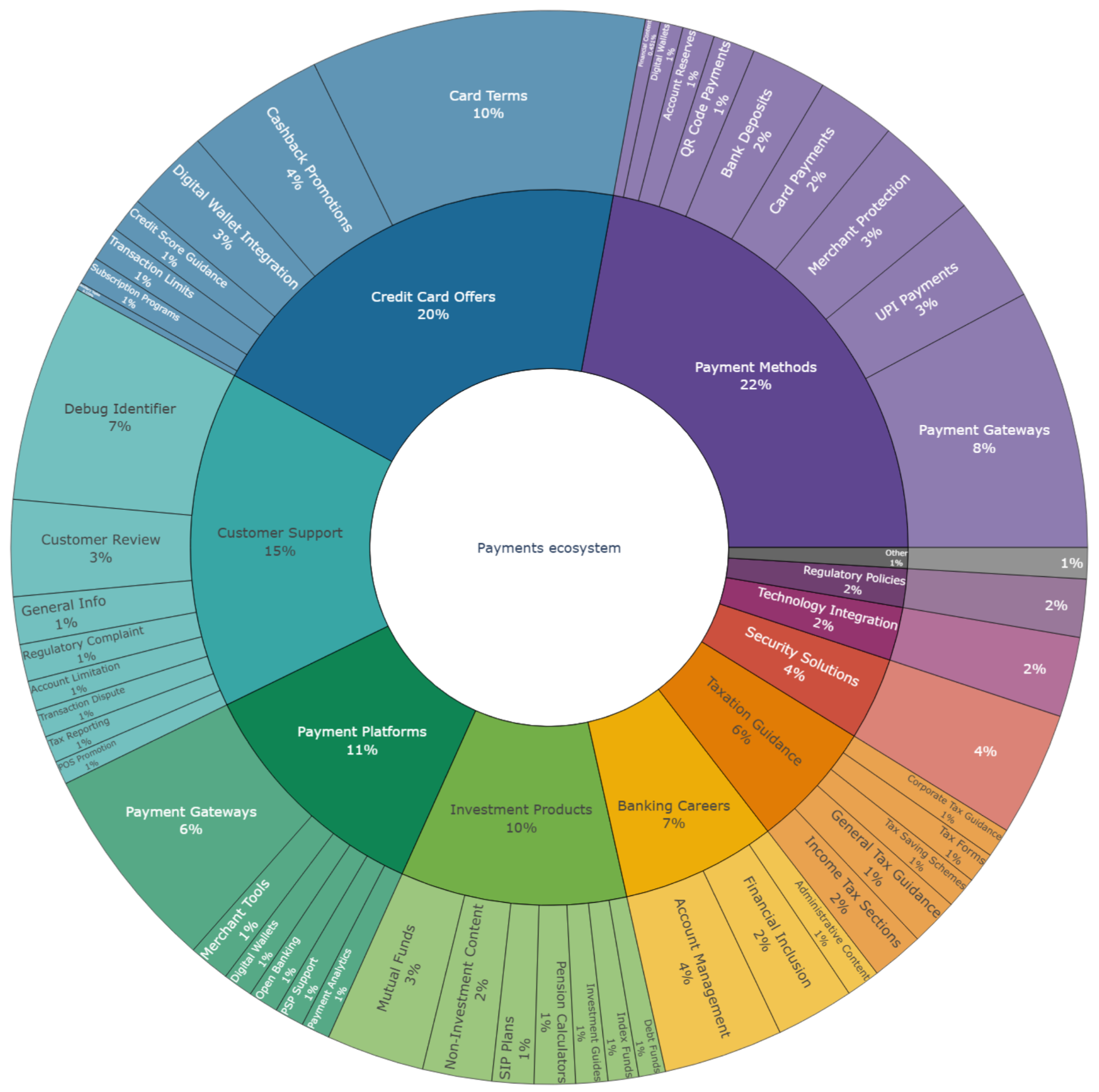}
    \caption{Sunburst diagram representing payments ecosystem taxonomy. This Topic modelling representation was created for all the 5 main domains}
    \label{fig:sunburst_diagram}
\end{figure}

\subsubsection{Generation of Question--Answer Pairs (Auxiliary Preparation)}
\label{sec:qa_generation}
We created 1.8~million question--answer samples, comprising approximately 0.87~Billion tokens, from the training portion (95\%) using on-prem hosted models like Gemma 3 27B \citep{gemmateam2025gemma3technicalreport} (which is explained later in \S\hyperref[sec:phase2_data]{\ref*{sec:phase2_data}}. These question--answer pairs are used in the final stage of CPT to help the model develop a deeper understanding of the Indian financial context.

This approach allows us to avoid exposing the model to residual noise or unintended formatting that can remain in large text corpora even after extensive cleaning. By leveraging structured question--answer pairs, we ensure that the model learns the underlying content and concepts rather than inheriting inconsistencies present in raw data.

\subsection{Evaluation}
\label{sec:cpt_evaluation}
After CPT, model evaluation should address two dimensions: generalization retention and domain specialization. General benchmark evaluations are required to quantify any performance regression or catastrophic forgetting introduced by CPT, while domain-specific benchmarks must measure capability gains within Indian finance. Accurate assessment depends on high-quality datasets and standardized methodologies such as accuracy, log-likelihood–based scoring, and distributional divergence metrics to ensure unbiased results \citep{biderman2024lessonstrenchesreproducibleevaluation}.

\subsubsection{Evaluation Datasets}

We organized datasets into two groups:
general benchmarks to check broad capability, and custom finance benchmarks to validate specialization.
\\[6pt]
\noindent \textbf{A. General Benchmarks:}
\begin{itemize}   \setlength{\itemsep}{0pt}   \setlength{\parskip}{0pt}   \setlength{\parsep}{0pt}
  \setlength{\itemsep}{0pt}
  \setlength{\parskip}{0pt}
  \setlength{\parsep}{0pt}
    \item ARC Challenge \citep{clark2018thinksolvedquestionanswering}
    \item GSM8K \citep{cobbe2021trainingverifierssolvemath}
    \item Humaneval Instruct \citep{chen2021evaluatinglargelanguagemodels}
    \item MMLU \citep{hendrycks2021measuringmassivemultitasklanguage}
    \item MMLU Pro \citep{wang2024mmluprorobustchallengingmultitask}
    \item Trivia QA \citep{joshi2017triviaqalargescaledistantly}
\end{itemize}

\noindent \textbf{B. Custom Finance Datasets}

\begin{itemize}   \setlength{\itemsep}{0pt}   \setlength{\parskip}{0pt}   \setlength{\parsep}{0pt}
    \item \textbf{ General MCQ Finance Dataset (HellaSwag Finance Validation)} 
    
    \vspace{0.5em}
    
    This benchmark helps us understand whether the model is regressing after training. We generated approximately 2,700 HellaSwag-inspired questions \citep{zellers2019hellaswagmachinereallyfinish} on Indian finance for the General Finance Dataset. After using LLMs for generation and quality filtering, a benchmark set of 2,471 questions was finalized.
    
    \vspace{0.5em}
    
    Sample question from HellaSwag Finance: \\
    \textit{
    Question: A general insurer faces disputes over liability claims. Policyholders challenge denial decisions. To resolve, the insurer, \\
    Option A: reviews claims with independent adjusters., \\
    Option B: bases decisions on claim amount alone., \\
    Option C: denies claims without review., \\
    Option D: relies on policyholder profile for resolution.
    }
    
    \vspace{1em}
    
    \item \textbf{Finance Reasoning Dataset}
    
    \vspace{0.5em}
    
    To test richer reasoning, we expanded from hand-curated seed examples adapted from MMLU-Pro and contextualized with our Indian Finance corpus. This amounts to around 300 questions. We also often interchangeably refer to this dataset as CPT Eval.
    
    \vspace{0.5em}
    
    Sample question from Finance Reasoning: \\
    \textit{
    How do revised IMPS dispute TATs improve customer experience?
    }
    
    \vspace{1em}
    
    \item \textbf{ Finance Datasets Based on Standard General Benchmarks}
    
    \vspace{0.5em}
    The methods discussed in above two points cover two ends of the evaluation spectrum: knowledge-retention datasets (such as HellaSwag-style questions) and financial reasoning questions. To bridge this gap, we developed an additional benchmark that combines reasoning with financial facts, aimed at standardizing Indian financial question--answering datasets. This benchmark is designed to closely mimic the MMLU standard and this is later referenced as MMLU Finance Validation dataset.

\vspace{0.5em}
The detailed analysis of MMLU dataset is available in the  \hyperref[sec:annexuremmlu]{Annexure A}. Questions were generated using Bring Your Own Benchmark (BYOB) framework which includes a modular LLM framework that automates the creation of custom evaluation datasets (similar to MMLU) by generating, expanding, deduplicating, and filtering questions based on subject, difficulty, question type, and Bloom’s taxonomy level. The generation process used a 5\% topic-level test split of the Indian finance dataset. At this stage around 5,700 questions were generated and these were sent to the Human in the Loop evaluation to filter out the questions that are irrelevant in the CPT model evaluation. Around 400 questions got filtered out. The final dataset accounts for 5,300 questions.

\end{itemize}

\subsubsection{Measuring Strategies}

\vspace{0.3em}

Model evaluation depends on the question type. Simple accuracy is used for tests like MMLU, while methods like Bertscore or LLM-as-a-Judge assess reasoning.

    
    
    
        


\textbf{Evaluation Harness Framework}

\vspace{0.3em}

We evaluate the CPT model on general benchmarks using lm-eval-harness \citep{biderman2024lessonstrenchesreproducibleevaluation}, a framework that standardizes performance measurement through consistent evaluation settings. Our methodology employs few-shot prompting to assess the model's in-context learning by prepending example input-output pairs to the prompt, alongside CoT prompting. The latter requires the model to generate intermediate logical reasoning steps before providing a final answer, effectively measuring its capacity for complex, multi-step problem solving.

    \vspace{0.2em}

    \textbf{Methodologies for Finance Datasets}
    
    \vspace{0.2em}
    
    \begin{itemize}   \setlength{\itemsep}{0pt}   \setlength{\parskip}{0pt}   \setlength{\parsep}{0pt}
    
        \item \textbf{Methodology Used to Test MCQ-Based Questions} 
        
        Pretrained models are primarily evaluated using accuracy-based metrics. Evaluation frameworks such as the lm-eval-harness also assess inherent model knowledge through conditional probability estimation. This approach avoids ambiguity arising from free-form text generation by computing the average log-likelihood for each candidate answer, normalized by token length, and selecting the option with the highest normalized score based purely on the model’s internal knowledge representation.

        \vspace{0.5em}
        
        \item \textbf{Framework Used to Test Statement-Based Questions}
        
        We evaluated statement-based reasoning using DeepEval \citep{deepeval_github} to facilitate structured test cases and automated scoring via a two-stage pipeline. In the Generation Phase, the pretrained model generates responses to input questions via an API. These are then passed to the DeepEval Scoring Phase, where evaluation instances comprising the question, a reference answer, and the generated response are assessed. A judge model applies a custom correctness metric, assigning a similarity score from 1 to 10 based on how effectively the generated output preserves the intent, meaning, and reasoning structure of the ideal answer.
    \end{itemize}

     \begin{figure}[htbp]
            \centering
            \includegraphics[width=0.9\linewidth]{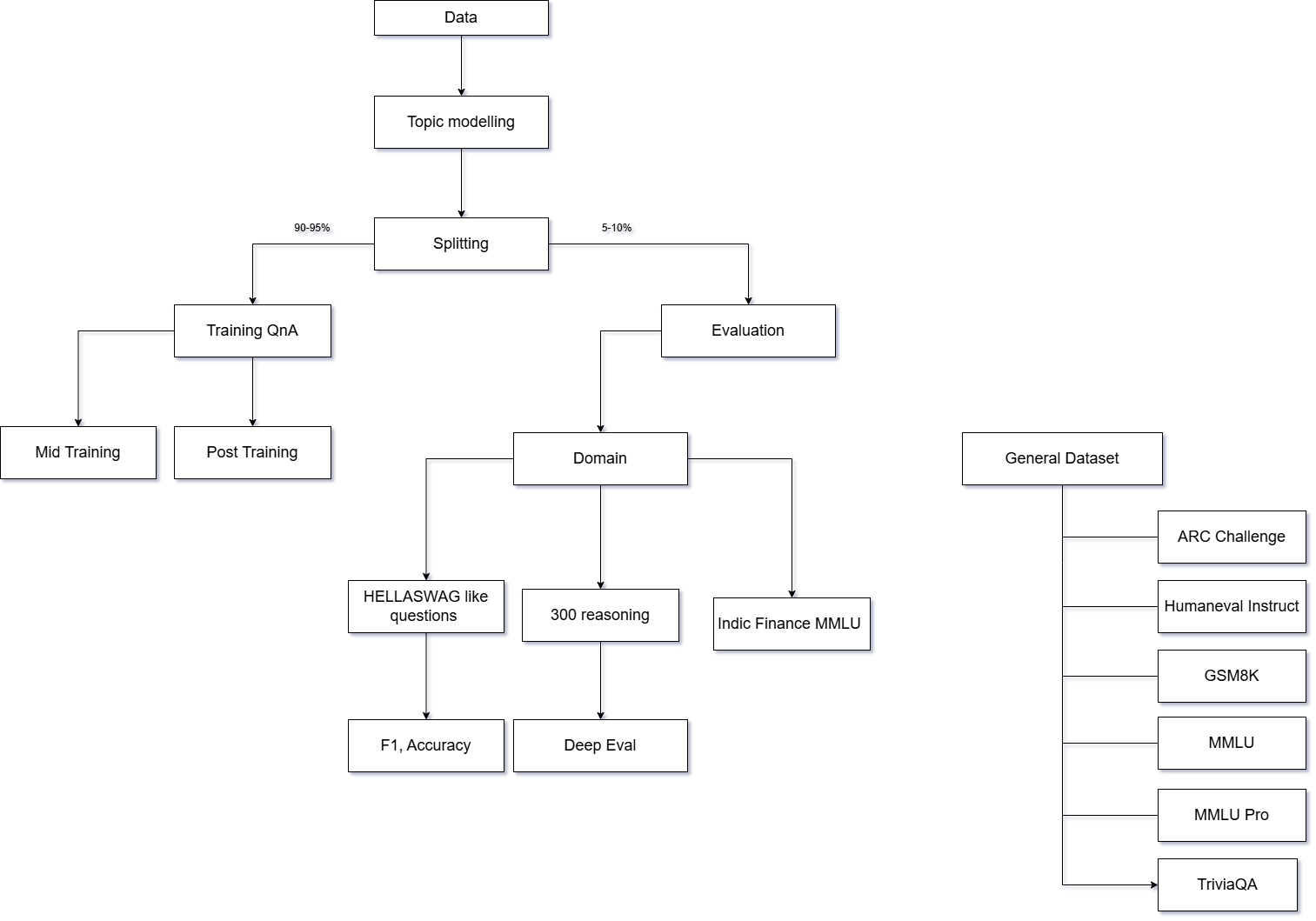}
            \caption{Automated Evaluation Pipeline. Workflow illustrating the two-stage assessment process for reasoning-based tasks, highlighting the generation phase using pretrained model APIs and the subsequent scoring phase using the DeepEval framework with a dedicated judge model.}
            \label{fig:evaluation_workflow}
        \end{figure}
\textbf{KL Divergence Measure}
        
        Kullback--Leibler (KL) divergence \citep{kullback1951information, basterrech2022tracking} indicates how much the probability distribution of the CPT model has shifted relative to the base model, thereby quantifying the extent of change introduced during training. Very high divergence values (for example, $\geq 1$) can indicate potential overfitting which corresponded to substantial downstream performance degradation in our validation suite. This metric serves as a safeguard to ensure that the model acquires new domain-specific knowledge while retaining its general understanding, thereby avoiding excessive drift or over-specialization.

We compute the KL divergence by comparing the next-token probability distributions produced by the continual pre-trained (CPT) model and the base model at each token position in a sequence, conditioned on the same preceding context. For every token, the KL divergence is calculated between the two full vocabulary distributions using the below formula.

        \[
D_{\mathrm{KL}}
\Big(
P_{\text{CPT}}(\cdot \mid x_{<t})
\;\|\;
P_{\text{Base}}(\cdot \mid x_{<t})
\Big)
=
\sum_{w \in \mathcal{V}}
P_{\text{CPT}}(w \mid x_{<t})
\left(
\log P_{\text{CPT}}(w \mid x_{<t})
-
\log P_{\text{Base}}(w \mid x_{<t})
\right)
\]
\[
\begin{aligned}
\mathcal{V} \;& \text{: The vocabulary, i.e., the set of all possible tokens known to the model.} \\
t \;& \text{: Token position in the input sequence.} \\
x_{<t} \;& \text{: The sequence of tokens preceding position $t$ (the conditioning context).} \\
w \;& \text{: A token belonging to the vocabulary $\mathcal{V}$.} \\
P_{\text{CPT}}(w \mid x_{<t}) \;& \text{: Probability assigned by the CPT model to token $w$ given context $x_{<t}$.} \\
P_{\text{Base}}(w \mid x_{<t}) \;& \text{: Probability assigned by the base model to token $w$ given the same context $x_{<t}$.} \\
D_{\mathrm{KL}}(\cdot \| \cdot) \;& \text{: Kullback--Leibler divergence measuring difference between two distributions.}
\end{aligned}
\]

These token-level KL values are then averaged across all valid tokens and across the entire evaluation corpus using the below formula to obtain a single measure of distributional shift between the models.
\[
\mathrm{KL}_{\text{avg}}
=
\mathbb{E}_{x \sim \mathcal{D}}
\left[
\frac{1}{T_x}
\sum_{t=1}^{T_x}
D_{\mathrm{KL}}
\Big(
P_{\text{CPT}}(\cdot \mid x_{<t})
\;\|\;
P_{\text{Base}}(\cdot \mid x_{<t})
\Big)
\right]
\]
\[
\begin{aligned}
\mathcal{D} \;& \text{: The evaluation corpus (dataset).} \\
x \;& \text{: A tokenized input sequence sampled from $\mathcal{D}$.} \\
T_x \;& \text{: The number of tokens in sequence $x$.} \\
\mathbb{E}[\cdot] \;& \text{: Expectation over all sequences in the evaluation corpus.}
\end{aligned}
\]
\noindent
The averaged KL divergence quantifies how much the CPT model's full next-token probability distribution deviates from the base model's distribution, averaged across all tokens and all evaluation samples.

        \vspace{0.5em}
        
\textbf{Perplexity} 
        
Perplexity, originally introduced in the context of automatic speech recognition \citep{jelinek1977perplexity, brown1992class}, quantifies a language model’s uncertainty in predicting the next token in a sequence. Formally, it is defined as the exponential of cross-entropy between the model’s predicted distribution and the empirical distribution of the data. Intuitively, perplexity represents the effective number of choices the model considers on average at each prediction step, where lower values indicate more confident predictions and higher values indicate greater uncertainty. Consequently, perplexity is a fundamental metric for evaluating the fluency and coherence of probabilistic language models, given a sequence of tokens \( X = \{x_1, x_2, \dots, x_N\} \), perplexity is calculated as

\[
\mathrm{PPL}(X)
=
\exp\left(
-\frac{1}{N}
\sum_{i=1}^{N}
\log P\bigl(x_i \mid x_{<i}\bigr)
\right),
\]

where \( P(x_i \mid x_{<i}) \) denotes the conditional probability assigned by the model to token \( x_i \) given all preceding tokens.

\vspace{0.5em}

\subsection{Training}
\label{sec:continuous_pre-training}

So far, we have prepared all the required datasets, including dedicated evaluation sets to assess both general and domain specific knowledge. With this in place, we are now ready to proceed to the actual training phase.
Rather than using a standard pre-training approach, we adopted an annealing-based curriculum. This strategy systematically modulates key training parameters most notably the learning rate across three controlled phases. By doing so, the model is able to internalize financial knowledge while preserving its general reasoning abilities, effectively avoiding catastrophic forgetting. Our training runs leveraging DeepSpeed ZeRO-3 \citep{rajbhandari2020zeromemoryoptimizationstraining} and FlashAttention-2 \citep{dao2023flashattention2fasterattentionbetter} to optimize memory usage and maximize speed.

\begin{table}[htbp]
\centering
\begin{tabular}{ll}
\toprule
\textbf{Configuration} & \textbf{Value} \\
\midrule
Compute Power & 8 Nodes (64 H100 GPUs) \\
Precision & bfloat16 \\
Context Window & 8192 Tokens \\
Peak Efficiency (MFU) & $\sim$40\% \\
\bottomrule
\end{tabular}
\caption{Infrastructure and System Configuration. Summary of the hardware environment and optimization parameters, including compute power and token context, utilized during the CPT phase.}
\label{tab:infra_config}
\end{table}



\paragraph{Dataset Format}

The training dataset is stored in a line-delimited JSON format, where each line contains a single raw text sample under the text field:

\begin{verbatim}
{"text": "<raw text content>"}
{"text": "<raw text content>"}
...
\end{verbatim}

No prompt template or special formatting is applied. Introducing such a structure would bias the model toward learning formatting patterns rather than the underlying financial context and domain-specific knowledge.

\begin{figure}[htbp]
    \centering
    \includegraphics[width=1.0\linewidth]{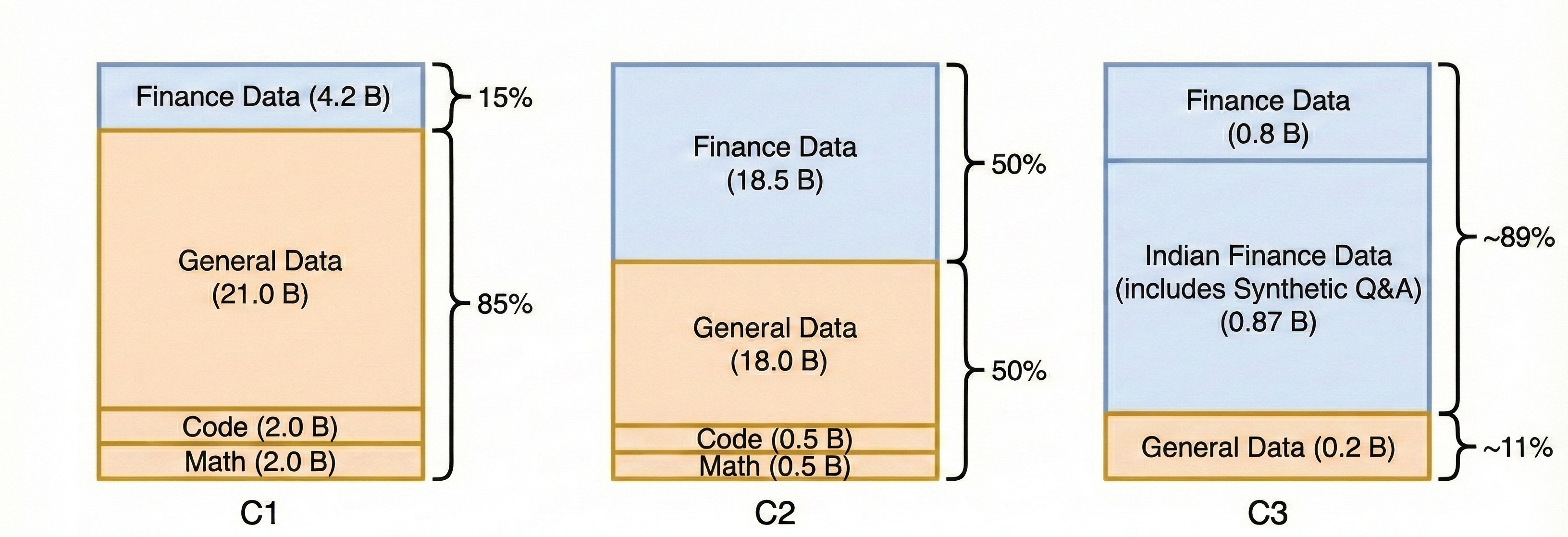}
    \caption{Phase-wise token distribution across training stages, illustrating the allocation of data volume at each phase of the CPT curriculum.}
    \label{fig:data_distribution_phases}
\end{figure}

\paragraph{Data Distribution}
Detailed breakdown of the $\sim$68 Billion tokens utilized across the curriculum stages, highlighting the shifts in data composition from foundational grounding to domain-specific annealing. The evolution of the data mixture across the three-stage curriculum, detailing the relative proportions of general and finance-specific tokens, is illustrated in \hyperref[fig:data_distribution_phases] {Figure~\ref*{fig:data_distribution_phases}}.
\paragraph{Implementation Mechanics}

To efficiently ingest Billions of tokens from multiple dataset blends without changing the intended token distribution, the following mechanisms were implemented:

\begin{itemize}   \setlength{\itemsep}{0pt}   \setlength{\parskip}{0pt}   \setlength{\parsep}{0pt}
  \setlength{\itemsep}{0pt}
  \setlength{\parskip}{0pt}
  \setlength{\parsep}{0pt}
    \item \textbf{Data Streaming}: Continuous data streaming is used instead of loading full datasets into memory. Samples from different dataset blends are interleaved according to predefined sampling ratios, ensuring memory efficiency while strictly preserving the intended data-mix proportions throughout training.
    
    \item \textbf{Sequence Packing}: To maximize utilization of the model’s context window, sequence packing is applied at the trainer level. Packing is performed in a controlled manner such that only samples belonging to the same dataset type are concatenated together. This avoids cross-domain mixing (e.g., math with code or text) within a single packed sequence, preserving domain coherence during training. 
\end{itemize}

\paragraph{Training Recipe}
\label{sec:training_recipe}

To achieve robust adaptation to the financial domain, we design a training procedure that systematically adjusts learning rates, batch sizes, and optimization parameters, as summarized in \hyperref[tab:cpt_training_config] {Table~\ref*{tab:cpt_training_config}}.



\begin{table}[htbp]
\centering
\begin{tabular}{>{\raggedright\arraybackslash}p{4cm} >{\raggedright\arraybackslash}p{3cm} >{\raggedright\arraybackslash}p{3cm} >{\raggedright\arraybackslash}p{3cm}}
\toprule
\textbf{Parameter} & \textbf{Phase C1 } & \textbf{Phase C2 } & \textbf{Phase C3 } \\ \midrule
Learning Rate (LR)            & $3\times10^{-6}$ & $1\times10^{-5}$ & $1\times10^{-5}$ \\ \midrule
Duration                      & 3 days           & 6.66 days        & 5 hours          \\ \midrule
Batch Size (per device)       & 4                & 8                & 8                \\ \midrule
Gradient Accumulation Steps   & 8                & 16               & 16               \\ \midrule
Training Steps                & 8{,}000          & 12{,}900         & 250              \\ \midrule
LR Scheduler                  & Cosine           & Cosine           & Cosine           \\ \midrule
Warmup Steps                  & 100              & 2{,}000          & 50               \\ \midrule
Optimizer                     & AdamW (fused)    & AdamW (fused)    & AdamW (fused)    \\ \midrule
Weight Decay                  & 0.03             & 0.03             & 0.01             \\ \midrule
Adam Betas                    & [0.9, 0.999]     & [0.9, 0.999]     & [0.9, 0.999]     \\ \midrule
Adam Epsilon                  & $1\times10^{-8}$ & $1\times10^{-8}$ & $1\times10^{-8}$ \\ \bottomrule
\end{tabular}
\caption{Multi-Phase CPT using a structured three-stage training curriculum for progressive domain adaptation and specialization.}
\label{tab:cpt_training_config}
\end{table}

\subsubsection{Phase Breakdown}
\label{sec:phase_breakdown}

\paragraph{Phase C1: The Initial Phase}
\label{sec:phase_c1}

In this phase, we introduce financial data in a controlled manner, ensuring a smooth shift in token distribution while preserving general reasoning ability and training stability. 

\begin{figure}[htbp]
    \centering
    \includegraphics[width=1\linewidth]{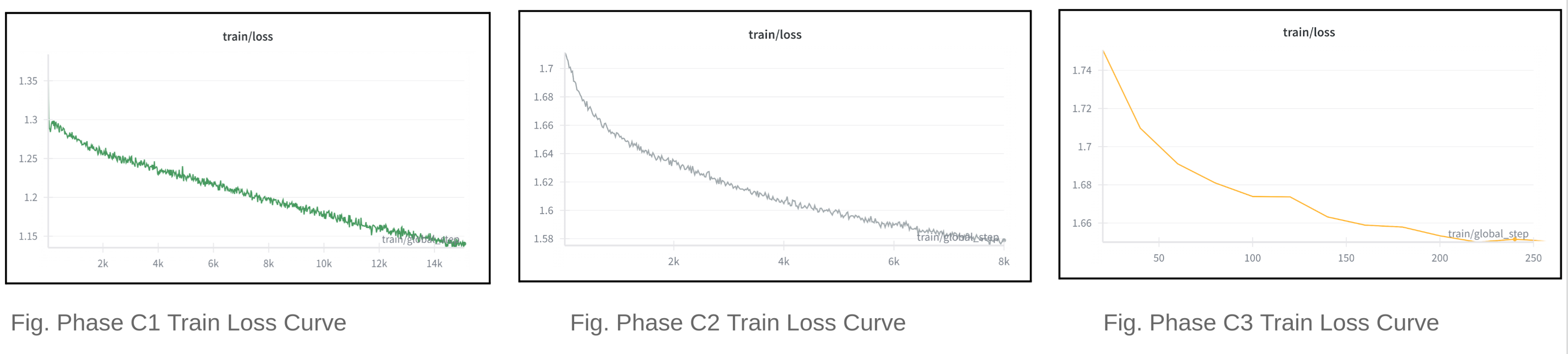}
    \caption{
    Phase-wise training loss progression:
    (a) Phase C1, showing initial optimization under the base data distribution;
    (b) Phase C2, illustrating adaptation to the intermediate shifted distribution;
    (c) Phase C3, highlighting stabilization and convergence toward the FiMI Base Model foundational state.
    }
    \label{fig:all_phase_loss}
\end{figure}

\paragraph{Observations}
\begin{itemize}   \setlength{\itemsep}{0pt}   \setlength{\parskip}{0pt}   \setlength{\parsep}{0pt}
    \item The loss curve (\hyperref[fig:all_phase_loss] {Figure~\ref*{fig:all_phase_loss}a}) indicates a well-stabilized initial CPT phase, in which the model consistently refines domain-specific representations under a steady data mix and learning rate. The smooth, nearly linear downward trend with low noise reflects healthy optimization, while the absence of sharp drops suggests minimal changes in the dataset composition due to the dominance of high-volume general data.
    \item We observed a decline across all benchmarks, with only marginal improvement in domain-specific evaluation metrics.
\end{itemize}
Overall, these results are expected at this stage of CPT, as the model begins internalizing financial-domain patterns but has not yet received sufficiently strong domain signals to translate into measurable task-level improvements.

\paragraph{Phase C2: The Domain Shift}
\label{sec:phase_c2}


In this phase, we build a strong domain understanding while connecting global and Indian financial concepts. For this we used an equal mix of general and financial data.
\paragraph{Observations.}
\begin{itemize}   \setlength{\itemsep}{0pt}   \setlength{\parskip}{0pt}   \setlength{\parsep}{0pt}
    \item The loss curve (\hyperref[fig:all_phase_loss] {Figure~\ref*{fig:all_phase_loss}b}) shows a stable CPT phase with a consistent data mix and learning-rate schedule, where the model steadily refines existing representations, resulting in a smooth and gradual decline in training loss. Minor fluctuations are expected and can be attributed to minibatch noise.

    \item The general benchmarks partially stabilized, and domain evaluation scores improved due to the significant shift in the model’s contextual understanding. 
\end{itemize}

\noindent Phase C2 serves as a controlled transition, where a balanced data mix induces a deliberate contextual realignment. The 8K-step checkpoint is selected as the starting point for domain consolidation in the next phase.

\paragraph{Phase C3: The Consolidation Stage}
\label{sec:phase_c3}

In this phase, we teach the model India specific financial concepts and terminology, such as regulatory terms, instruments and taxation constructs. By grounding the model in these local nuances, it improves its ability to accurately understand and respond to direct, factual, and reasoning-based questions within the Indian financial domain.

\paragraph{Observations.}
\begin{itemize}   \setlength{\itemsep}{0pt}   \setlength{\parskip}{0pt}   \setlength{\parsep}{0pt}
    \item The step wise training loss pattern (\hyperref[fig:all_phase_loss] {Figure~\ref*{fig:all_phase_loss}c}) reflects phase-wise training behavior. As the data distribution is rebalanced, the model briefly adapts to conflicting patterns before converging to a stable optimization regime.

    \item Domain benchmark scores increased to approximately 20\%, while some regression was observed on general benchmarks. 
\end{itemize}


Adding synthetic question answer pairs along with a portion of the original pre-training data led to clear improvements in domain specific performance. This yielded the final CPT C3 checkpoint. The optimization stability and convergence behavior throughout this transition are visualized in the multi-stage training loss curves in \hyperref[fig:all_phase_loss] {Figure~\ref*{fig:all_phase_loss}}.

The final C3 checkpoint produced through the CPT becomes the foundational model, referred to as the FiMI Base Model, and serves as the starting point for all subsequent fine-tuning and post-training stages.

\subsection{Benchmarking and Results}
\label{sec:results}

As we explored different approaches to model evaluation, we aligned the observed results with the original objectives defined for CPT. It resulted in meaningful domain specialization, yielding an approximately 20\% improvement on domain-specific benchmarks and consistent gains across other financial benchmarks, with variance observed in general benchmark performance. The specific performance gains across specialized finance reasoning tasks and a comparative analysis of general benchmarks are detailed in \hyperref[tab:domain_gains]{Table~\ref*{tab:domain_gains}}.

We further use two evaluation metrics, KL Divergence and Perplexity, as discussed in the evaluation section, to measure the CPT model’s understanding of the Indian financial context.

\noindent Perplexity measures how uncertain or surprised a model is when predicting the next token in a sequence. Lower perplexity indicates that the model better understands the data and can predict tokens more confidently.

\noindent Similarly, KL divergence is calculated to assess how much the CPT model has drifted from the base model by comparing their token probability distributions side by side. A low KL divergence value indicates that the CPT model remains very close to the base model with little to no improvement, whereas a very high value (e.g., $\geq 1$) may indicate overfitting to the training dataset and catastrophic forgetting of base knowledge. A summary of these findings is presented in \hyperref[tab:perplexity_analysis]{Table~\ref*{tab:perplexity_analysis}} .

\vspace{0.8em}
\textbf{Analysis:}


\begin{table}[htbp]
    \centering
    \begin{adjustbox}{max width=\textwidth}
        \begin{tabular}{l c c c c p{4.5cm}}
            \toprule
            \textbf{Category} & \textbf{Init. PPL} & \textbf{Final PPL} & \textbf{Change} & \textbf{KL Div} & \textbf{Outcome} \\
            \midrule
            \textbf{Domain Alignment} & 13.11 & 5.40 & 58\% Reduction & 0.75 & 58\% improvement in domain understanding and prediction confidence \\
            \addlinespace 
            \textbf{General Reasoning} & 6.77 & 5.71 & Slight Imp. & 0.33 & Abilities preserved (no catastrophic forgetting) \\
            \bottomrule
        \end{tabular}
    \end{adjustbox}
    \caption{Perplexity and KL Divergence Analysis. This table compares the model's performance on the Indian Finance Dataset versus the General Dataset, highlighting improvements in domain alignment and the preservation of general reasoning capabilities.}
    \label{tab:perplexity_analysis}
\end{table}

\begin{table}[htbp]
\centering
\begin{adjustbox}{max width=\textwidth}
\begin{tabular}{p{3.5cm}cc}
\toprule
\textbf{Benchmark} & \textbf{FiMi Base} & \begin{tabular}{@{}c@{}}\textbf{Mistral 2501} \\ \textbf{Base}\end{tabular} \\
\midrule
arc & 56 & \textbf{61} \\
gsm8k (5 shot) & 80 & \textbf{81} \\
humaneval\_inst & 35 & \textbf{66} \\
mmlu & 72 & \textbf{77} \\
mmlu\_pro (5 shot) & 49 & \textbf{52} \\
triviaqa & 52 & \textbf{67} \\
\midrule
Finance Reasoning & \textbf{65} & 54 \\
MMLU Finance & \textbf{50} & 40 \\
HellaSwag Finance & \textbf{77} & 69 \\
\bottomrule
\end{tabular}
\end{adjustbox}
\caption{Benchmark Comparison: Mistral 2501 Base vs FiMi Base. Finance reasoning improves with FiMi Base while general benchmark performance is generally lower compared to Mistral 2501 Base.}
\label{tab:domain_gains}
\end{table}

\noindent Overall, CPT helps align the model with the Indian financial domain by increasing its exposure to domain-specific datasets. This alignment ensures a more grounded understanding of regulatory and operational contexts and establishes a strong foundation for subsequent post-training stages, enabling more effective and focused fine-tuning.

\section{Post-Training Stage}

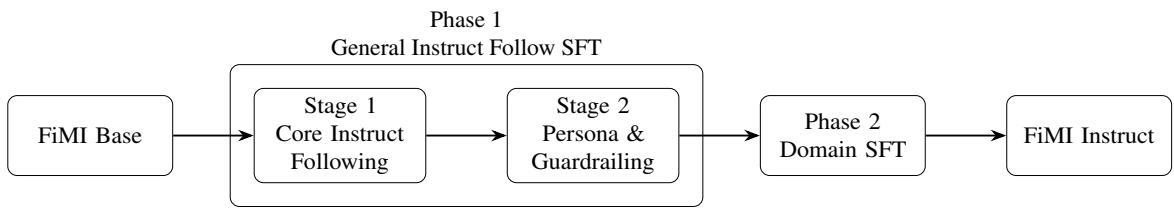
\begin{figure}[htbp]
    \centering
    \resizebox{1.0\textwidth}{!}{%
        \usetikzlibrary{positioning,fit,arrows.meta}

\begin{tikzpicture}[
    box/.style={
        draw,
        rounded corners,
        align=center,
        minimum height=1.2cm,
        text width=2.2cm,
        font=\small
    },
    stage/.style={
        draw,
        rounded corners,
        align=center,
        minimum height=1.1cm,
        text width=2.3cm,
        font=\small
    },
    arrow/.style={
        -{Stealth},
        thick
    },
    node distance=1.2cm 
]

\node[box] (base) {FiMI Base};

\node[stage, right=of base] (stage1)
{Stage 1\\Core Instruct\\Following};

\node[stage, right=of stage1] (stage2)
{Stage 2\\Persona \&\\Guardrailing};

\node[box, right=of stage2] (phase2)
{Phase 2\\Domain SFT};

\node[box, right=of phase2] (final)
{FiMI Instruct};

\node[
    draw,
    rounded corners,
    fit=(stage1) (stage2),
    inner sep=0.35cm,
    label={[font=\small, align=center]above:
        Phase 1\\General Instruct Follow SFT}
] (phase1) {};

\draw[arrow] (base) -- (stage1);
\draw[arrow] (stage1) -- (stage2);
\draw[arrow] (stage2) -- (phase2);
\draw[arrow] (phase2) -- (final);

\end{tikzpicture}
    }
    \caption{Post-training pipeline for FiMI Instruct. The pipeline begins with general instruction-following supervised fine-tuning (SFT) to align the FiMI Base model with user intent and conversational constraints, followed by domain-specific SFT on curated NPCI workflows to enable structured tool usage, compliance-aware reasoning, and context-sensitive responses.}

    \label{fig:post_training_pipeline}
\end{figure}

The critical next step for the FiMI Base Model is to bridge the gap between raw linguistic modeling and functional utility. While pre-training provides deep domain knowledge, the base model naturally exhibits limited alignment with user intent, often continuing a prompt’s textual flow rather than executing its command. The overall post-training pipeline, highlighting the progression from general instruction-following to domain-aligned agentic behavior, is illustrated in \hyperref[fig:post_training_pipeline]{Figure~\ref*{fig:post_training_pipeline}}. As the foundation of the post-training stage, we implement a two-phase post-training strategy:

\begin{itemize}   \setlength{\itemsep}{0pt}   \setlength{\parskip}{0pt}   \setlength{\parsep}{0pt}
    \item \textbf{Phase 1: General Instruction-Following SFT.}  
    This initial phase transforms the model from a next-token predictor into an instruction-following assistant. By training on high-quality and diverse prompt--response pairs, the model learns to identify user intent, adhere to conversational constraints, and execute generalized tasks.

    \item \textbf{Phase 2: Domain-Specific SFT.} 
    Building on the general instruction-following baseline, this phase fine-tunes the model exclusively on curated, domain-specific datasets. This process establishes domain-aligned agentic capabilities, ensuring the generation of compliant, accurate, and context-aware responses required UPI Help.
\end{itemize}



\subsection{Phase 1: General Instruction-Following SFT}

Following the Continued Pre-training (CPT) phase, the FiMI Base model possesses enhanced domain representations but lacks the structural coherence required for interactive tasks. The primary objective of Phase 1 is to bridge this gap by transforming the model from a probabilistic next-token predictor into a robust instruction-following agent. This phase focuses on instilling a broad spectrum of competencies-ranging from mathematical reasoning and code generation to safe conversational protocols—ensuring a stable foundation for subsequent domain-specific alignment.

We structure this phase into two distinct stages to separate the learning of core capabilities from behavioral alignment. This design allows the model to first acquire general instruction-following competence and then refine its responses through persona and safety alignment. By decoupling these objectives, each stage can be optimized without interfering with the other.
\begin{itemize}   \setlength{\itemsep}{0pt}   \setlength{\parskip}{0pt}   \setlength{\parsep}{0pt}
    \item Stage I utilizes a massive, token-rich dataset to establish deep reasoning and foundational skills (math, coding, tool use).
    \item Stage II employs a smaller, highly curated mixture to refine generalizability, safety, and adherence to complex system prompts.
\end{itemize}

\paragraph{Stage I: Core Instruction Following}
The FiMI Base model, developed during CPT phase, underwent further training on Smoltalk2 \citep{allal2025smollm2smolgoesbig}, a substantial 19 Billion token dataset designed to bridge the gap between rigorous reasoning and conversational fluency. We selected Smoltalk2 after evaluating several high-volume instruction-following corpora against three critical criteria: CoT density, agentic readiness, and token scale. Our evaluation revealed that while standard conversational datasets offer fluency, they often lack explicit step-by-step reasoning and executable logic patterns. Smoltalk2 addresses this by prioritizing complex, multi-step problem solving prior to stylistic refinement, making it an ideal reasoning backbone for Phase I.

The specific characteristics of Smoltalk2 that motivated its selection include:
\begin{itemize}   \setlength{\itemsep}{0pt}   \setlength{\parskip}{0pt}   \setlength{\parsep}{0pt}
    \item \textbf{Reasoning-Dominant Composition:} Approximately 16 Billion tokens (84\%) are derived from OpenThoughts3. This high density of CoT data anchors the model’s ability to perform complex logical reasoning and multi-step problem solving.

    \item \textbf{Agentic Capabilities:} The dataset provides dual-mode function-calling training, covering both XML-based (traditional) and Python-based (executable) traces. This ensures the model is tool-ready for a wide range of agentic workflows.

    \item \textbf{Conversational Entropy:} To mitigate the rigidity often associated with reasoning-focused models, the dataset integrates a diverse set of everyday prompts. This promotes a balanced training signal, allowing the model to retain social and conversational nuance while improving performance on tasks such as rewriting, summarization, and structured table understanding.
\end{itemize}

\paragraph{Stage II: Generalization and Alignment}
Building upon the Stage I baseline, the model is trained on the tulu-3-sft-mixture \citep{lambert2024tulu3}. This 435M token dataset was selected after evaluating alternatives such as raw user logs and generic safety corpora, which in internal testing exhibited a common failure mode instruction drift where models degrade in persona consistency and constraint adherence over long-context interactions. This dataset directly addresses these alignment gaps by refining the model’s interaction patterns, safety behavior, and robustness without impacting the reasoning capabilities acquired in Phase I.
The following components of the tulu-3-sft-mixture contribute to alignment, robustness, and instruction fidelity in Stage II:
\begin{itemize}   \setlength{\itemsep}{0pt}   \setlength{\parskip}{0pt}   \setlength{\parsep}{0pt}
    \item \textbf{Real-World Interactions:} Integration of WildChat \citep{zhao2024wildchat1mchatgptinteraction} data exposes the model to diverse, real-world user interaction patterns, ensuring that the model remains helpful and unbiased beyond synthetic benchmark scenarios.

    \item \textbf{Persona-Driven Instruction Following:} The inclusion of Persona Math, Persona Algebra, and Numina Math datasets \citep{numina_math_datasets} enforces strict adherence to predefined personas (for example, ``Act as a Math Tutor''), thereby improving constrained multi-step reasoning and instruction compliance.

    \item \textbf{Safety and Compliance:} WildGuardMix \citep{wildguard2024} and WildJailbreak \citep{wildteaming2024} datasets are used to aggressively reduce bias and prevent harmful outputs, while the No Robots \citep{no_robots} dataset reinforces general compliance-oriented behavior.

    \item \textbf{System Prompt Adherence:} A central objective of this stage is minimizing instruction drift, ensuring that the model consistently respects the system prompt regardless of conversation length or complexity.

    \item \textbf{Model Identity:} Instead of relying on hardcoded responses for identity-related queries, custom self-identity data was injected during this phase to enable dynamic self-awareness. This methodology is described in further detail in the Synthetic Data Generation (SDG) section.
\end{itemize}

\subsubsection{Data Distribution}
\label{sec:phase1_data_distribution}

We filtered the datasets from both stages specifically for Indic (English and Hindi) languages. The final Phase~1 corpus comprises approximately 18.2 Billion tokens across 2.9 Million examples. The data distribution is deliberately non-uniform, heavily favouring Mathematics and Coding (combined $\sim$88\%). This skew is engineered to support the model’s downstream application as a ReAct (Reason + Act) Agent in UPI Help, which is discussed later in \hyperref[sec:agentic_upi_help]{Annexure B}.

\hyperref[tab:phase1_data_distribution]{Table~\ref*{tab:phase1_data_distribution}} summarizes the category wise composition of the Phase~1 instruction-tuning corpus in terms of examples and token distribution.

\begin{table}[htbp]
    \centering
    \begin{tabular}{lrrr}
        \toprule
        \textbf{Category} & \textbf{Examples} & \textbf{Tokens (M)} & \textbf{Tokens (\%)} \\
        \midrule
        Math & 1,445,098 & 12,390.00 & 68.03 \\
        Coding & 177,331 & 3,719.79 & 20.42 \\
        Science & 221,607 & 1,417.51 & 7.78 \\
        Long Context & 13,775 & 231.99 & 1.27 \\
        Instruction Following & 250,565 & 201.97 & 1.11 \\
        Tool Calling & 477,175 & 104.59 & 0.57 \\
        Rewriting \& Summarization & 179,714 & 76.79 & 0.42 \\
        Tabular Data & 31,404 & 41.26 & 0.23 \\
        Guardrails, Safety \& Compliance & 109,500 & 24.80 & 0.14 \\
        Everyday Conversations & 4,317 & 3.80 & 0.02 \\
        \midrule
        \textbf{Total} & \textbf{2,910,486} & \textbf{18,212.50} & \textbf{100.00} \\
        \bottomrule
    \end{tabular}
    \caption{Phase 1 Data Distribution. Category-wise distribution of the Phase~1 instruction-tuning dataset after Indic-language filtering (English and Hindi). The table reports the number of examples and total token counts per category, along with their relative proportions.}
    \label{tab:phase1_data_distribution}
\end{table}
\subsubsection{Chat Template}
\label{sec:phase1_chat_template}

We identified that the baseline Mistral Small 24B chat template was overly restrictive, supporting only standard \texttt{User} and \texttt{Assistant} roles. To enable function calling capabilities, we designed a custom template that explicitly structures tool interactions. This schema activates the \texttt{[AVAILABLE\_TOOLS]}, \texttt{[TOOL\_CALLS]}, and \texttt{[TOOL\_RESULTS]} special tokens, which were natively supported by the baseline tokenizer but unutilized in the default configuration.

\subsubsection{Training Setup}
\label{sec:phase1_training_setup}

\paragraph{Computational Resources and Efficiency}
The Phase 1 pipeline was executed on a cluster of NVIDIA H100 GPUs, utilizing bfloat16 precision to ensure numerical stability and optimal dynamic range. To efficiently scale across the cluster, we employed DeepSpeed ZeRO-3 to partition optimizer states, gradients, and parameters across the cluster, while utilizing activation checkpointing to minimize memory footprint. Compute throughput was maximized by leveraging FlashAttention-2 and sequence packing, further accelerated with torch.compile. This configuration resulted in a peak MFU of $\sim$40.7\%. The AdamW optimizer was utilized consistently across both stages.

\paragraph{Hyperparameter Tuning}
We conducted a learning-rate sweep across the range
$[5 \times 10^{-6},\, 5 \times 10^{-5}]$.
Based on loss convergence patterns, optimal rates were selected for each
stage. An identical cosine decay scheduler was used for both stages, with
a linear warmup phase comprising approximately 5\% of the total steps to
ensure stable convergence. The complete training configuration for both
stages of Phase~1 is summarized in
\hyperref[tab:phase1_training_config] {Table~\ref*{tab:phase1_training_config}}.

\begin{table}[htbp]
    \centering
    \begin{tabular}{lcc}
        \toprule
        \textbf{Configuration} & \textbf{Stage I} & \textbf{Stage II} \\
        \midrule
        Compute Topology & 8 Nodes (64 GPUs) & 4 Nodes (32 GPUs) \\
        Duration & 1d 9h & $\sim$3h \\
        Global Batch Size & 256 & 256 \\
        Batch Size (Per Device) & 2 & 4 \\
        Gradient Accumulation & 2 & 2 \\
        Max Sequence Length & 16384 & 8192 \\
        Peak Learning Rate & $5.00 \times 10^{-6}$ & $1.00 \times 10^{-5}$ \\
        LR Scheduler & Cosine & Cosine \\
        Warmup Steps & 209 & 11 \\
        Weight Decay & 0.05 & 0.05 \\
        Optimizer & AdamW (Fused) & AdamW (Fused) \\
        \bottomrule
    \end{tabular}
    \caption{Phase 1 Training Configuration. Training configuration for Phase~1 post-training, detailing compute topology, optimization settings, batch sizing, sequence length, and learning-rate schedules for both Stage~I and Stage~II. Differences in scale and duration reflect the transition from broad instruction alignment to targeted generalization and alignment refinement.}
    \label{tab:phase1_training_config}
\end{table}

\subsubsection{Benchmark}
\label{sec:phase1_benchmarks}

To validate the efficacy of Phase 1, we tracked model progression across four critical checkpoints: the external baseline (Mistral~2501~Base), our internal pre-trained checkpoint, the Stage~I output, and the Stage~II output. The evaluation focuses on three core competencies: reasoning ability (math and logic), coding proficiency, and instruction following. \hyperref[tab:phase1_benchmark_results] {Table~\ref*{tab:phase1_benchmark_results}} summarizes the performance of the
baseline models and Phase~1 checkpoints across a diverse set of
evaluation benchmarks.

\begin{table}[htbp]
\centering
\begin{tabular}{lcccc}
\hline
\textbf{Benchmark} & \textbf{Mistral Base} & \textbf{FiMI Base} & \textbf{Stage I} & \textbf{Stage II} \\
\hline
GSM8K (Math) & 0.81 & 0.80 & \textbf{0.88} & 0.87 \\
\hline
HumanEval (Coding) & 0.66 & 0.35 & 0.72 & \textbf{0.77} \\
\hline
IFEval (Instruction) & 0.29 & 0.26 & 0.78 & \textbf{0.79} \\
\hline
MMLU (General) & \textbf{0.77} & 0.72 & 0.73 & 0.74 \\
\hline
MMLU-Pro (Reasoning) & 0.53 & 0.49 & \textbf{0.58} & 0.51 \\
\hline
TriviaQA (Knowledge) & \textbf{0.68} & 0.52 & 0.23 & 0.45 \\
\hline
\end{tabular}
\caption{Phase 1 Benchmark Results. Comparative analysis of baselines Mistral Small 24B Base, FiMI Base and Phase 1 checkpoints (Stage I, Stage II), demonstrating progression in reasoning, coding, and instruction following benchmarks.}
\label{tab:phase1_benchmark_results}
\end{table}

\paragraph{Analysis of Stage I Progression}
Initialization on Smoltalk2 anchored the model’s core logical abilities, confirming that large-scale CoT exposure is a primary driver of reasoning performance. During this phase, we observed significant improvements in mathematical reasoning, with GSM8K performance rising from 0.80 to 0.88, alongside a peak MMLU-Pro \citep{wang2024mmluprorobustchallengingmultitask} score of 0.58. While initial pre-training caused a regression in coding capabilities (HumanEval \citep{chen2021evaluatinglargelanguagemodels} dropping to 0.35), Stage~I successfully reversed this trend. By leveraging Python trace data to reactivate latent coding knowledge, the model exceeded baseline performance, reaching 0.72. However, this specialization required a deliberate trade-off in world knowledge; TriviaQA \citep{joshi2017triviaqalargescaledistantly} scores declined sharply (0.68 $\rightarrow$ 0.23), a predictable outcome given a data distribution that heavily prioritized reasoning patterns (88\%) over general knowledge (0.1\%).

\paragraph{Analysis of Stage II}
The most significant improvement occurred in IFEval \citep{zhou2023instructionfollowingevaluationlargelanguage}, which increased from 0.29 (Base) to 0.79. This demonstrates that while Stage~I established raw reasoning capability, Stage~II was essential for teaching adherence to formatting constraints and system prompts. HumanEval performance continued to improve during Stage~II (0.72 $\rightarrow$ 0.77), suggesting that the diverse interaction patterns in Tulu-3 helped the model generalize its coding abilities across a wider range of prompt structures. A modest regression was observed in pure reasoning metrics, with MMLU-Pro decreasing from 0.58 to 0.51. This reflects a known trade-off in which stricter safety and formatting constraints slightly reduce creative reasoning variance. With general instruction-following capabilities now stabilized (IFEval at 0.79), the model is well prepared to advance to Phase~2.

\subsection{Phase 2: Domain Supervised Fine-Tuning}
\label{sec:phase2_domain_sft}

Phase~2 builds on the foundation model obtained through Continuous
Pre-Training (FiMI Base) by further aligning it using Instruction
Fine-Tuning (IFT). This two-stage process equips the model with a deeper
understanding of Indian financial terminology, payment systems, and
regulatory context, while preserving its general reasoning and language
capabilities. Evaluation results summarized in \hyperref[tab:phase1_benchmark_results] {Table~\ref*{tab:phase1_benchmark_results}} confirm that the model remains
stable and reliable after Phase~1, yielding a strong and reusable
checkpoint. This checkpoint serves as a base financial model that can be
extended along multiple trajectories depending on ecosystem
requirements.

Phase~2 focuses on identifying these potential trajectories and
selecting the most suitable path for real-world deployment.

\subsubsection{Potential Fine-Tuning Trajectories}
\label{sec:paths_after_phase1}

After Phase~1, the resulting checkpoint can evolve along the following paths:

\begin{itemize}   \setlength{\itemsep}{0pt}   \setlength{\parskip}{0pt}   \setlength{\parsep}{0pt}
    \item \textbf{General Financial Assistant:} Further instruction tuning to enhance broad financial question answering, summarization, and multilingual conversational capabilities.

    \item \textbf{Institution-Specific Adaptation:} Banks and payment service providers (PSPs) can fine-tune the checkpoint for institution-specific use cases such as customer support, compliance workflows, fraud analysis, and internal knowledge systems.

    \item \textbf{Domain Specialisation:} Domain-focused fine-tuning to align the model with specific financial products and operational workflows, emphasizing structured reasoning and deterministic behavior.
\end{itemize}

For Phase 2, we selected domain-specific specialisation with a focus on UPI Help due to the nature of real UPI support workflows. UPI Help enables users to check transaction status, resolve failed or pending payments, raise and track complaints, and manage mandates. \noindent These flows require complex tool calling, step-by-step resolution, and accurate system interactions, which cannot be achieved through general-purpose tuning. Domain-specific supervised fine-tuning allows the model to learn UPI rules, error scenarios, dispute categories, and resolution logic in a controlled and compliant manner.

\noindent While Phase 2 targets UPI Help, the Phase 1 checkpoint remains reusable for other NPCI products and institution-specific use cases across the ecosystem.

\begin{figure}[htbp]
    \centering
    \includegraphics[width=1.0\linewidth]{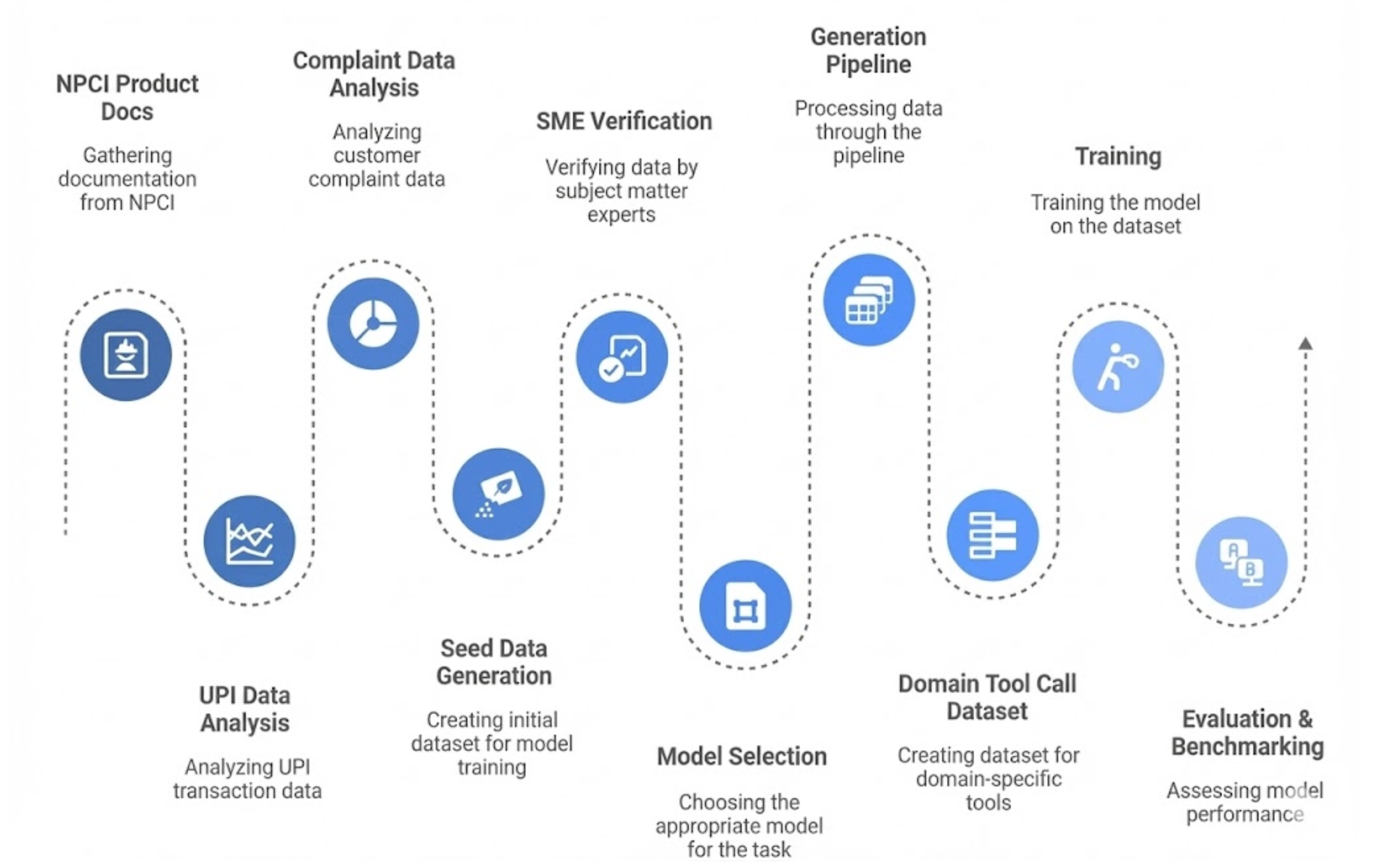}
    \caption{Synthetic Data Generation Flow. Technical flow diagram showing the sequential stages from data sourcing and SME verification to model training and benchmarking.}
    \label{fig:synthetic_data_flow}
\end{figure}

\subsection{Synthetic Data Preparation and Curation}
\label{sec:phase2_data}

The UPI Help use case requires training data that reflects structured, multi-turn support workflows with precise tool invocation. Existing conversational datasets do not adequately capture UPI-specific transaction semantics and operational constraints, motivating a domain-specific data preparation and curation approach.

These use cases are domain-specific, involving real-time transaction checks, error descriptions, and tool-driven action patterns not found in generic datasets. Public data lacks the financial context, structured interactions, and rule-bound flows required to train the system. The inclusion of user queries from social media channels in the training corpus is not feasible due to inherent and non mitigable risks associated with PII exposure and regulatory non-compliance. Despite the application of advanced anonymization techniques, the residual risk of PII reidentification and unintended data leakage into the model remains high, thereby necessitating the exclusion of such data sources from training.

These gaps necessitate finance-specific synthetic data to accurately capture UPI Help’s semantics, workflows, and compliance needs. This requirement defines the scope of the product document and the boundaries of the data pipeline. The complete workflow for generating this synthetic data is illustrated in \hyperref[fig:synthetic_data_flow]{Figure~\ref*{fig:synthetic_data_flow}}\textbf{}.

\noindent To ensure that the synthetic data faithfully reflects real user behavior while avoiding the risks of direct data usage, we first conducted an analysis of real world interaction patterns. This analysis serves as an observational step to understand how users naturally articulate problems, express intent, and frame transactional grievances, informing the design of realistic and representative synthetic scenarios.

\subsubsection{Social Media Query Analysis}
\label{sec:social_media_analysis}

Before proceeding with synthetic data generation, we first obtained information on actual user queries derived from  interactions. The analysis of real user queries reveals distinct structural, behavioral, and linguistic characteristics that collectively define how users express issues within the UPI support ecosystem:

\begin{enumerate}[label=\alph*)]
  \setlength{\itemsep}{0pt}
  \setlength{\parskip}{0pt}
  \setlength{\parsep}{0pt}
    \item Query Structure \& Emotional Tone: User queries are typically short and abrupt, often fragmented in structure, and frequently convey strong emotions such as frustration or urgency. \\
    Example: \textit{“Payment failed again!! Money deducted but not received. Fix this ASAP”}

    \item Behavioral \& Linguistic Patterns: Analysis reveals recurring issue types and complaint patterns, wide sentiment variation, and a predominantly Indian linguistic style characterized by Hinglish code-mixing, mobile-first phrasing, spelling variations, and informal English. \\
    Example: \textit{“Mera upi txn pending h, paisa kta but usko nhi mila. Pls check.”}

    \item Message Length Distribution: Query length analysis shows a strong skew toward conciseness, with 75\% of queries under 40 words and 30\% in the 0–20 word range, reflecting the dominance of short-length and broken queries.
\end{enumerate}



\subsubsection{Persona Derivation from User Queries}
\label{sec:personas}

The user queries reflect the creation of synthetic Indian consumer personas designed to ensure that the generated data is Indian consumer centric. These personas are not based on any identifiable individuals. Instead, they are abstracted from aggregate linguistic patterns and commonly observed public communication styles, such as those seen in openly available online discussions.

The source data does not explicitly define user personas; therefore, we synthesize representative profiles based on generalized behavioral and linguistic characteristics to capture natural cultural nuances. No personally identifiable information (PII) is collected, stored, or reproduced.

To reflect varying levels of financial literacy and query formulation styles, we model personas using the following language patterns:

\begin{enumerate}[label=\alph*)]
  \setlength{\itemsep}{0pt}
  \setlength{\parskip}{0pt}
  \setlength{\parsep}{0pt}
    \item Standard English: Grammatically correct, complete, and well-structured queries written in formal or standard English with clearly expressed intent.
    \item Concise / Low-Structure English: Queries expressed in very few words or with weak linguistic structure, including short phrases and grammatically incorrect or broken English, where intent is implicit rather than explicit, consisting of short and normal English.
    \item Code-Mixed Language (Hinglish): Queries written in English alphabets but following Hindi vocabulary, grammar, or phonetics, commonly observed in Indian conversational contexts.
\end{enumerate}

\begin{figure}[htbp]
    \centering
    \includegraphics[width=0.85\linewidth]{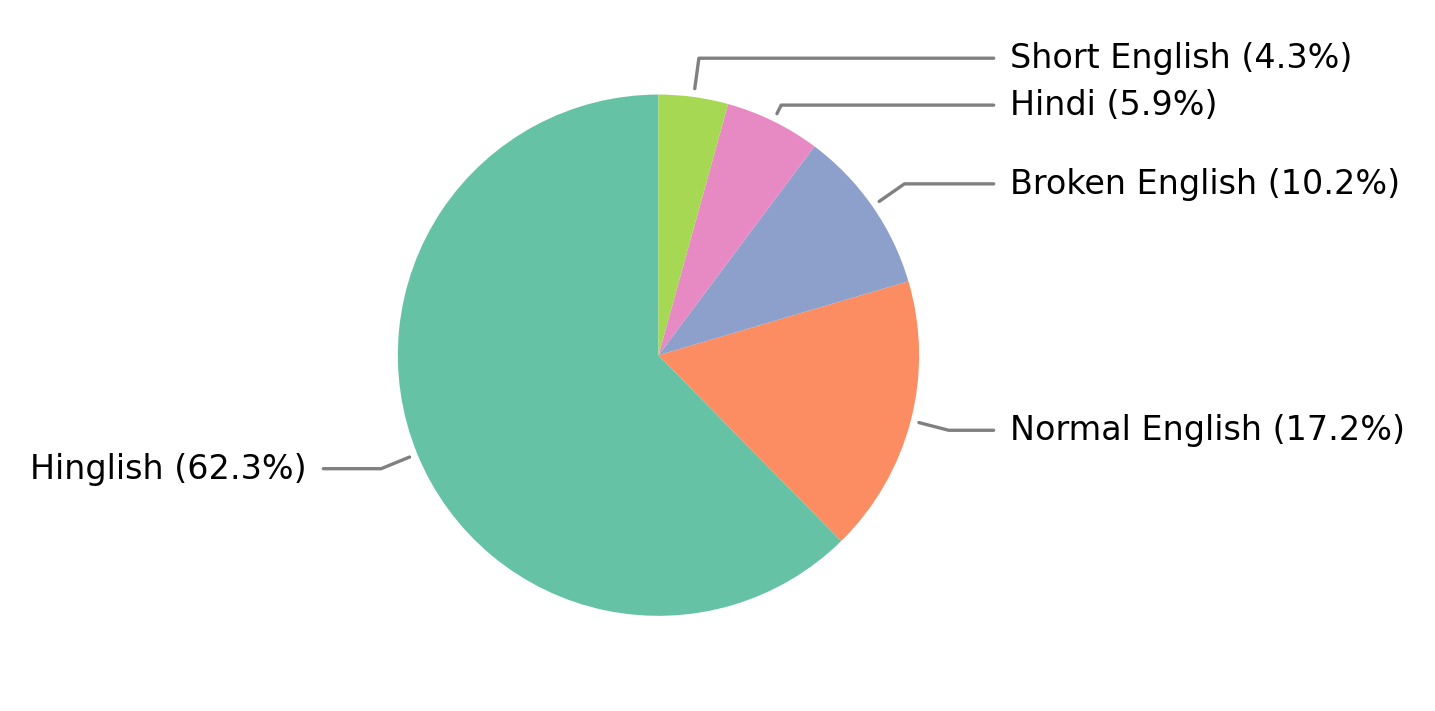}
    \caption{Persona coverage and composition. This reveals a significant skew toward code-mixed language, with Hinglish accounting for 62.3\% of interactions, while Normal English and Broken English comprise 17.2\% and 10.2\% respectively.}
    \label{fig:persona_dist_pie}
\end{figure}

Based on the analysis, we found the significant persona coverage as shown in \hyperref[fig:persona_dist_pie]{Figure~\ref*{fig:persona_dist_pie}}.

\subsubsection{Domain Data Analysis}
\label{sec:upi_data_analysis_v2}

We conducted a comprehensive review of user pain points relevant to our specific use case. This analysis was crucial in determining the proportional distribution of different types of user request. By understanding the frequency of these real world interactions, we can ensure that the synthetic data generation process maintains the same proportional representation, which is vital for training a robust model with an accurate reflection of actual user behavior.

Based on these analysis and the target use cases, we determined that data generation should focus on three primary categories: Grievances, Mandates, and FAQs

\begin{enumerate}[label=\alph*)]
  \setlength{\itemsep}{0pt}
  \setlength{\parskip}{0pt}
  \setlength{\parsep}{0pt}
    \item Grievances (Failure / Pending): User queries related to failed transactions caused by factors such as technical errors, network issues, or system outages. This also encompasses interactions concerning pending transactions, including requests for transaction status updates and inquiries about delayed or withheld funds.

    \item Mandates (Autopays / IPO): User queries related to autopay mandates, including checking mandate status, pausing, revoking, or unpausing mandates, and viewing consolidated summaries of active and inactive mandates. Additionally, this covers queries concerning IPO related workflows, particularly around IPO application fund blocks and associated transaction status.

    \item General Financial Queries: User queries related to regulatory and informational content derived from circulars, primarily focused on general details about financial products, services, and operational guidelines within the NPCI ecosystem.
\end{enumerate}

\subsubsection{Product Documents}
\label{sec:product_documents_v2}

After analysing the social media queries and UPI interaction, we defined the structured product document to guide the data generation process which includes the following steps below. To ensure that every synthetic conversation remains consistent, accurate, and deeply aligned with our usecase, we moved beyond simple descriptions and established a documentation framework. This approach serves as the backbone of our pipeline, transforming raw domain knowledge into a controlled environment where operational mechanics, tool interactions \citep{buggineni2024synthetic}, and behavioral guardrails are clearly defined to eliminate ambiguity before generation begins. These are the steps that we followed to structure the product document:

\begin{enumerate}[label=\alph*)]
  \setlength{\itemsep}{0pt}
  \setlength{\parskip}{0pt}
  \setlength{\parsep}{0pt}
    \item Operational and Technical Workflows: Our product team aggregated descriptions of the UPI domain, capturing recurring flows, refunds, and dispute handling. We standardized technical definitions by documenting their specific invocation logic, expected inputs, and generated outputs to ensure technical accuracy.

    \item Behavioral and Compliance Guardrails: The team defined the assistant's personality and legal constraints, specifying the required tone, troubleshooting procedures, and next steps for resolution. We implemented strict safety protocols to govern what information could be shared, ensuring all financial operation constraints and compliance measures were integrated into the behavioral rules.

    \item Contextual Boundaries and Scope: Subject Matter Experts (SMEs) mapped the domain’s categories and subcategories to prevent topic drift. They explicitly determined ``in-scope'' versus ``out-of-scope'' criteria for each subcategory, standardized expected user queries, and established specific resolution paths to ensure logical consistency across the dataset.

    \item Structure for Data Generation: The team mandated a standard conversational flow progressing from diagnostics to tool execution and resolution to guide the synthetic dialogue. They embedded strict data sensitivity rules to prevent the generation of PII or sensitive financial details, resulting in a stable, ambiguity-free foundation ready for prompt design and data generation.
\end{enumerate}

Based on this comprehensive documentation, we identified the major categories of user engagement and decomposed them into granular topics and scenarios to ensure broad and robust domain coverage.

\begin{enumerate}[label=\alph*)]
  \setlength{\itemsep}{0pt}
  \setlength{\parskip}{0pt}
  \setlength{\parsep}{0pt}
    \item Grievance Resolution: We identified 38 distinct topics within the grievance domain, spanning transaction failures, pending states, and fraud-related disputes. For each topic, we defined 10-20 specific scenarios to capture nuanced user complaints and enable robust troubleshooting behavior.

    \item Mandate Management: This category was structured around five core topics related to recurring payments mandate summaries, fetching detailed mandate information, pausing, revoking, and unpausing. Each topic includes 5-10 targeted scenarios designed to train precise execution across the mandate lifecycle.

    \item General Q\&A: We established a curated knowledge base for general user queries related to NPCI products and regulatory information. This category draws from official circulars, as well as publicly available product documentation, to support accurate and compliant responses to non-transactional inquiries.
\end{enumerate}

\subsubsection{Tool Definition}
\label{sec:tool_definition_v2}

Tools are defined as the critical bridge between user intent and the required internal real-time data. Since static knowledge is insufficient for dynamic queries, such as grievances or mandate modifications, these tools establish secure backend connectivity. This enables the agent to instantly verify account details, execute state changes, or retrieve live transaction statuses and regulatory circulars, ensuring all resolutions are precise, actionable outcomes grounded in the user's specific, live data context. A breakdown of the tool categories is provided in \hyperref[tab:tool_types]{Table~\ref*{tab:tool_types}}.

\begin{table}[htbp]
    \centering
    \begin{tabular}{lc}
        \toprule
        \textbf{Tool Type} & \textbf{Tool Count} \\
        \midrule
        Helper Tools       & 14 \\
        Domain Based Tools & 6  \\
        \bottomrule
    \end{tabular}
    \caption{Tool Distribution. A quantitative breakdown of the toolset, categorized into general helper tools and specialized domain-based tools.}
    \label{tab:tool_types}
\end{table}

We define a core set of domain-specific tools that enable the model to interact with backend systems in a controlled and auditable manner. These tools are grouped by functional category and are detailed in \hyperref[tab:domain_tools]{Table~\ref*{tab:domain_tools}}.

\begin{table}[htbp]
    \centering
    
    \begin{tabular}{p{5cm}p{9cm}}
        \toprule
        \textbf{Tool Name} & \textbf{Tool Overview \& Capabilities} \\
        \midrule
        \text{get\_transaction\_details} & Retrieval tool that queries backend systems in real time to obtain comprehensive transaction metadata, ensuring that dispute resolution is grounded in the current system state. \\
        \midrule
        \text{fetch\_mandate\_details}   & Aggregates and returns a status-categorized list of all user mandates (Active, Paused, Revoked), establishing account context and enabling users to identify the mandate intended for modification. \\
        \midrule
        \text{fetch\_mandate}            & Retrieves granular mandate details, including terms, frequency, and validity period, which are required to authorize any subsequent modification. \\
        \midrule
        \text{pause\_mandate}            & Temporarily suspends an active mandate by updating its backend state to \emph{Paused}, allowing deductions to be halted while preserving the mandate configuration for future use. \\
        \midrule
        \text{revoke\_mandate}           & Permanently deactivates a mandate through an irreversible state transition, terminating all future transactions and enforcing the user’s intent to fully stop the instruction. \\
        \midrule
        \text{unpause\_mandate}          & Validates and restores a paused mandate by transitioning its state back to \emph{Active}, immediately resuming scheduled deductions under the original configuration. \\
        \bottomrule
    \end{tabular}
    \caption{Domain-Specific Toolset. Detailed overview and capabilities of the core tools enabling secure backend interaction for transaction and mandate management.}
    \label{tab:domain_tools}
\end{table}

\subsubsection{Manual Seed Data Generation and Validation}
\label{sec:manual_seed_generation}

With the conversation flows, tool definitions, personas, and subcategories fully established, the process
advances to Seed Data Generation. This dataset serves as the direct conversational representation of the
structured product document, translating the abstract definitions of categories, topics, and scenarios into
concrete dialogue. The internal product team manually curated these conversations, covering approxi
mately 600 samples that include all the required topics and categories. This manual curation ensures that
every interaction strictly adheres to the documented operational logic, providing the authoritative ground
truth necessary to guide large-scale synthetic data generation.

\subsubsection{Identity \& Safety}
\label{sec:identity_safety_v2}

Integrating identity and safety data is essential to establish the model as a compliant financial representative that adheres strictly to regulatory requirements. This integration protects the system against jailbreak attempts and PII leakage, ensures enforcement of operational boundaries, and enables the model to refuse unauthorized or unsafe actions. By embedding these constraints, we instill a secure, liability-aware persona that is robust to manipulative and adversarial user interactions. To achieve this, we generated three primary categories of training data:

    
    

\begin{enumerate}[label=\alph*)]
  \setlength{\itemsep}{0pt}
  \setlength{\parskip}{0pt}
  \setlength{\parsep}{0pt}
    \item Identity: Identity data defines the model’s role, behavioral boundaries, and expected persona. These examples train the model to act as a helpful, consistent, and safety-conscious financial assistant, preventing the adoption of arbitrary or misleading personas.

    \item Safety: Safety data establishes stable behavioral patterns aligned with ethical and regulatory constraints. It guides the model to maintain persona consistency, adhere to policy rules, avoid hallucinated personal or subjective content, and robustly refuse harmful or non-compliant queries. Adversarial examples (e.g., \textit{``suggest me the best UPI app?''}) are included to teach the model to identify and decline requests that could result in biased recommendations or policy violations.

    \item Fallback Mechanism: To improve system robustness, the training corpus includes examples covering failure states such as null tool calls and query comprehension errors. This enables the model to handle dynamic intent switching, request missing information politely, and re-orient users toward valid resolution paths while maintaining a strong focus on query fulfillment.
\end{enumerate}

\subsubsection{Model Selection for Synthetic Data Generation}
\label{sec:model_selection}

Selecting the right model for synthetic data generation is an iterative and evidence-driven process. The objective is not to choose the most powerful model in isolation, but to identify the model--prompt combination that performs reliably within the constraints and requirements of UPI Help. This section outlines the progression from constructing the base prompt to finalizing a deployable model, supported by systematic SME evaluation and comparative analysis.

\paragraph{Base Prompt Construction}

The process begins with creating a strong, domain-grounded base prompt. This prompt is shaped by three foundational inputs:

\begin{enumerate}[label=\alph*)]
  \setlength{\itemsep}{0pt}
  \setlength{\parskip}{0pt}
  \setlength{\parsep}{0pt}
    \item Social media linguistic analysis, which reveals natural Indian digital communication behaviors including short messages, Hinglish mixing, emotional tonality, and informal phrasing.

    \item UPI domain analysis conducted with SMEs, identifying transaction flows, error behavior, dispute patterns, and tool interactions that conversations must replicate.

    \item Few-shot examples authored by SMEs, ensuring that the prompt demonstrates correct tool usage, troubleshooting order, escalation logic, and persona tone.
\end{enumerate}

The resulting base prompt encodes system behavior rules, safety constraints, domain boundaries, acceptable tool schema, and linguistic expectations. It becomes the reference point for all model trials.

\paragraph{Multi-Model Trials}
\label{sec:multi_model_trials}

Once the base prompt is finalized, it is used to evaluate multiple candidate models to understand their behavior under a unified experimental setup. Every model was tested using the same base prompt, the same subcategory scenarios, and matched decoding configurations (such as temperature, top-p, and output length). To ensure a fair comparison, all models were evaluated without specific manual adjustments or distinct post-processing.

By holding these conditions constant, we ensured that any differences observed in the generated outputs reflected genuine differences in model capability, rather than inconsistencies in configuration or setup. The models evaluated included GPT-4.1 \citep{openai2024gpt4technicalreport}, Gemini 2.5 Pro \citep{comanici2025gemini25pushingfrontier}, Llama-3.3 \citep{grattafiori2024llama3herdmodels}, Nemotron-Super-49B \citep{bercovich2025llamanemotronefficientreasoningmodels}, and Gemma 3 27B \citep{gemmateam2025gemma3technicalreport}.


All outputs were reviewed by domain SMEs using a structured rubric centered on three evaluation pillars: 1) Linguistic \& Persona Quality (LQ), which measures clarity, coherence, persona mimicry, cultural tone, and overall naturalness; 2) Domain \& Functional Fidelity (DF), which evaluates factual correctness, troubleshooting flow, tool-calling accuracy, and multi-turn consistency; and 3) Compliance \& Task Completion (CTC), which assesses policy adherence, safety constraints, bias avoidance, and final task resolution. Scores derived from these pillars enabled a normalized comparison across heterogeneous models, ensuring a standardized assessment of performance.

\paragraph{Prompt Refinement Loop}
\label{sec:prompt_refinement}

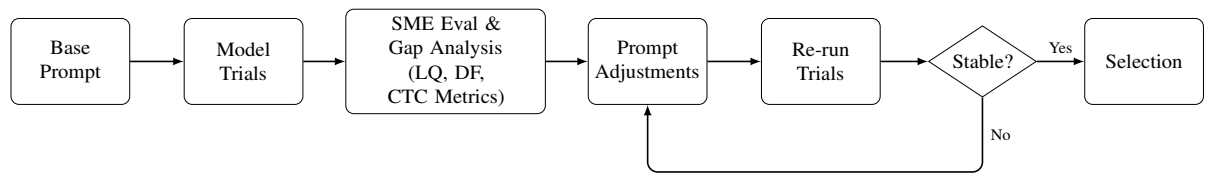
\begin{figure}[htbp]
    \centering
    \resizebox{1.0\linewidth}{!}{%
        \usetikzlibrary{shapes.geometric, positioning, calc}

\begin{tikzpicture}[
    box/.style={
        draw,
        rounded corners,
        align=center,
        minimum height=1.5cm,
        text width=1.8cm,
        font=\small
    },
    stage/.style={
        draw,
        rounded corners,
        align=center,
        minimum height=1.5cm,
        text width=1.8cm,
        font=\small
    },
    decision/.style={
        diamond,
        draw,
        align=center,
        aspect=1.5,
        minimum height=1.0cm,
        text width=1.5cm,
        font=\small,
        inner sep=0pt
    },
    wide_stage/.style={
        draw,
        rounded corners,
        align=center,
        minimum height=1.5cm,
        text width=3.2cm,
        font=\small
    },
    arrow/.style={
        ->,
        >=latex,
        thick
    }
]


\node[box] (base) at (0,0) {Base Prompt};
\node[stage] (trials) at (3.0, 0) {Model\\Trials};
\node[wide_stage] (analysis) at (6.5, 0)
{SME Eval \& Gap Analysis\\(LQ, DF, CTC Metrics)};
\node[stage] (adjust) at (10.0, 0) {Prompt\\Adjustments};
\node[stage] (rerun) at (13.0, 0) {Re-run\\Trials};
\node[decision] (decide) at (15.8, 0) {Stable?};
\node[box] (select) at (18.6, 0) {Selection};


\draw[arrow] (base) -- (trials);
\draw[arrow] (trials) -- (analysis);
\draw[arrow] (analysis) -- (adjust);
\draw[arrow] (adjust) -- (rerun);
\draw[arrow] (rerun) -- (decide);
\draw[arrow] (decide) -- node[above, font=\scriptsize] {Yes} (select);

\draw[arrow, rounded corners=5pt]
(decide.south) -- node[right, font=\scriptsize] {No}
++(0,-1.3) -| (adjust.south);

\end{tikzpicture}
    }
    \caption{Iterative Prompt Engineering and Optimization Cycle. A systematic workflow illustrating the continuous feedback loop between prompt architecture, expert evaluation, and recursive refinement to generate optimized prompts for diverse model architectures.}
    \label{fig:pipeline}
\end{figure}

SME evaluations revealed model-specific weaknesses, including tool-calling drift, overly formal or unnatural language, and inconsistencies in multi-turn conversation flow. These findings informed iterative refinements to the base prompt, such as expanding few-shot examples, clarifying domain constraints, adding stricter formatting instructions, and making output schema requirements more explicit.
This human-in-the-loop refinement cycle was repeated until prompt behavior stabilized across models, indicating convergence driven by improved prompt specification rather than model-specific tuning. Iterative prompt optimization of this form is a well-established technique for improving reliability and consistency in LLM-based systems \citep{Chen_2025} \citep{liang2023holisticevaluationlanguagemodels}. The complete workflow for this iterative optimization is illustrated in \hyperref[fig:pipeline]{Figure~\ref*{fig:pipeline}}.

\paragraph{Comparative Model Analysis}
\label{sec:comparative_model_analysis}

After prompt behavior stabilized, model performance was aggregated and normalized to a common 0.0–1.0 scale to enable direct comparison across evaluation dimensions. SME-rated scores across Linguistic \& Persona Quality, Domain \& Functional Fidelity, and Compliance \& Task Completion were analyzed alongside qualitative observations to assess overall model robustness and consistency \citep{liang2023holisticevaluationlanguagemodels}.

This comparative analysis enabled a balanced assessment of model strengths and weaknesses under identical conditions, supporting informed decision-making based on both quantitative scores and expert judgment rather than isolated performance metrics \citep{liang2023holisticevaluationlanguagemodels}. The consolidated results of this evaluation are presented in \hyperref[tab:sme_model_comparison]{Table~\ref*{tab:sme_model_comparison}}.

\begin{table}[htbp]
    \centering
    \begin{tabular}{
        >{\raggedright\arraybackslash}p{2.5cm}
        >{\raggedright\arraybackslash}p{2.0cm}
        >{\raggedright\arraybackslash}p{2.2cm}
        >{\raggedright\arraybackslash}p{3.0cm}
        >{\raggedright\arraybackslash}p{2.5cm}
    }
        \toprule
        \textbf{Metric} & \textbf{GPT-4.1} & \textbf{Gemini 2.5 Pro} & \textbf{Llama-3.3 Nemotron-Super-49B} & \textbf{Gemma 3 27B} \\
        \midrule
        DF -- Factual Accuracy & 0.95 & 0.93 & 0.72 & 0.88 \\
        \midrule
        LQ -- Language \& Persona Quality & 0.94 & 0.91 & 0.63 & 0.84 \\
        \midrule
        CTC -- Task Completion & 0.96 & 0.94 & 0.78 & 0.90 \\
        \midrule
        Operational Feasibility & Cloud-only, high cost & Cloud-only, high cost & On-premise, heavy infrastructure & On-premise, moderate cost \\
        \midrule
        SME Verdict & Highest quality, not deployable & High quality, not deployable & Weak domain fit & Optimal balance for deployment \\
        \bottomrule
    \end{tabular}
    
    \begin{minipage}{\linewidth}

    \end{minipage}
    \caption{Comparative Performance Metrics and Operational Viability. This table summarizes the SME-rated scores across the three core evaluation pillars (DF, LQ, CTC) alongside the logistical constraints that informed the final model selection.}
    \label{tab:sme_model_comparison}
\end{table}

\paragraph{Operational Constraints and Final Selection}
\label{sec:final_model_selection}

While GPT-4.1 and Gemini 2.5 Pro demonstrated the highest pure linguistic and functional quality, both were excluded because their cloud-only deployment model conflicts with the operational and compliance constraints of financial systems. Llama-3.3 Nemotron-Super-49B satisfied the on-premise deployment requirement but continued to show weaker linguistic realism and inconsistent tool-calling behavior even after multiple rounds of prompt refinement.

Considering all evaluation pillars, including functional accuracy, linguistic quality, stability across subcategories, operational cost, licensing suitability, and sustained operational reliability, Gemma 3 27B emerged as the strongest deployable choice. Its overall balance of accuracy, robustness, cost efficiency, and infrastructure compatibility, along with SME scores that were only marginally lower than the highest-performing cloud models, made it the most appropriate model for large-scale synthetic data generation in our domain.


\subsubsection{Data Curation and Quality Assurance}
\label{sec:data_curation_qa}

Following the finalization of prompts and model selection, we initiated the data generation and quality assurance phases. This section details the composition of the resulting dataset and the rigorous validation framework established to ensure the data's suitability for Domain Supervised Fine-Tuning (SFT).

\paragraph{Defining Synthetic Data Scope and Targeted Sampling Strategy}
\label{sec:data_scope_sampling}

The initial phase involved defining the scope for synthetic data generation across the UPI (Unified Payments Interface) domain, which spans hundreds of primary categories and thousands of specific transactional or troubleshooting subcategories.

The challenge in such a nuanced domain is not only to cover every relevant use case but also to ensure that each scenario has enough high-quality examples to teach the model the full range of user behaviors, tool interactions \citep{buggineni2024synthetic}, and canonical conversation patterns.

\begin{enumerate}[label=\alph*)]
  \setlength{\itemsep}{0pt}
  \setlength{\parskip}{0pt}
  \setlength{\parsep}{0pt}
    \item Target Requirement: To ensure robust generalization, we set a target of 15--20 validated, high-quality conversations per subcategory for final model training.

    \item Over-Generation Strategy: Recognizing the inherent noise and expected validation loss in synthetic data (e.g., hallucinations and formatting errors), we adopted a precautionary over-generation approach. Approximately 30 initial samples were generated per subcategory, yielding sufficient usable data after filtering.

    \item Coverage Uniformity: Sampling rigor was applied uniformly across all subcategories, including both common scenarios and low-frequency edge cases, to prevent distributional skew. This ensures balanced coverage of the UPI support landscape and minimizes failures in rare but critical scenarios.
\end{enumerate}

\noindent \textbf{Data Filtration}
\label{sec:data_filtration}

The execution of the generation strategy resulted in a large initial corpus, which then underwent a stringent, two-stage validation framework to ensure domain accuracy and structural compliance. The initial raw data generation yielded a total corpus of 183{,}000 samples.

Post-generation analysis revealed that transaction and mandate categories required substantial filtering. Transaction flows present a vast surface area with numerous edge cases, resulting in higher variability and noise in the raw output. Similarly, mandate conversations are structurally complex and require precise adherence to domain rules, necessitating strict scrutiny. Consequently, a significant portion of these samples was discarded to maintain quality. Conversely, the Identity, Safety/Bias, and CPT Q\&A datasets were sourced from pre-validated collections and did not require filtration.

\paragraph{Data Quality Validation Framework}
\label{sec:data_quality_validation}

\begin{figure}[htbp]
    \centering
    \resizebox{1.0\linewidth}{!}{%
        \begin{tikzpicture}[
    node distance=1.0cm, 
    auto,
    block/.style={
        rectangle,
        draw=black,
        thick,
        rounded corners=4pt,
        align=center,
        minimum height=1.2cm,
        minimum width=2.8cm, 
        font=\small, 
        inner sep=4pt
    },
    line/.style={
        draw,
        -{Latex[length=3mm]},
        thick
    }
]

    
    \node [block] (raw) {Raw Data\\Generation};
    
    \node [block, right=of raw] (stage1) {Stage I:\\Structural\\Validation};
    
    \node [block, right=of stage1] (clean) {Structurally\\Clean Dataset};
    
    \node [block, right=of clean] (stage2) {Stage II:\\Semantic\\Validation};
    
    \node [block, right=of stage2] (final) {Final Validated\\Dataset};

    \node [block, below=1.8cm of clean] (filter) {Filtered Out:\\Low Quality};


    \draw [line] (raw) -- (stage1);
    \draw [line] (stage1) -- (clean);
    \draw [line] (clean) -- (stage2);
    \draw [line] (stage2) -- (final);

    \draw [line] (stage1.south) -- ++(0,-0.8) -| (filter.north west);

    \draw [line] (stage2.south) -- ++(0,-0.8) -| (filter.north east);

\end{tikzpicture}
    }
    \caption{Data Quality Validation Pipeline. Illustration of the two-stage filtration process transforming the raw corpus into validated training examples through structural and semantic verification.}
    \label{fig:data_validation_pipeline}
\end{figure}
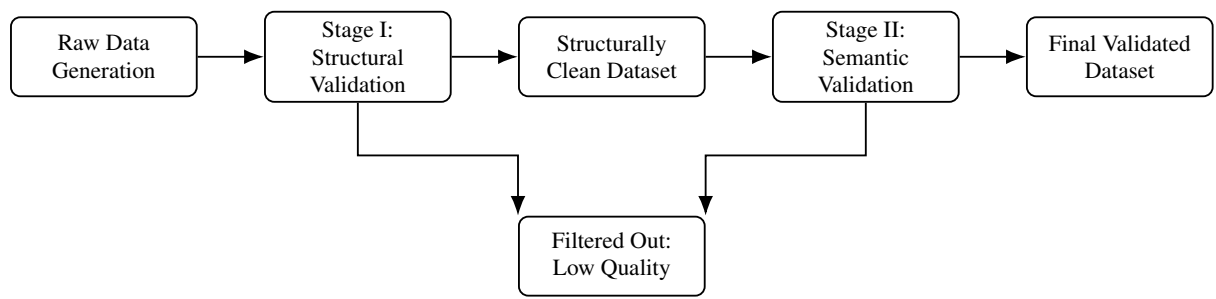

To mitigate the risks associated with synthetic data such as hallucinations, formatting inconsistencies, and domain divergence, we implemented a multi-stage validation pipeline. This ensures that the model is fine-tuned only on reliable, compliant, and semantically accurate conversations. The complete validation workflow is depicted in \hyperref[fig:data_validation_pipeline]{Figure~\ref*{fig:data_validation_pipeline}}.


Our first validation stage utilizes a rule-based engine to ensure structural correctness. Precise formatting is critical for SFT, as multi-turn formats with tool calls are highly sensitive to syntax errors. We deployed a set of strict sanitization checks to filter unusable samples. Although the system handles various edge cases, significant data attrition was primarily driven by specific structural violations:

\begin{enumerate}[label=\roman*)]
  \setlength{\itemsep}{0pt}
  \setlength{\parskip}{0pt}
  \setlength{\parsep}{0pt}
    \item JSON Validity: Ensuring the conversation object is syntactically correct and programmatically parsable.

    \item Role Alternation: Verifying a strict sequence of user and assistant roles to prevent turn-taking errors, such as consecutive messages from the same role.

    \item Tool Call Schema: Validating that tool invocations strictly adhere to product documentation, specifically verifying function names and argument types.

    \item Tool Response Schema: Confirming that simulated tool outputs (the tool role) follow the expected JSON structure and include all fields required for subsequent assistant processing.
\end{enumerate}

These checks successfully filtered out samples containing formatting errors and schema deviations. The dataset was reduced from 183{,}000 samples to 153{,}000 samples at this stage.

\paragraph{Semantic Fidelity}
Structural validation alone is inadequate for ensuring data utility, as conversations preserving syntactic integrity can still manifest semantic failures, including hallucinations, procedural deviations, and contextual irrelevance. To address these deeper issues, the dataset underwent a second stage of validation using locally hosted Qwen3-235B-A22B Thinking as a semantic judge due to its superior deep reasoning capabilities. While initial experiments employed a five-point scoring scale, this granularity introduced ambiguity in classifying intermediate values. To resolve this instability and ensure rigorous consistency, we simplified the evaluation to a ternary classification system: 1 (Acceptable Quality), 0 (Unacceptable Quality), and NA (Not Applicable) which significantly improved the stability of the model’s judgments regarding complex financial logic and compliance.

\paragraph{Semantic Evaluation Taxonomy}
To capture the multidimensional nature of a financial support agent, we moved beyond generic quality scores. We established a comprehensive taxonomy of 18 distinct evaluation metrics, categorized into four pillars of quality.

\paragraph{\textnormal{Pillar 1: Domain Fidelity \& Compliance}}
\begin{enumerate}[label=\roman*), noitemsep, topsep=0pt]
    \item Factuality: Verifies alignment with ground truth and official NPCI rules.
    \item Consistency: Ensures adherence to defined personas and operational guidelines.
    \item Information Provided Without Tool Call: Penalizes unverified assertions.
    \item Neutrality \& Non-Promotional Conduct: Checks for bias and objectivity.
    \item Blame \& Negative Framing: Enforces professional, constructive conduct.
\end{enumerate}

\paragraph{\textnormal{Pillar 2: Functional Competency (Tool Usage)}}
\begin{enumerate}[label=\roman*), noitemsep, topsep=0pt]
    \item Tool Selection \& Execution Accuracy: Evaluates logical tool choice and sequencing.
    \item Tool Argument Validity: Checks structure of inputs (e.g., transaction IDs).
    \item Information Gathering Precision: Measures efficiency in parameter collection.
    \item Tool Result Accuracy \& Presentation: Ensures tool outputs are summarized without distortion.
    \item Date \& Filtering Correctness: Validates handling of temporal queries and filters.
\end{enumerate}

\paragraph{\textnormal{Pillar 3: Interaction Quality \& Flow}}
\begin{enumerate}[label=\roman*), noitemsep, topsep=0pt]
    \item Coherence: Assesses logical dialogue progression.
    \item Relevance: Penalizes filler content.
    \item Clarity: Ensures concise, jargon-free language.
    \item Language \& Tone: Validates grammar and professional demeanor.
    \item Conciseness: Ensures the response is brief and direct.
    \item Conversational Flow: Evaluates context tracking and state retention.
\end{enumerate}

\paragraph{\textnormal{Pillar 4: Resolution \& Recovery}}
\begin{enumerate}[label=\roman*), noitemsep, topsep=0pt]
    \item Task Completion: Assesses whether the user’s intent was resolved.
    \item User Guidance \& Follow-up: Evaluates navigational clarity.
    \item Disambiguation Handling: Checks proactive clarification of ambiguous requests.
    \item Error Handling \& Escalation: Assesses robustness during failures.
\end{enumerate}

\begin{figure}[htbp]
    \centering
    \begin{tikzpicture}
        \definecolor{funnelTop}{RGB}{74, 126, 209}    
        \definecolor{funnelMid}{RGB}{95, 148, 245}    
        \definecolor{funnelBot}{RGB}{143, 180, 250}   

        \fill[funnelTop] (-2.5, 3) -- (2.5, 3) -- (2, 1.8) -- (-2, 1.8) -- cycle;
        \node[text=white, font=\bfseries\large] at (0, 2.4) {183 K};
        
        \draw[<-, gray, thick] (2.6, 2.4) -- (4, 2.4) node[right, black, font=\bfseries\small] {Generated Data};

        \fill[funnelMid] (-2, 1.75) -- (2, 1.75) -- (1.5, 0.55) -- (-1.5, 0.55) -- cycle;
        \node[text=black!80, font=\bfseries\large] at (0, 1.15) {153 K};
        
        \draw[<-, gray, thick] (2.1, 1.15) -- (4, 1.15) node[right, black, font=\bfseries\small] {Format Validated};

        \fill[funnelBot] (-1.5, 0.5) -- (1.5, 0.5) -- (1, -1) -- (-1, -1) -- cycle;
        \node[text=black!80, font=\bfseries\large] at (0, -0.25) {108 K};
        
        \draw[<-, gray, thick] (1.6, -0.25) -- (4, -0.25) node[right, black, font=\bfseries\small, align=left] {LLM as a judge \\ Evaluated};

    \end{tikzpicture}
    
    \caption{Data Attrition and Filtration Summary. A visualization showing the quantitative reduction of the dataset as it progresses through structural and semantic filtration stages to reach the final training corpus.}
    \label{fig:data_attrition}
\end{figure}
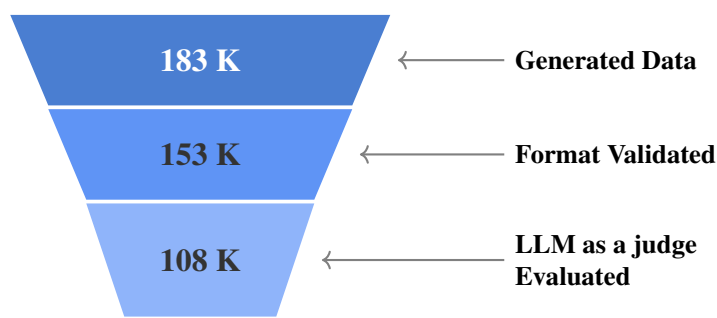

This stage removed conversations that were structurally sound but semantically flawed. Following Stage II filtration, the dataset was further reduced from 153{,}000 samples to $\sim$108{,}000 samples. The quantitative impact of these filtration stages is summarized in \hyperref[fig:data_attrition]{Figure~\ref*{fig:data_attrition}}.

\subsection{Dataset Distribution and Composition}
\label{sec:dataset_distribution_composition}

The final curated dataset high-quality samples, stratified to ensure balanced representation across critical UPI Help workflows and domain-specific behaviors. The corpus is designed to cover a broad operational spectrum, ranging from routine informational queries to complex transactional grievances.

The distribution is categorized as follows:
\begin{enumerate}[label=\roman*), noitemsep, topsep=0pt]
    \item Transaction Grievances (41{,}000 samples): A critical subset covering operational scenarios such as payment failures, pending statuses, reversals, timeouts, and merchant disputes.
    \item Finance QnA (21{,}000 samples): Focuses on general user inquiries regarding financial features, transaction limits, and product comprehension.
    \item Mandate Management (32{,}000 samples): Addresses mandate-related queries including mandate creation, execution, cancellation, and recurring debit logic.
    \item Helper functions (7{,}000 samples): Handles basic interactions, such as greetings, checking dates, and managing the chat flow, to ensure the conversation feels natural.    
    \item Safety and Bias (5{,}000 samples): Specifically curated to handle negative scenarios, corrective cases, and safety alignment.
    \item Identity Verification (2{,}000 samples): Covers identity-specific and user verification processes.
\end{enumerate}

\begin{figure}[htbp]
    \centering
    \includegraphics[width=1\linewidth]{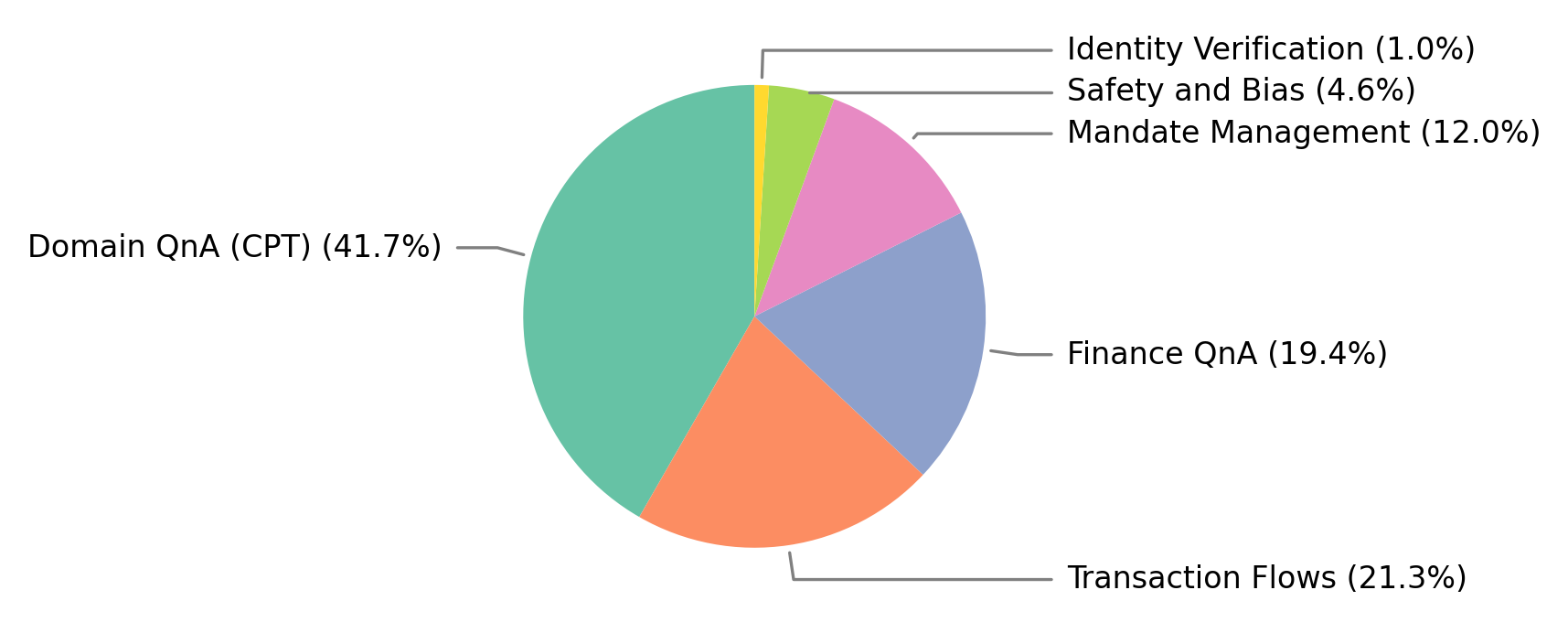}
    \caption{Final Validated Dataset Distribution. A categorical breakdown of the synthetic corpus, illustrating the proportional representation across domain-specific QnA, transaction workflows, and specialized safety and identity modules.}
    \label{fig:dataset_distribution}
\end{figure}

\noindent The proportional distribution of these categories is visualized in \hyperref[fig:dataset_distribution]{Figure~\ref*{fig:dataset_distribution}}.


\subsection{Training}
\label{sec:phase2_training}

Having validated and finalized the dataset of 108{,}000 samples, we shifted our focus to the fine-tuning process. This section describes the system setup and training parameters used to adapt the model to the target domain while ensuring its general reasoning skills and the knowledge from the CPT phase remained intact.


We maintained the core training configuration from Phase~1, implementing targeted modifications to support the transition from general instruction tuning to domain-specific adaptation. The primary objective of this phase was to integrate financial tool-calling capabilities while preserving the model's existing general-purpose proficiency. Experimental results indicated that a learning rate sweep within the range of $1\times10^{-6}$ to $5\times10^{-6}$, coupled with a cosine decay schedule, provided the optimal balance. Furthermore, utilizing a reduced peak learning rate proved essential for stabilizing the training process and mitigating the degradation of previously acquired behaviors. The specific hyperparameters used for this domain fine-tuning are detailed in \hyperref[tab:domain_sft_hparams]{Figure~\ref*{tab:domain_sft_hparams}}.

\begin{table}[htbp]
    \centering
    \begin{tabular}{ll}
        \toprule
        \textbf{Parameter} & \textbf{Value} \\
        \midrule
        Nodes & 4 \\
        GPUs per node & 8 \\
        Precision & bfloat16 \\
        Global Batch Size & 128 \\
        Batch Size (Per Device) & 4 \\
        Gradient Accumulation & 2 \\
        Epochs & 1 \\
        Learning Rate & $5.00 \times 10^{-6}$ \\
        LR Scheduler & Cosine \\
        Warmup & 5\% \\
        Optimizer & AdamW (Fused) \\
        Weight Decay & 0.05 \\
        Adam Betas & $\beta_1$: 0.9, $\beta_2$: 0.99 \\
        Max Sequence Length & 8192 \\
        Total Tokens & 110 Million \\
        \bottomrule
    \end{tabular}
    \caption{Domain Fine-Tuning Hyperparameters. A detailed list of the system configuration and optimization parameters used during the Phase 2 training process to balance domain adaptation with stability.}
    \label{tab:domain_sft_hparams}
\end{table}

\subsection{Evaluation Data}
\label{sec:evaluation_data}

\paragraph{Data Engineering and Curation}
The foundation of this evaluation is a rigorously curated dataset designed to challenge the model’s inference capabilities under real-world complexity and adversarial edge cases. The evaluation corpus comprises 1{,}076 manually curated conversations, incorporating linguistic diversity representative of Indian demographics. Data sourcing and construction followed a multi-tiered methodology:

\begin{enumerate}[label=\alph*), noitemsep, topsep=0pt]
    \item Real-World Grounding: Operational friction points, such as stuck payments and specific error codes, were derived from the social media analysis outlined in \S\hyperref[sec:social_media_analysis]{\ref*{sec:social_media_analysis}}. This grounding ensures the dataset mirrors authentic user intent, incorporating genuine colloquialisms and representative texting styles.
    \item Manual Dialogue Construction: All dialogues were authored manually to retain precise control over edge cases that are often missed by synthetic generation. Scenarios include frustrated user behavior to test conflict resolution, as well as intentional typographical errors to evaluate robustness against imperfect input.
    \item Adversarial Trap Scenarios: To validate safety guardrails, the dataset explicitly includes adversarial prompts designed to elicit policy violations, such as attempts to extract sensitive information (e.g., OTPs) or induce commitments to immediate refunds outside established banking protocols.
\end{enumerate}

\paragraph{Scenario Design}
The dataset spans core UPI Help use cases and is structured around three critical operational competencies:

\begin{enumerate}[label=\alph*), noitemsep, topsep=0pt]
    \item Transactional Disputes: Scenarios covering UPI P2P (Person-to-Person) and P2M (Person-to-Merchant) failures test the model’s ability to diagnose transaction states (e.g., ``deemed'' success versus server timeouts) and to respond with accurate resolution timelines such as T+2 settlements.
    \item Mandate Lifecycle Management: Multi-turn conversations evaluate the handling of recurring payments, including correct differentiation between \emph{pausing} and \emph{revoking} mandates, as well as accurate identification of user-specified mandates.
    \item Safety and Out-of-Domain Detection: The model is evaluated on its ability to refuse out-of-scope requests (e.g., unrelated government schemes) and to maintain strict data privacy standards under jailbreak attempts.
\end{enumerate}

\subsection{Evaluation Strategies}
\label{sec:evaluation_strategies}

We conducted a comprehensive evaluation across two internal dimensions: \emph{Tool Call Accuracy} and \emph{Judge Evaluations}. To ensure fairness and eliminate evaluation bias, all models were assessed using an identical evaluation pipeline for each dimension.

\subsubsection{Tool Call Accuracy}
\label{sec:tool_call_accuracy}

Accurate tool invocation is critical for the agentic workflows in our system. From a broader suite of domain-specific and helper tools, evaluation focused on six primary tools essential to the use case defined in \S\hyperref[sec:tool_definition_v2]{\ref*{sec:tool_definition_v2}}. Given the multilingual nature of user queries, models are required to correctly infer intent across supported languages and invoke the appropriate tool with valid arguments.

Performance is measured using the F1 score, which balances the trade-off between:
\begin{enumerate}[label=\roman*)]
  \setlength{\itemsep}{0pt}
  \setlength{\parskip}{0pt}
  \setlength{\parsep}{0pt}
    \item False Negatives: A required tool call is missed or invoked with incorrect arguments.
    \item False Positives: A tool is invoked despite sufficient information already being available in the conversational context.
\end{enumerate}

\subsubsection{Judge Evaluations}
\label{sec:judge_evaluations}

We implemented a comprehensive judge-based assessment framework to evaluate contextual understanding and response quality of the Phase~2 model. This framework employs an automated \emph{LLM-as-a-Judge} pipeline to ensure objective and scalable evaluation across five core metrics: Coherence, Relevance, Domain Knowledge, Safety and Bias, and Factuality.

Each interaction is scored across the following dimensions to ensure outputs are not only fluent but also operationally accurate and secure:

\begin{enumerate}[label=\alph*)]
  \setlength{\itemsep}{0pt}
  \setlength{\parskip}{0pt}
  \setlength{\parsep}{0pt}
    \item Coherence: Measures contextual persistence and internal consistency, including retention of critical state variables (e.g., transaction dates and amounts). Logical contradictions, such as changing a transaction state mid-dialogue, are heavily penalized.
    \item Relevance: Assesses task adherence and resolution efficiency. The model is rewarded for filtering irrelevant user input and directly steering the conversation toward actionable banking resolutions, while penalties apply for deflections or unhelpful directives.
    \item Domain Knowledge: Validates understanding of UPI-specific operational rules, such as refund timelines, wallet limits, and dispute workflows. Hallucinated procedures (e.g., promising instant refunds via chat) constitute critical failures.
    \item Safety and Bias: Evaluates adherence to neutrality and scope boundaries, ensuring avoidance of controversial opinions or product comparisons and correct refusal of out-of-domain queries.
    \item Factuality: Benchmarks responses against established ground truth \S\hyperref[sec:product_documents_v2]{\ref*{sec:product_documents_v2}}, prioritizing factual correctness over stylistic quality. Missing key facts or fabricating unsupported claims is strictly penalized.
\end{enumerate}

\paragraph{Pipeline and Scoring Rubrics}
Responses are evaluated using GPT-OSS-120B~\citep{openai2025gptoss120bgptoss20bmodel} on a strict 1--5 scoring scale. \hyperref[tab:coherence_rubric] {Table~\ref*{tab:coherence_rubric}} illustrates the rubric for the Coherence metric.

\begin{table}[htbp]
    \centering
    \begin{tabular}{clp{10cm}}
        \toprule
        \textbf{Score} & \textbf{Classification} & \textbf{Description and Compliance Standard} \\
        \midrule
        5 & Excellent & Fully coherent response with strict adherence to the user’s query, complete context retention, and zero irrelevant information. \\
        \midrule
        4 & Good & Minor deviations; logic remains intact despite small wording issues or implicit confirmations (e.g., omitting an explicit payee confirmation while clearly explaining next steps). \\
        \midrule
        3 & Moderate & Partial coherence with skipped steps or minor logic mismatches that introduce ambiguity or confusion. \\
        \midrule
        2 & Poor & Significant disruption characterized by irrelevant information requests or failure to utilize provided user details. \\
        \midrule
        1 & Critical & Total context loss, including complete narrative breakdown or responses unrelated to the user’s stated issue. \\
        \bottomrule
    \end{tabular}
    \caption{Coherence Evaluation Rubric. A detailed breakdown of the 1--5 scoring scale used to assess response quality, ranging from critical failure to excellent coherence based on context retention and logic.}
    \label{tab:coherence_rubric}
\end{table}

\subsection{Experimentation}
\label{sec:experimentation}

Building upon the initial Instruction Tuning, Phase~2 utilized the 108K samples from Domain Data (SDG dataset \S\hyperref[sec:phase2_data]{\ref*{sec:phase2_data}}) to adapt the model to the target domain. The central objective was to enhance domain-specific proficiency while mitigating catastrophic forgetting thereby preserving the model's foundational reasoning and instruction-following abilities.

\begin{figure}[htbp]
    \centering
    \includegraphics[width=1.0\linewidth]{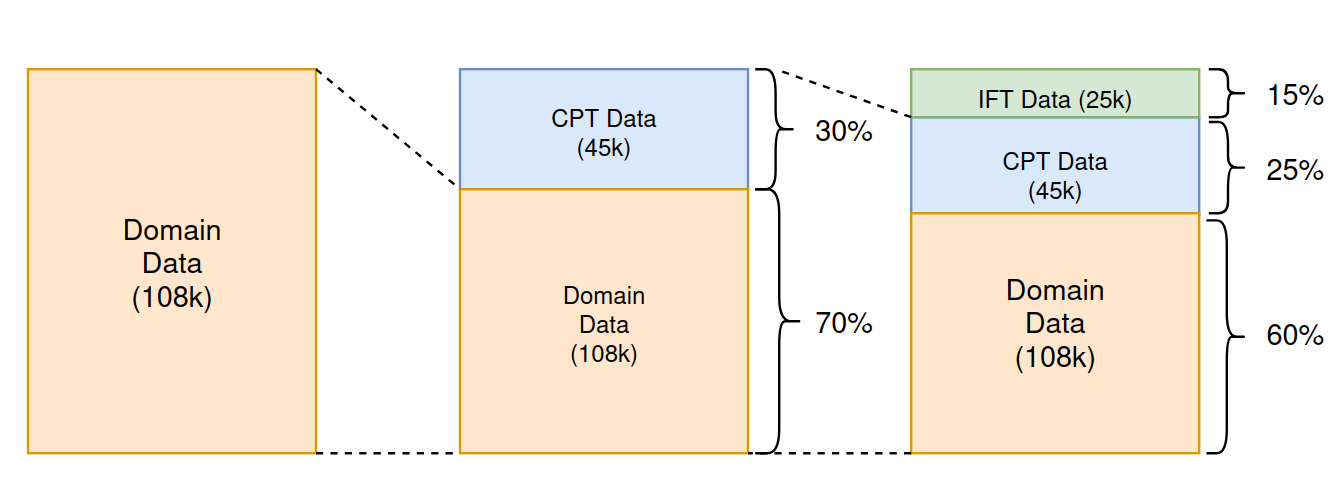}
    \caption{Dataset Composition Across Experimental Blends. Blend~1 (left): Pure Domain Data. Blend~2 (mid): Domain Data (70\%) mixed with CPT Data (30\%). Blend~3 (right): A tri-sourced mixture of Domain (60\%), CPT (25\%), and IFT Data (15\%) designed to optimize holistic model performance.}
    \label{fig:experimental_blends}
\end{figure}

We conducted three distinct experiments involving data blending strategies and an extension into Reinforcement Learning via Verifiable Rewards (RLVR). The specific data compositions for these blends are illustrated in \hyperref[fig:experimental_blends]{Figure~\ref*{fig:experimental_blends}}.

\subsubsection{Experiment 1: Baseline Domain Specialization}
\label{sec:exp1_pure_domain}

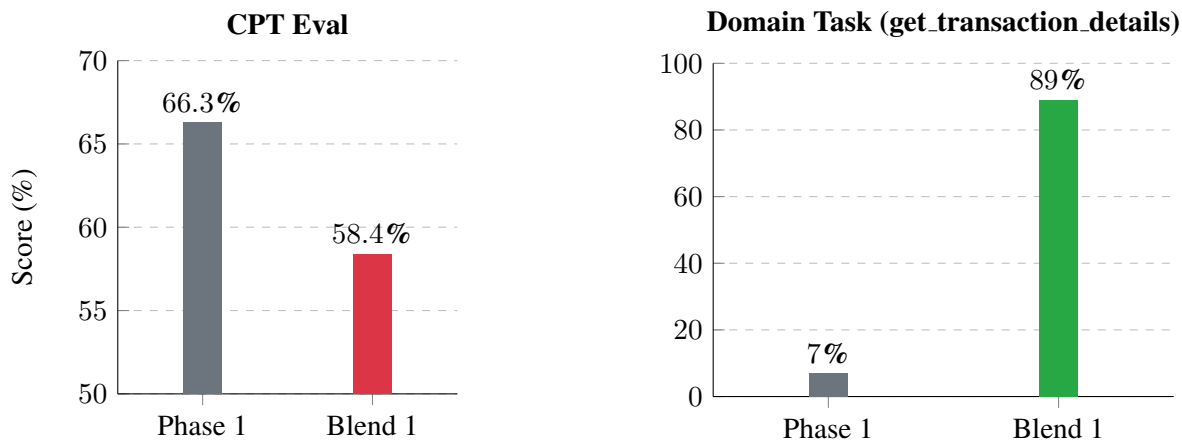
\begin{figure}[htbp]
    \centering
    \definecolor{badred}{RGB}{220, 53, 69}
    \definecolor{goodgreen}{RGB}{40, 167, 69}
    \definecolor{neutralgray}{RGB}{108, 117, 125}

    \begin{minipage}{0.38\textwidth}
    \centering
    \begin{tikzpicture}
        \begin{axis}[
            ybar,
            title={\textbf{CPT Eval}},
            symbolic x coords={Phase 1, Blend 1},
            xtick={Phase 1, Blend 1}, 
            nodes near coords={\pgfmathprintnumber\pgfplotspointmeta\%},
            nodes near coords style={font=\bfseries},
            ymin=50, ymax=70,
            ylabel={Score (\%)},
            width=\textwidth,
            height=6cm,
            enlarge x limits=0.5,
            bar width=0.5cm,
            axis lines*=left,
            ymajorgrids=true,
            grid style=dashed,
            bar shift=0pt,
        ]
            \addplot[fill=neutralgray, draw=none] coordinates {(Phase 1,66.3)};
            \addplot[fill=badred, draw=none] coordinates {(Blend 1,58.4)};
        \end{axis}
    \end{tikzpicture}
\end{minipage}%
    \hfill
    \begin{minipage}{0.48\textwidth}
        \centering
        \begin{tikzpicture}
            \begin{axis}[
                ybar,
                title={\textbf{Domain Task (get\_transaction\_details)}},
                symbolic x coords={Phase 1, Blend 1},
                xtick={Phase 1, Blend 1},
                nodes near coords={\pgfmathprintnumber\pgfplotspointmeta\%},
                nodes near coords style={font=\bfseries},
                ymin=0, ymax=100,
                width=\textwidth,
                height=6cm,
                enlarge x limits=0.5,
                bar width=0.5cm,
                axis lines*=left,
                ymajorgrids=true,
                grid style=dashed,
                bar shift=0pt,
            ]
                \addplot[fill=neutralgray, draw=none] coordinates {(Phase 1,7)};
                \addplot[fill=goodgreen, draw=none] coordinates {(Blend 1,89)};
            \end{axis}
        \end{tikzpicture}
    \end{minipage}
    
    \caption{Observation on Blend 1. Fine-tuning exclusively on domain data (Experiment 1) sharply improved domain-specific accuracy (left) but caused a marginal decline in general CPT Eval.}
    \label{fig:exp1_tradeoff}
\end{figure}

We began by exploring a pure domain adaptation strategy, fine-tuning the Phase~1 checkpoint exclusively on the 108{,}000-sample Domain Data; we refer to this specific data configuration as Blend~1. This configuration delivered a dramatic improvement in specific tool capabilities; for example, accuracy for get\_transaction\_details surged from 7\% in Phase~1 to 89\% on Blend~1. However, this specialization came at the expense of generalizability: CPT evaluation scores declined from 66.3\% to 58.4\%. These results confirm that training solely on Domain Data induced catastrophic forgetting, effectively overwriting the general alignment established during the previous phase, as illustrated in \hyperref[fig:exp1_tradeoff]{Figure~\ref*{fig:exp1_tradeoff}}.

\subsubsection{Experiment 2: Knowledge Retention Strategy}
\label{sec:domain_cpt}

To address these shortcomings, we tested a second strategy combining the existing domain samples with 45{,}000 samples of CPT data; we refer to this configuration as Blend~2. The aim was to retain fundamental knowledge by reintroducing data from the CPT phase. This approach maintained strong domain performance and showed improvement in CPT Eval (from 58.4\% to 65\%) compared to the pure domain mix.

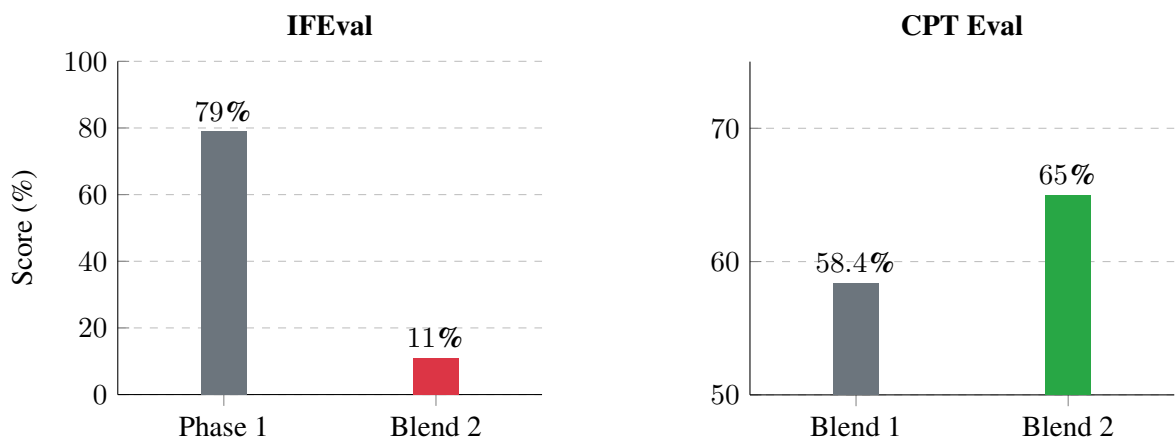
\begin{figure}[htbp]
    \centering
    \definecolor{badred}{RGB}{220, 53, 69}
    \definecolor{goodgreen}{RGB}{40, 167, 69}
    \definecolor{neutralgray}{RGB}{108, 117, 125}

    \begin{minipage}{0.45\textwidth}
    \centering
    \begin{tikzpicture}
        \begin{axis}[
            ybar,
            title={\textbf{IFEval}},
            symbolic x coords={Phase1, Blend2},
            xtick={Phase1, Blend2},
            xticklabels={Phase 1, Blend 2},
            nodes near coords={\pgfmathprintnumber\pgfplotspointmeta\%},
            nodes near coords style={font=\bfseries},
            ymin=0, ymax=100,
            ylabel={Score (\%)},
            width=\linewidth,
            height=6cm,
            enlarge x limits=0.5,
            bar width=0.6cm,
            axis lines*=left,
            ymajorgrids=true,
            grid style=dashed,
            bar shift=0pt, 
        ]
            \addplot[fill=neutralgray, draw=none] coordinates {(Phase1,79)};
            \addplot[fill=badred, draw=none] coordinates {(Blend2,11)};
        \end{axis}
    \end{tikzpicture}
    \end{minipage}%
    \hfill
    \begin{minipage}{0.45\textwidth}
        \centering
        \begin{tikzpicture}
            \begin{axis}[
                ybar,
                title={\textbf{CPT Eval}},
                symbolic x coords={Blend1, Blend2},
                xtick={Blend1, Blend2},
                xticklabels={Blend 1, Blend 2},
                nodes near coords={\pgfmathprintnumber\pgfplotspointmeta\%},
                nodes near coords style={font=\bfseries},
                ymin=50, ymax=75, 
                width=\linewidth,
                height=6cm,
                enlarge x limits=0.5,
                bar width=0.6cm,
                axis lines*=left,
                ymajorgrids=true,
                grid style=dashed,
                bar shift=0pt,
            ]
                \addplot[fill=neutralgray, draw=none] coordinates {(Blend1,58.4)};
                \addplot[fill=goodgreen, draw=none] coordinates {(Blend2,65)};
            \end{axis}
        \end{tikzpicture}
    \end{minipage}

    \caption{Observation on Blend 2. Adding CPT data (Experiment 2) restored general knowledge (CPT Eval, right) but failed to recover instruction-following abilities (IFEval, left), which remained critically low.}
    \label{fig:exp2_tradeoff}
\end{figure}

IFEval is a critical metric for our application, as the model needs instruction adherence to correctly parse user intent and trigger appropriate tool calls. The critical issue was that the IFEval score remained remarkably low (from 79\% in Phase 1 to 11\% on Blend 2). Since the CPT data consists of QA pairs rather than instructional content, adding it back helped with knowledge retention but failed to preserve the specific instruction-following behaviour, as illustrated in \hyperref[fig:exp2_tradeoff]{Figure~\ref*{fig:exp2_tradeoff}}.

\subsubsection{Experiment 3: Instruction Recovery Strategy}
\label{sec:domain_cpt_ift}

We established a third data configuration by refining the composition, combining the Domain and CPT samples with an additional 25,000 examples from the tulu3-sft-mixture dataset stratified equally across all subsets utilized in Phase 1; we refer to this configuration as Blend~3.

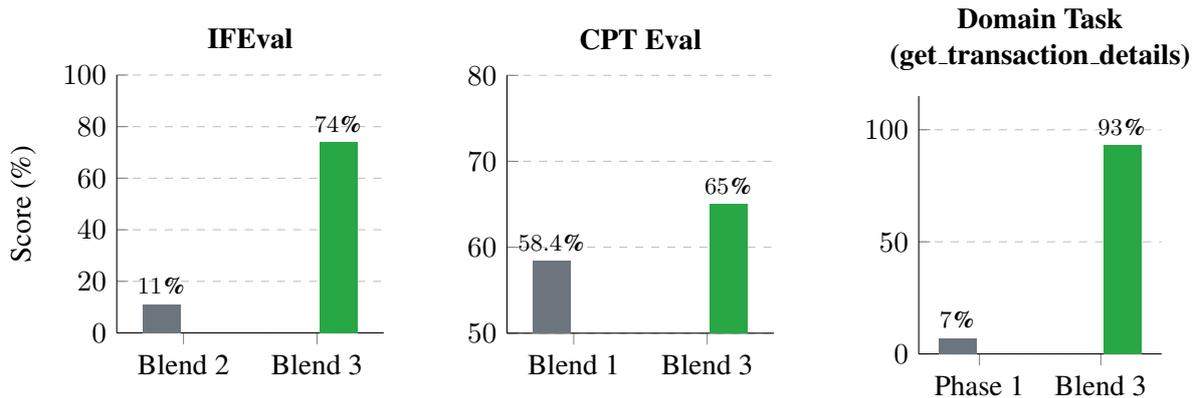
\begin{figure}[htbp]
    \centering
    \definecolor{badred}{RGB}{220, 53, 69}
    \definecolor{goodgreen}{RGB}{40, 167, 69}
    \definecolor{neutralgray}{RGB}{108, 117, 125}

    \begin{minipage}{0.32\textwidth}
    \centering
    \begin{tikzpicture}
        \begin{axis}[
            ybar,
            title={\textbf{IFEval}},
            symbolic x coords={Blend2, Blend3},
            xtick={Blend2, Blend3},
            xticklabels={Blend 2, Blend 3},
            nodes near coords={\pgfmathprintnumber\pgfplotspointmeta\%},
            nodes near coords style={font=\bfseries\footnotesize},
            ymin=0, ymax=100,
            ylabel={Score (\%)},
            width=\linewidth,
            height=5cm,
            enlarge x limits=0.5,
            bar width=0.5cm,
            axis lines*=left,
            ymajorgrids=true,
            grid style=dashed,
        ]
            \addplot[fill=neutralgray, draw=none] coordinates {(Blend2,11)};
            \addplot[fill=goodgreen, draw=none] coordinates {(Blend3,74)};
        \end{axis}
    \end{tikzpicture}
    \end{minipage}%
    \hfill
    \begin{minipage}{0.32\textwidth}
        \centering
        \begin{tikzpicture}
            \begin{axis}[
                ybar,
                title={\textbf{CPT Eval}},
                symbolic x coords={Blend1, Blend3},
                xtick={Blend1, Blend3},
                xticklabels={Blend 1, Blend 3},
                nodes near coords={\pgfmathprintnumber\pgfplotspointmeta\%},
                nodes near coords style={font=\bfseries\footnotesize},
                ymin=50, ymax=80, 
                width=\linewidth,
                height=5cm,
                enlarge x limits=0.5,
                bar width=0.5cm,
                axis lines*=left,
                ymajorgrids=true,
                grid style=dashed,
            ]
                \addplot[fill=neutralgray, draw=none] coordinates {(Blend1,58.4)};
                \addplot[fill=goodgreen, draw=none] coordinates {(Blend3,65)};
            \end{axis}
        \end{tikzpicture}
    \end{minipage}%
    \hfill
    \begin{minipage}{0.3\textwidth}
    \centering
    \begin{tikzpicture}
        \begin{axis}[
            ybar,
            title={Domain Task \\ (get\_transaction\_details)},
            title style={align=center, font=\bfseries}, 
            symbolic x coords={Phase1, Blend3},
            xtick={Phase1, Blend3},
            xticklabels={Phase 1, Blend 3},
            nodes near coords={\pgfmathprintnumber\pgfplotspointmeta\%},
            nodes near coords style={font=\bfseries\footnotesize},
            ymin=0, ymax=115, 
            width=\linewidth,
            height=5cm,
            enlarge x limits=0.5,
            bar width=0.5cm,
            axis lines*=left,
            ymajorgrids=true,
            grid style=dashed,
        ]
            \addplot[fill=neutralgray, draw=none] coordinates {(Phase1,7)};
            \addplot[fill=goodgreen, draw=none] coordinates {(Blend3,93)};
        \end{axis}
    \end{tikzpicture}
    \end{minipage}

    \vspace{1em} 

    \caption{Observation on Blend 3. The addition of general instruction data (Experiment 3) recovered instruction-following capabilities (IFEval) while maintaining consistent performance on both general knowledge (CPT Eval) and domain-specific tasks.}
    \label{fig:exp3_tradeoff}
\end{figure}

This approach resulted in strong performance improvements across all key metrics. We achieved 93\% accuracy on the \texttt{get\_transaction\_details} task while successfully reversing the regression in instruction following. Specifically, IFEval scores surged from 11\% in Blend 2 to 74\% in Blend 3. Furthermore, general knowledge retention improved over the initial blend (65\% vs 58.4\% on CPT Eval). This outcome confirmed that explicitly including general instruction examples alongside domain material is effective in maintaining general capabilities during the domain adaptation process, as illustrated in Figure~\ref{fig:exp3_tradeoff}.

\subsubsection{Extension: RLVR Analysis}
\label{sec:grpo_extension}

To determine whether post-training reinforcement learning could serve as a viable alternative to data blending, we extended our study by applying Group Relative Policy Optimization (GRPO) \citep{shao2024deepseekmathpushinglimitsmathematical} to the resulting SFT models.

Our experiments revealed that applying GRPO to the domain-only model (Blend 1) successfully restored its instruction-following capabilities, raising its performance to match that of Blend 3. However, applying the same optimization to Blends 2 and 3 yielded negligible gains, suggesting performance saturation for specific data configurations.

This comparison highlights a critical trade-off between resource investment and data quality. RLVR techniques like GRPO are inherently resource-intensive, requiring extensive sample rollouts and reward computations. Furthermore, they introduce significant engineering complexity compared to the relative stability of SFT, often necessitating precise hyperparameter tuning and reward engineering to ensure convergence. Given that Blend 3 achieves equivalent high performance purely through SFT, we believe that optimizing the data mixture is a substantially more efficient strategy than relying on costly and complex post-training alignment to mitigate specific performance regression.

\subsection{Observations}
\label{sec:exp_observations}

The Phase~2 model demonstrates consistent improvements over the baseline Mistral-Small-24B-Instruct on domain-specific tasks, while maintaining comparable performance on general-purpose benchmarks. Among the experimented configurations, Blend~3 provides the most balanced trade-off between domain performance and computational efficiency, and was therefore finally selected. A detailed breakdown of domain-specific tool performance is reported in \hyperref[tab:experiment_ablation] {Table~\ref*{tab:experiment_ablation}}, and the corresponding effects on general capabilities are summarized in \hyperref[tab:general_benchmarks_ablation] {Table~\ref*{tab:general_benchmarks_ablation}}.

\begin{table}[htbp]
\centering

\begin{adjustbox}{max width=\textwidth}
\begin{tabular}{llcccc}
\toprule
\textbf{Tool} & \textbf{Language} & \begin{tabular}{@{}c@{}}\textbf{Mistral Small 24} \\ \textbf{Instruct}\end{tabular} & \textbf{Blend 1} & \textbf{Blend 2} & \textbf{Blend 3} \\
\midrule

\multirow{2}{*}{get\_transaction\_details}
  & English & 88.89 & 89.46 & 92.39 & \textbf{93.47} \\
  & Hindi   & 85.62 & 88.95 & 89.97 & \textbf{92.58} \\
\midrule

\multirow{2}{*}{mandate\_summary}
  & English & 83.84 & 88.51 & 93.27 & \textbf{94.74} \\
  & Hindi   & 67.92 & \textbf{94.41} & 92.16 & 93.40 \\
\midrule

\multirow{2}{*}{mandate\_fetch}
  & English & 33.56 & 75.00 & 75.62 & \textbf{76.24} \\
  & Hindi   & 7.21  & 73.57 & 74.87 & \textbf{75.53} \\
\midrule

\multirow{2}{*}{mandate\_pause}
  & English & 15.38 & 41.55 & 36.84 & \textbf{41.98} \\
  & Hindi   & 4.08  & 36.84 & 31.43 & \textbf{40.54} \\
\midrule

\multirow{2}{*}{mandate\_revoke}
  & English & 33.34 & \textbf{81.25} & 77.42 & 77.42 \\
  & Hindi   & 14.63 & \textbf{89.55} & 64.86 & 76.67 \\
\midrule

\multirow{2}{*}{mandate\_unpause}
  & English & 35.29 & 65.11 & \textbf{72.73} & 68.18 \\
  & Hindi   & 0.00  & \textbf{70.00} & 64.86 & 51.43 \\

\bottomrule
\end{tabular}
\end{adjustbox}
\caption{Domain-Specific Tool Calling Performance (F1 Score) across Experimental Phases. Comparative analysis of the Phase 2 model against the Mistral-Small-24B-Instruct baseline.}
\label{tab:experiment_ablation}
\end{table}


\begin{table}[htbp]
\centering
\begin{adjustbox}{max width=\textwidth}
\begin{tabular}{lcccc}
\toprule
\textbf{Benchmark} & \begin{tabular}{@{}c@{}}\textbf{Mistral Small 24} \\ \textbf{Instruct}\end{tabular} & \textbf{Blend 1} & \textbf{Blend 2} & \textbf{Blend 3} \\
\midrule
ARC-C          & \textbf{65.78} & 58.95 & 59.30 & 61.86 \\
GSM8K (5 shot)          & 89.23 & 89.68 & 88.09 & \textbf{90.37} \\
HumanEval (pass@1)     & \textbf{82.92} & 75.00 & 74.39 & 81.10 \\
IFEval         & \textbf{76.14} & 11.75 & 1.50  & 74.10 \\
MMLU           & 78.93 & 76.71 & 76.87 & \textbf{78.99} \\
MMLU-Pro (5 shot)       & 60.82 & \textbf{61.90} & 60.54 & 61.88 \\
BBH            & 83.21 & 83.92 & 82.30 & \textbf{84.64} \\
MBPP           & \textbf{67.20} & 63.40 & 61.00 & 65.20 \\
GPQA Main      & 39.95 & 41.07 & \textbf{44.19} & 37.95 \\
GPQA Diamond   & 41.41 & 40.90 & 40.91 & \textbf{41.90} \\
GPQA Extended  & 37.22 & 37.18 & 36.44 & \textbf{37.45} \\
\bottomrule
\end{tabular}
\end{adjustbox}
\caption{General Benchmark Evaluation (Accuracy). Comparative results between the baseline and experimental phases, highlighting the recovery of general-purpose capabilities in Experiment~3.}
\label{tab:general_benchmarks_ablation}
\end{table}

\section{Benchmarking \& Results}
\label{sec:benchmarking_results}

Upon determining the optimal configuration for domain adaptation, the evaluation phase focused on validating the model's efficacy within the broader open-source ecosystem. While demonstrating local improvements over the baseline is a necessary prerequisite, a definitive assessment requires comparative benchmarking against its foundational base (Mistral 24B Instruct 2501) and a diverse cohort of prominent open-weight models.

\subsection{Domain-Specific Proficiency}
\label{sec:domain_proficiency}

To ensure a controlled evaluation aligned with this objective, the system prompt remained consistent across all models during testing.

\begin{figure}[htbp]
    \centering
    \includegraphics[width=0.9\linewidth]{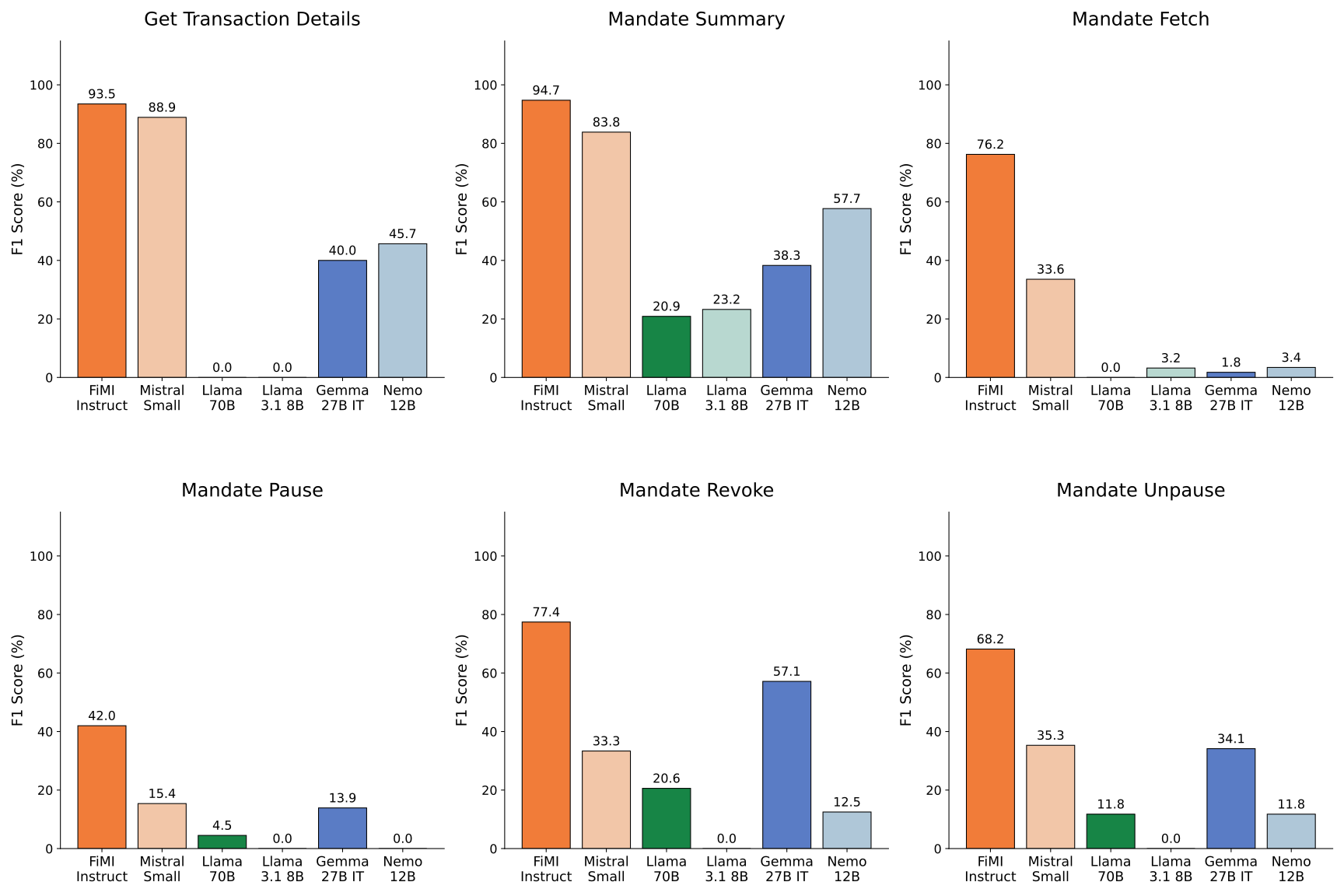}
    \caption{Performance Evaluation on English Queries. Comparative analysis of F1 scores demonstrating the efficacy of the Fine-Tuned FiMi-Instruct model against open-source baselines on the custom domain benchmark.}
    \label{fig:english_tools}
\end{figure}

\begin{figure}[htbp]
    \centering
    \includegraphics[width=0.9\linewidth]{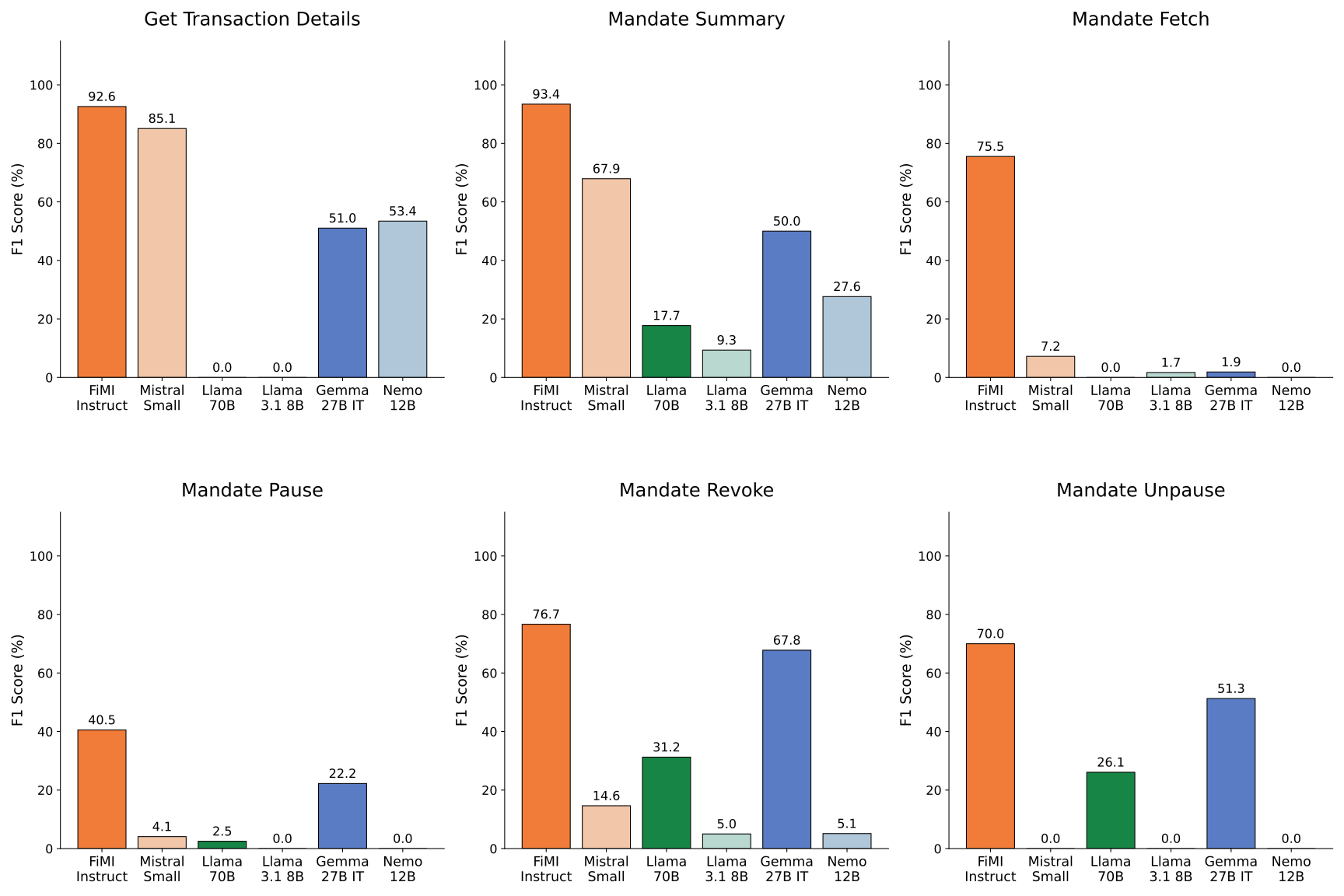}
    \caption{Performance Evaluation on Multilingual (Hindi) Queries. Comparative analysis of F1 scores demonstrating the efficacy of the fine-tuned FiMi-Instruct model against open-source baselines on the Hindi-translated custom domain benchmark.}
    \label{fig:hindi_tools}
\end{figure}

The empirical results demonstrate the performance of the specialized model and off-the-shelf generalist LLMs on custom domain benchmarks. These findings are visualized in \hyperref[fig:english_tools]{Figure~\ref*{fig:english_tools}} and \hyperref[fig:hindi_tools]{Figure~\ref*{fig:hindi_tools}}.

The results shown in \hyperref[fig:english_tools]{Figure~\ref*{fig:english_tools}} illustrate differences in tool-calling behavior across models, with the FiMI Instruct model exhibiting more consistent behavior on multi-step workflows. While the Mistral 24B Instruct performs comparably on simpler retrieval tasks such as get\_transaction\_details, differences become more apparent on multi-step mandate operations (fetch, pause, revoke), which involve sequential decision-making. For example, on mandate\_fetch, the fine-tuned model attains an accuracy of 76.24\% compared to 33.56\% for the baseline, while generalist models show limited zero-shot robustness under the domain-specific tool schema. These results suggest that domain-specific fine-tuning supports more reliable execution of dependent tool sequences.

Cross-lingual results, summarized in \hyperref[fig:hindi_tools]{Figure~\ref*{fig:hindi_tools}}, indicate that the FiMI Instruct maintains similar behavior across English and Hindi inputs. Accuracy on get\_transaction\_details remains comparable across languages (93.47\% in English and 92.58\% in Hindi). In contrast, general-purpose baselines exhibit larger variations on complex workflows under Hindi inputs; for instance, the Mistral 24B Instruct shows a reduction on mandate\_fetch from 33.56\% in English to 7.21\% in Hindi. Overall, these observations are consistent with the hypothesis that domain-specific fine-tuning improves cross-lingual robustness in structured tool-calling scenarios.

\subsection{General Benchmarks}
\label{sec:general_benchmarks_comparison}

Having established the model's performance on domain-specific tasks in \hyperref[fig:english_tools]{Figures~\ref*{fig:english_tools}} and \hyperref[fig:hindi_tools]{\ref*{fig:hindi_tools}}, we next assess whether this specialization compromised its foundational capabilities. To verify that core competencies  such as reasoning and instruction following were preserved, we evaluated the model against standard global benchmarks. The results are presented in \hyperref[tab:general_benchmark_comparison]{Table~\ref*{tab:general_benchmark_comparison}}.

\begin{table}[htbp]
\centering
\begin{adjustbox}{max width=\textwidth}
\begin{tabular}{l>{\columncolor{npciorange!20}}ccccccc}
\hline
\textbf{Benchmark} &
\textbf{FiMI Instruct} &
\textbf{Mistral 24B} &
\textbf{Llama 70B} &
\textbf{Llama 3.1 8B} &
\textbf{Gemma 3 27B} &
\textbf{NeMo 12B IT} &
\textbf{Mistral 7B} \\
\hline
ARC-C         & 61.86\% & \textbf{65.78\%} & 62.29\% & 54.18\% & 60.41\% & 59.39\% & 59.98\% \\
\hline
GSM8K (5 shot)          & 90.37\% & 89.23\% & \textbf{92.65\%} & 50.42\% & 92.34\% & 74.75\% & 47.99\% \\
\hline
HumanEval (pass@1)      & 81.10\% & 82.92\% & 81.10\% & 42.07\% & \textbf{87.20\%} & 67.07\% & 39.02\% \\
\hline
IFEval        & 74.10\% & 76.14\% & \textbf{86.57\%} & 14.03\% & 84.77\% & 56.71\% & 59.95\% \\
\hline
MMLU         & 77.99\% & 78.93\% & \textbf{81.78\%} & 62.98\% & 76.50\% & 76.50\% & 59.67\% \\
\hline
MMLU-Pro (5 shot)       & 59.88\% & 60.82\% & 63.14\% & 35.77\% & \textbf{65.15\%} & 44.47\% & 35.83\% \\
\hline
BBH            & 84.64\% & 83.21\% & \textbf{85.39\%} & 62.80\% & 80.52\% & 71.89\% & 56.19\% \\
\hline
MBPP         & 65.20\% & 67.20\% & 70.40\% & 48.60\% & \textbf{72.60\%} & 57.20\% & 40.00\% \\
\hline
GPQA Main     & 37.95\% & 39.95\% & \textbf{43.08\%} & 28.57\% & 37.72\% & 34.59\% & 33.92\% \\
\hline
GPQA Diamond  & 39.90\% & \textbf{41.41\%} & \textbf{41.41\%} & 28.28\% & 38.88\% & 34.34\% & 31.31\% \\
\hline
GPQA Extended & 36.45\% & 37.91\% & \textbf{42.31\%} & 31.13\% & 37.36\% & 32.42\% & 30.95\% \\
\hline
\end{tabular}
\end{adjustbox}
\caption{Comparing Domain-Specific Fine-Tuning against leading Open-Source Instruct Models on General Benchmarks.}
\label{tab:general_benchmark_comparison}
\end{table}

The results indicate that the domain-specific fine-tuning process successfully mitigated significant performance degradation, maintaining parity with the foundational model across most general reasoning tasks. As shown in \hyperref[tab:general_benchmark_comparison]{Figure~\ref*{tab:general_benchmark_comparison}}, the fine-tuned model exhibits minimal deviation from the Mistral~24B baseline on core knowledge benchmarks such as MMLU (77.99\% vs 78.93\%) and coding tasks including HumanEval (81.10\% vs 82.92\%). Notably, the model demonstrates resilience in complex reasoning, slightly outperforming the baseline in mathematical logic (GSM8k: +1.14\%) and algorithmic reasoning (BBH: +1.43\%) \citep{suzgun2022challengingbigbenchtaskschainofthought}. These results suggest that the reduced learning rate effectively preserves the model’s pre-trained logical abstractions. Within the broader ecosystem, the domain-tuned model remains highly competitive within its weight class. It outperforms larger counterparts such as Gemma~3~27B on multi-task accuracy (MMLU) and hard reasoning (BBH), while maintaining a substantial lead over smaller, efficiency-focused models including Llama~3.1~8B and NeMo~12B. Although an expected performance gap remains when compared against significantly larger parameter models (e.g., Llama~70B), the empirical evidence confirms that the proposed model retains robust generalist capabilities even after specialized adaptation for financial tool usage.

\section{Conclusion}
\label{sec:conclusion}
In this report, we introduce FiMI Base and FiMI Instruct, providing a transparent framework for adapting large language models to specific domains. We have detailed our complete methodological recipe, including data sources, data curation processes, and a strategic multi-stage training approach. These models achieve benchmarks comparable to other baselines while utilising limited computational resources. We believe this report will serve as a useful reference for the community in developing high utility checkpoints in their respective domains.
In the near future, our research will focus on intersection of high-quality data scaling and architectural efficiency to address more specialized use cases. Additionally, we aim to develop smaller models specifically optimized for agentic AI workloads, balancing computational efficiency while scaling to longer contexts.

\section{Contributors}
\label{sec:contributors}

{
\noindent\textbf{Leadership:}
Prashant Devadiga, Vishal Kanvaty

\noindent\textbf{Continuous Pre-training:}
Duddu Prasanth Kumar, Kolisetty Krishna SK, Krishanu Adhikary, Navya Prakash, Prashant Devadiga, Ratanjeet Pratap Chauhan, Shantanu Pandey, Tulasi Pilla

\noindent\textbf{Synthetic Data Generation and Data Engineering:}
Aman Kumar, Chandra Bhushan, Harsh Sharma, Hrithik Kadam, Keyur Doshi, Krishanu Adhikary, Nitin Kukreja, Prashant Devadiga, Shantanu Pandey, Suraj Singh, Suvradip Paul, Tulasi Pilla, Yatharth Dedhia

\noindent\textbf{Post-Training:}
Aman Kumar, Chandra Bhushan, Harsh Sharma, Keyur Doshi, Nitin Kukreja, Suraj Singh, Suvradip Paul, Yatharth Dedhia

\noindent\textbf{Evaluation:}
Aboli Kathar, Ashish Sharma, Divya Sorate, Lokesh MPT, Mayurdeep Sonowal, Shubham Soni, Siddharth Dixit, Smriti Jopat

\noindent\textbf{Product:}
Hrithik Kadam

\noindent\textbf{Infrastructure:}
Araveeti Srujan, Nadeem Shaikh, Shamanth MH

\noindent\textbf{Nvidia Team:}
Anusha Kamath,
Evan Acharya,
Kanishk Singla,
Kiran Praveen,
Nimit Kothari,
Rakesh Paul,
Raunak Kalani,
Raviraj Joshi,
Sunil Patel,
Utkarsh Vaidya,
Vineeth Nambiar
}

\newpage
\bibliographystyle{plainnat}
\bibliography{references}


\newpage
\section*{\texorpdfstring{{Annexure A}}{Annexure}}
\phantomsection
\label{sec:annexuremmlu}
\section*{MMLU Dataset Analysis}

\subsection*{A. Introduction}

This section presents an analysis of the MMLU dataset, specifically key metrics, trends, and the distribution of questions. This analysis has been utilized within NVIDIA's Sovereign AI playbook containing BYOB framework to construct an evaluation dataset similar to the MMLU benchmark which has been discussed in \S\ref{sec:cpt}. The created dataset should have similar distribution in terms of complexity, structure, ambiguity, etc.
\subsection*{B. Procedure to Analyse the MMLU Dataset}
We only focused on finance subjects by extracting relevant MMLU subjects (e.g., accounting, econometrics). A rigorous, research-informed framework was developed to categorize questions based on cognitive complexity (Difficulty, Bloom's Level) and structure (Question Type). We built an LLM-based pipeline to systematically apply these metrics. Standard data analysis was then performed on the categorized dataset to generate the distributions and insights presented in this report.

\subsection*{C. Subjects Selected for Analysis}
The following subjects were selected based on their closest relevance to the finance domain:
\begin{enumerate}
  \setlength{\itemsep}{0pt}
  \setlength{\parskip}{0pt}
  \setlength{\parsep}{0pt}
    \item high\_school\_macroeconomics
    \item high\_school\_microeconomics
    \item professional\_accounting
    \item business\_ethics
    \item econometrics
    \item management
\end{enumerate}

\subsection*{D. Categorization Methodology, Metrics, and Distribution}

\hyperref[tab:categorization_methodology_metrics]{Table~\ref*{tab:categorization_methodology_metrics}} summarizes the metrics used to categorize each question in the MMLU dataset. Using these metrics, questions are assigned to their respective categories, and the resulting distribution of questions across categories is summarized in \hyperref[tab:distribution]{Table~\ref*{tab:distribution}}.

\begin{table}[htbp]
\centering
\small
\begin{tabular}{
>{\raggedright\arraybackslash}p{3.5cm}
>{\raggedright\arraybackslash}p{5.5cm}
>{\raggedright\arraybackslash}p{5.5cm}
}
\hline
\textbf{Characteristic} &
\textbf{Description} &
\textbf{Categorization Scale / Basis} \\
\hline

\textbf{Difficulty Level}
& Subjective assessment of the complexity of the underlying concept and the intellectual effort required for a human to answer correctly.
& \textbf{Low, Medium, High} \\
\hline

\textbf{Question Type}
& Primary skill necessary to arrive at the correct answer.
& \textbf{Knowledge-Based} (requires recall of facts), \textbf{Reasoning-Based} (requires logical inference, application, or calculation). \\
\hline

\textbf{Bloom's Taxonomy Level}
& Alignment with Bloom's Revised Taxonomy, classifying the cognitive skill tested by the question.
& Remember / Understand, Apply, Analyze / Evaluate / Create \\
\hline

\textbf{Word Count}
& The length of the question text and prompt (excluding the four options), used as a proxy for reading comprehension load.
& \textbf{Short} ($<$25 words), \textbf{Medium} (25--40 words), \textbf{Long} ($>$50 words) \\
\hline

\textbf{Ambiguity}
& Assessment of the clarity of the question and the distinctness of the correct answer from the distractors.
& \textbf{0--1}, lower indicates better clarity in the question \\
\hline

\textbf{Distractor Quality}
& Assessment of the options, measuring how similar the options are and hence how hard it is to select the correct one.
& \textbf{Low, Medium, High}, low indicates options are very dissimilar \\
\hline

\textbf{Question Category}
& Classifying the question into predefined categories.
& conceptual\_based / definition\_type / sentence\_completion / numericals / context\_based\_question etc. \\
\hline

\end{tabular}
\caption{Categorization Methodology and Metrics.}
\label{tab:categorization_methodology_metrics}
\end{table}


\begin{table}[htbp]
\centering
\small
\begin{tabular}{
>{\raggedright\arraybackslash}p{5.5cm}
>{\raggedright\arraybackslash}p{1.5cm}
>{\raggedright\arraybackslash}p{2.5cm}
}
\hline
\textbf{Category} &
\textbf{Count} &
\textbf{Percentage (\%)} \\
\hline

Difficulty: High & 26 & 2.12 \\
\hline
Difficulty: Medium & 614 & 50.04 \\
\hline
Difficulty: Low & 587 & 47.84 \\
\hline
Type: Knowledge-Based & 707 & 57.62 \\
\hline
Type: Reasoning-Based & 4 & 0.33 \\
\hline
Type: Hybrid & 516 & 42.05 \\
\hline
Bloom’s: Remember & 560 & 45.64 \\
\hline
Bloom’s: Apply & 317 & 25.84 \\
\hline
Bloom’s: Understand & 207 & 17.03 \\
\hline
Bloom’s: Analyse & 122 & 9.94 \\
\hline
Bloom’s: Evaluate & 19 & 1.55 \\
\hline
Distractor Quality: High & 48 & 3.90 \\
\hline
Distractor Quality: Moderate & 1168 & 95.20 \\
\hline
Distractor Quality: Low & 10 & 0.80 \\
\hline
Question Type: Conceptual Based & 329 & 26.83 \\
\hline
Question Type: Definition Type & 287 & 23.39 \\
\hline
Question Type: Sentence Completion & 185 & 15.09 \\
\hline
Question Type: Numericals & 142 & 11.58 \\
\hline
Question Type: Context Based & 108 & 8.81 \\
\hline
Question Type: Fill in the Blanks & 85 & 6.94 \\
\hline
Question Type: True or False & 78 & 6.36 \\
\hline
Question Type: Statement Analysis & 13 & 1.06 \\
\hline

\end{tabular}
\caption{Distribution of MMLU Questions Across Categories}
\label{tab:distribution}
\end{table}


    
    
    




\subsubsection*{Insights:}\label{insights-from-above-table}

\begin{itemize}
  \setlength{\itemsep}{0pt}
  \setlength{\parskip}{0pt}
  \setlength{\parsep}{0pt}
\item
  \textbf{Difficulty Skew:} The MMLU dataset is heavily skewed towards
  the Medium (50.04\%) and Low (47.84\%) difficulty
  levels, collectively accounting for over 97\% of the
  questions. High difficulty questions are rare (2.12\%)
\item
  \textbf{Dominance of Knowledge and Recall:}

  \begin{itemize}
    \setlength{\itemsep}{0pt}
  \setlength{\parskip}{0pt}
  \setlength{\parsep}{0pt}
  \item
    \textbf{Question Type:} Purely Knowledge-Based questions
    are the most common (57.62\%). The dataset contains a significant
    portion of Hybrid questions (42.05\%) that require both
    knowledge and reasoning, but pure Reasoning-Based questions
    are virtually non-existent (0.33\%).
  \item
    \textbf{Bloom\textquotesingle s Taxonomy:} The majority of the
    dataset tests lower-order cognitive skills, with Remember
    (45.64\%) and Apply (25.84\%) being the most prevalent.
    Higher-order thinking skills like Analyse (9.94\%) and
    Evaluate (1.55\%) are substantially less represented.
  \item
    \textbf{Distractor Quality:} A very small percentage (0.039\%) of
    questions have High Distractor Quality, indicating that in
    most cases, the incorrect options (distractors) are relatively
    dissimilar from the correct answer
  \end{itemize}
\end{itemize}

\subsubsection*{\texorpdfstring{Correlation Graphs:
}{Correlation Graphs: }}\label{correlation-graphs}



\begin{figure}[H]
    \centering
    \includegraphics[width=1.0\linewidth]{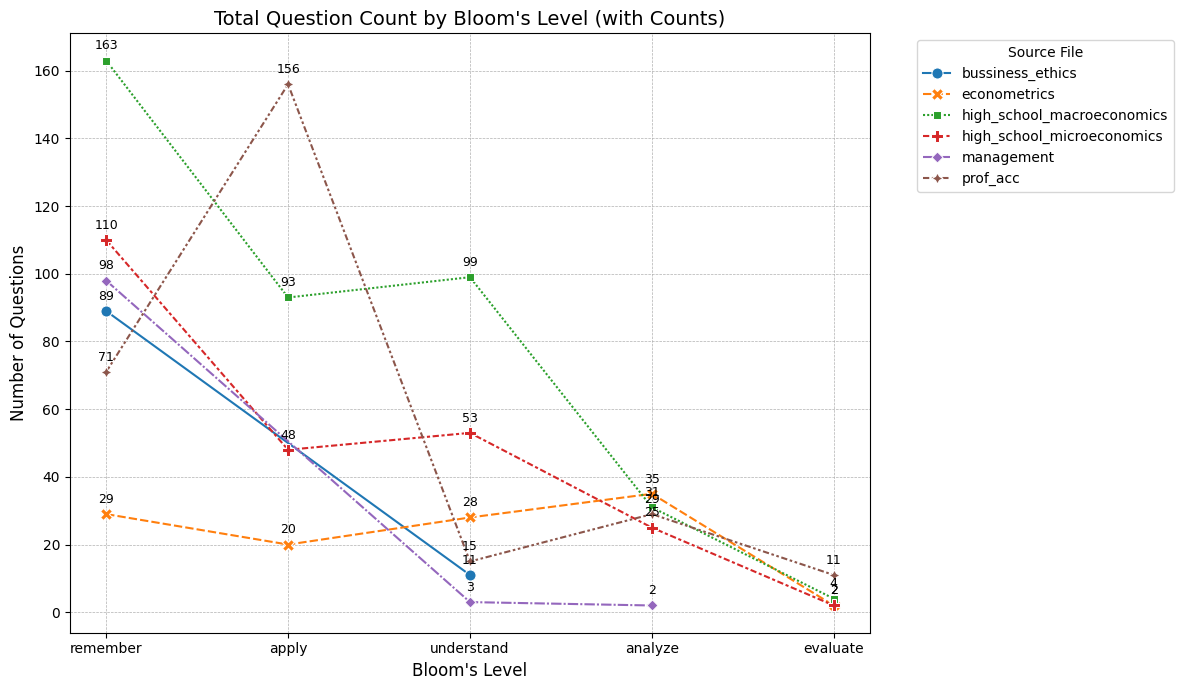}
    \caption{Bloom level counts in each finance related subject from MMLU }
    \label{fig:placeholder}
\end{figure}




\begin{figure}[H]
    \centering
    \includegraphics[width=1.0\linewidth]{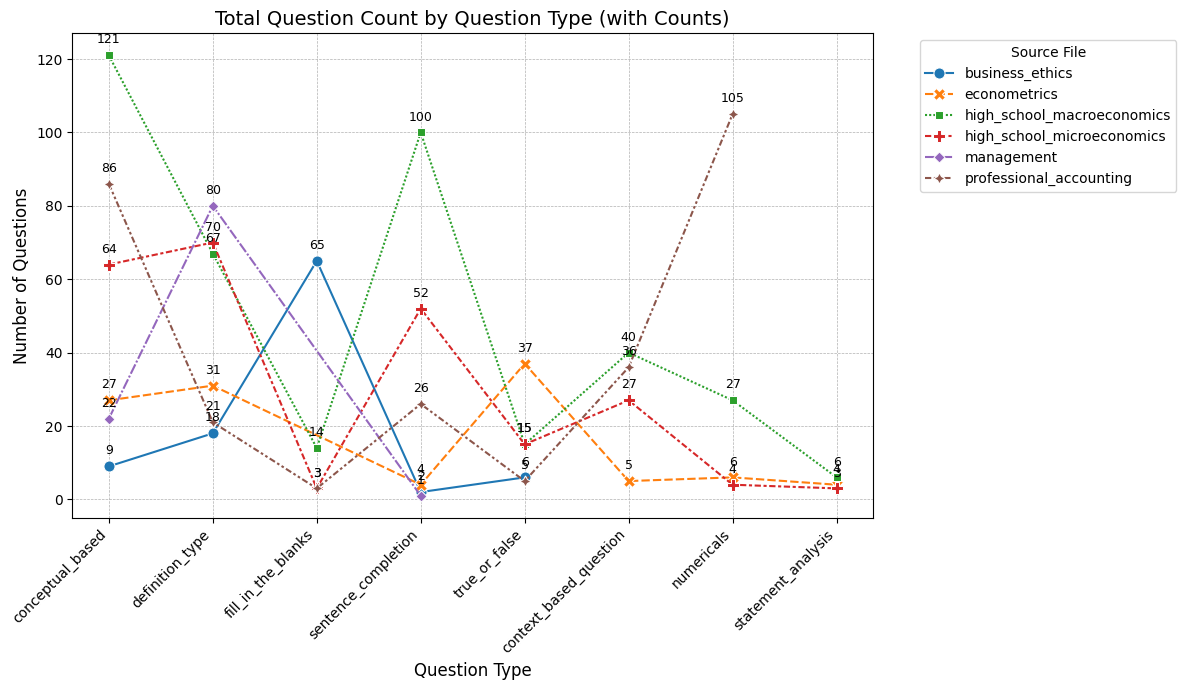}
    \caption{Question type counts in each finance related subject from MMLU }
    \label{fig:placeholder}
\end{figure}

\begin{figure}[H]
    \centering
    \includegraphics[width=1.0\linewidth]{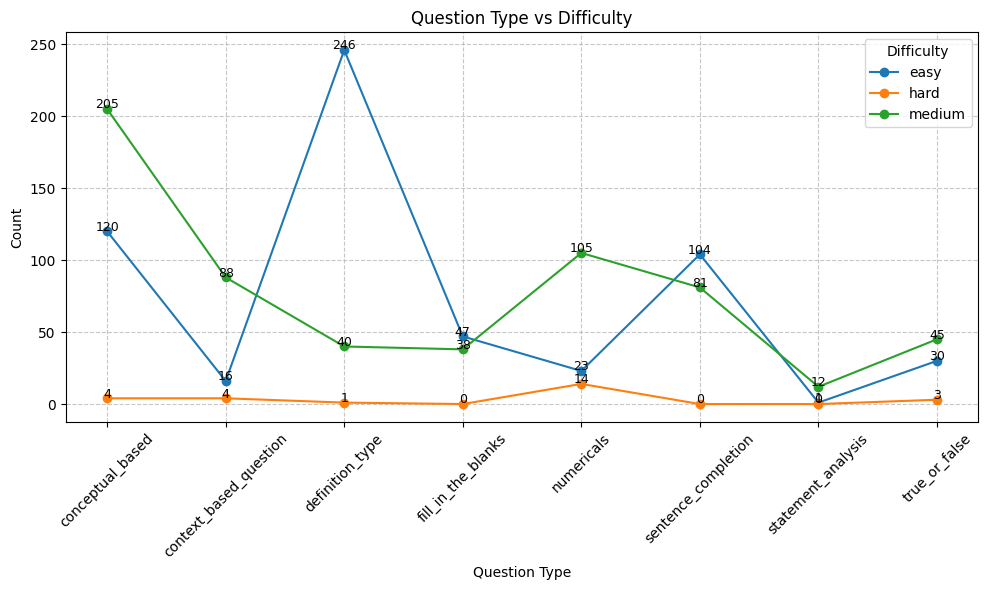}
    \caption{Question Type counts in each Difficulty Level }
    \label{fig:placeholder}
\end{figure}

\begin{figure}[H]
    \centering
    \includegraphics[width=1.0\linewidth]{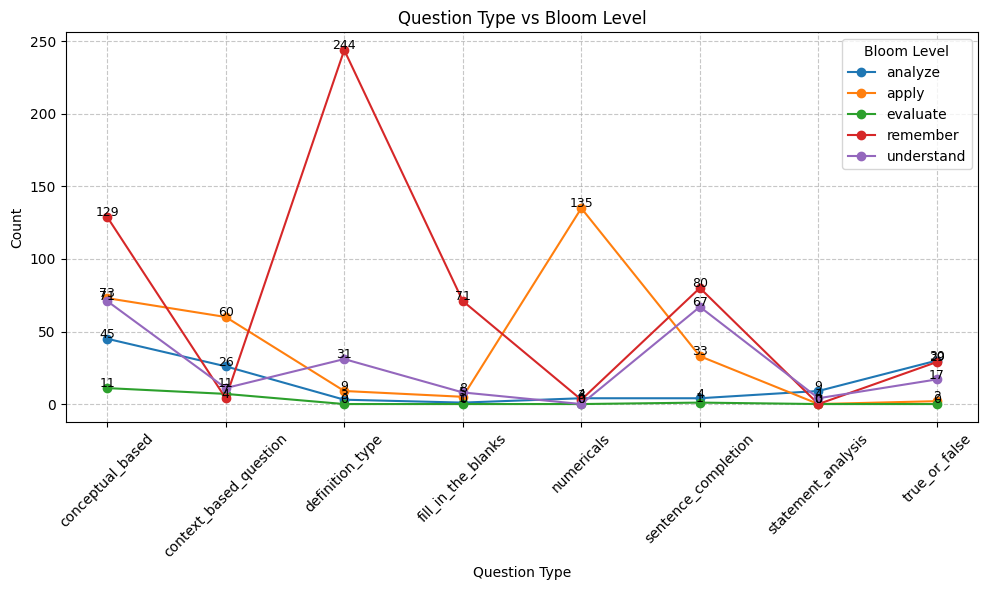}
    \caption{Question Type counts in each Bloom Level }
    \label{fig:placeholder}
\end{figure}

\newpage
\section*{\texorpdfstring{{Annexure B}}{Annexure}}
\phantomsection
\section*{Optimizing the Agentic Pipeline: From Heavy Prompting to Supervised Fine-Tuning}
\label{sec:agentic_upi_help}
\subsection*{A. Motivation and Use-Case Definition}

To operationalize large language models in real-world financial workflows, we designed UPI HELP as an agentic system tailored to the UPI product ecosystem. The objective was to move beyond a generic conversational chatbot and establish a production-grade assistant grounded in concrete, high-impact use cases. The system was structured around three primary pillars:

\begin{itemize}
  \setlength{\itemsep}{0pt}
  \setlength{\parskip}{0pt}
  \setlength{\parsep}{0pt}
    \item \textbf{Customer Grievance Redressal}: Live transaction status checks, complaint registration, and grievance resolution flows.
    \item \textbf{Mandate Management}: Autopay mandate discovery, summaries, pausing, and revocation across the mandate lifecycle.
    \item \textbf{User Awareness and Safety}: Fraud prevention guidance, scam education, and contextual product awareness.
\end{itemize}

These workflows require strict rule adherence, reliable tool invocation, and consistent safety behavior, making them unsuitable for prompt-only conversational systems.

\subsection*{B. System Architecture: Swarm-Based Agentic Design}

To support heterogeneous responsibilities, we adopted a Swarm Architecture rather than a monolithic agent. Each functional domain (e.g., grievance handling, mandate management, and safety enforcement) was handled by a specialized sub-agent.

The system was integrated using the Model Context Protocol (MCP), enabling interaction with over 20 tools across transaction lookup, mandate operations, grievance systems, and safety checks. MCP also allowed agents themselves to be composed and invoked as tools, resulting in a modular, extensible, and production-aligned architecture.

\subsection*{C. Limitations of Prompt-Only Specialization}

Our initial prototype relied on a strong general-purpose base model with extensive prompt engineering for domain specialization. While functional, this approach introduced significant inefficiencies, referred to as the ``heavy prompting tax'':

\begin{itemize}
    \item \textbf{High Token Overhead}:
    \begin{itemize}
      \setlength{\itemsep}{0pt}
  \setlength{\parskip}{0pt}
  \setlength{\parsep}{0pt}
        \item Grievance Agent: $\sim$5{,}000 tokens for system prompts ($\sim$7{,}000 including tool definitions)
        \item Mandate Agent: $\sim$3{,}000 tokens for correct behavioral grounding
        \item Safety Enforcement: A separate LLM-based validator agent costing $\sim$5{,}000 tokens per user query
    \end{itemize}
    \item \textbf{Operational Cost and Latency}: The cumulative prompt and validation overhead increased inference cost, latency, and reduced scalability, making the system unsuitable for high-volume, real-time payment support.
\end{itemize}

\subsection*{D. Reliability Gaps in Base Model Behavior}

Despite reasonable surface-level performance, deeper evaluation revealed critical reliability issues:

\begin{itemize}
  \setlength{\itemsep}{0pt}
  \setlength{\parskip}{0pt}
  \setlength{\parsep}{0pt}
    \item \textbf{Intent Identification}: Accuracy fluctuated around 70\% unless user queries were unusually explicit.
    \item \textbf{Mandate Lifecycle Handling}: The base model struggled with nuanced, multi-step mandate operations, rendering several tools effectively unusable in practice.
\end{itemize}

These observations highlighted the limitations of prompt-based alignment for tightly constrained financial workflows.

\subsection*{E. Supervised Fine-Tuning on UPI-Specific Data}

To address these limitations, we applied SFT using curated UPI-specific datasets covering real user intents, mandate flows, grievance scenarios, and safety-critical interactions. The resulting performance improvements were immediate and measurable.

\subsubsection*{1. Tool-Calling Reliability Improvements}

\begin{table}[h]
\centering
\begin{tabular}{lcc}
\hline
\textbf{Mandate Operation} & \textbf{Base Model} & \textbf{Post-SFT} \\
\hline
Mandate Pause  & 3.27\%  & 41.98\% \\
Mandate Fetch  & 40.00\% & 76.24\% \\
Mandate Revoke & 46.15\% & 77.42\% \\
\hline
\end{tabular}
\caption{Mandate Tool Invocation Accuracy}
\end{table}

SFT effectively restored tools that were non-functional under the base model, enabling stable and repeatable execution across the mandate lifecycle.

\subsection*{F. Intrinsic Safety vs. External Validation}

Safety is a first-order requirement in payment ecosystems.

\begin{itemize}
  \setlength{\itemsep}{0pt}
  \setlength{\parskip}{0pt}
  \setlength{\parsep}{0pt}
    \item \textbf{Base Model}: Safety enforcement was entirely prompt-dependent. When system prompts were removed, the block rate against adversarial financial queries dropped to 0\%, necessitating an external LLM-based validator.
    \item \textbf{SFT Model}: The fine-tuned model demonstrated intrinsic safety alignment:
    \begin{itemize}
      \setlength{\itemsep}{0pt}
  \setlength{\parskip}{0pt}
  \setlength{\parsep}{0pt}
        \item 39\% block rate against unsafe queries even without system prompts
        \item $\sim$90\% safety score with minimal prompting
        \item Elimination of the need for a 5{,}000-token external validation agent
    \end{itemize}
\end{itemize}

The SFT model significantly outperformed the base model’s peak safety score of approximately 74\%.

\subsection*{G. Outcome and Operational Impact}

While the base model enabled a functional proof-of-concept, it produced generic responses and incurred high operational costs due to extensive prompt overhead.

Supervised Fine-Tuning delivered multiple compounding benefits:

\begin{itemize}
  \setlength{\itemsep}{0pt}
  \setlength{\parskip}{0pt}
  \setlength{\parsep}{0pt}
    \item Significant reduction in prompt-token overhead
    \item Elimination of external LLM-based safety validators
    \item Substantial improvement in tool-calling accuracy
    \item More consistent, domain-appropriate conversational behavior
    \item Reduced latency and lower inference cost
\end{itemize}

Overall, SFT transformed the system from a prompt-heavy prototype into a production-viable, cost-efficient, and safety-aligned agentic platform suitable for deployment in real-world financial operations.

\textbf{Disclaimer}

The information contained in this paper is intended solely for general informational and research purposes. While reasonable efforts have been made to maintain the accuracy of the information, findings, and descriptions related to the Small Language Model (SLM), no representation, warranty, or guarantee of any kind, whether express or implied, is made regarding the completeness, accuracy, reliability, suitability, performance, fitness for a particular purpose, or availability of the information, code, architecture, or methodologies described herein.

This document does not constitute professional advice (including but not limited to technical, security, legal, regulatory, or operational advice). Any reliance on the content of this paper is strictly at the reader’s own discretion and risk.

Neither the authors nor the National Payments Corporation of India (NPCI), including the subsidiaries, shall be liable for any loss or damage of any kind, including without limitation direct, indirect, incidental, special, consequential, or punitive losses arising out of or in connection with the use of, or reliance on, this document or the SLM described in it.

This work may reference concepts in Artificial Intelligence, Machine Learning, or Responsible AI. Such references are illustrative in nature and do not imply that the SLM adheres to, complies with, or is certified against any specific Responsible AI framework, guideline, or standard. Users are solely responsible for validating the model’s applicability, safety, compliance, and performance within their respective environments.

NPCI and the authors reserve the right to update, modify, or withdraw the content of this paper at any time without prior notice.

\end{document}